\documentclass[11pt,english]{article}
\usepackage[T1]{fontenc}
\usepackage[latin9]{inputenc}
\usepackage[letterpaper]{geometry}
\usepackage{color}
\usepackage{babel}
\usepackage{float}
\usepackage{amsthm}
\usepackage{amsmath}
\usepackage{amssymb}
\usepackage{graphicx}
\usepackage[unicode=true,pdfusetitle,
 bookmarks=true,bookmarksnumbered=false,bookmarksopen=false,
 breaklinks=false,pdfborder={0 0 0},backref=false,colorlinks=true]
 {hyperref}
\hypersetup{
 linkcolor=blue, citecolor=blue}
\usepackage{breakurl}

\makeatletter

\providecommand{\tabularnewline}{\\}
\floatstyle{ruled}
\newfloat{algorithm}{tbp}{loa}
\providecommand{\algorithmname}{Algorithm}
\floatname{algorithm}{\protect\algorithmname}

\theoremstyle{plain}
\newtheorem{assumption}{\protect\assumptionname}
\theoremstyle{definition}
\newtheorem*{defn*}{\protect\definitionname}
\newtheorem{defn}{\protect\definitionname}
\theoremstyle{plain}
\newtheorem{thm}{\protect\theoremname}
\theoremstyle{plain}
\newtheorem{cor}{\protect\corollaryname}
\theoremstyle{remark}
\newtheorem*{rem*}{\protect\remarkname}
\theoremstyle{remark}
\newtheorem{rem}{\protect\remarkname}
\theoremstyle{plain}
\newtheorem{lem}{\protect\lemmaname}
\theoremstyle{plain}
\newtheorem{prop}{\protect\propositionname}

\usepackage{appendix}
\usepackage{algorithmic}

\makeatother

  \providecommand{\assumptionname}{Assumption}
  \providecommand{\definitionname}{Definition}
  \providecommand{\lemmaname}{Lemma}
  \providecommand{\propositionname}{Proposition}
  \providecommand{\remarkname}{Remark}
\providecommand{\corollaryname}{Corollary}
\providecommand{\theoremname}{Theorem}

\begin{document}
\global\long\def\P{\mathcal{P}}

\global\long\def\T{T}

\global\long\def\I{I}

\global\long\def\Ic{I^{c}}

\global\long\def\PT{\mathcal{\P}_{\bar{\T}}}

\global\long\def\PTO{\mathcal{\P}_{\T_{0}}}

\global\long\def\PI{\mathcal{\P}_{\bar{\I}}}
\global\long\def\PIc{\mathcal{\P}_{\bar{\I}^{c}}}

\global\long\def\PIO{\mathcal{\P}_{\I_{0}}}
\global\long\def\PIOc{\mathcal{\P}_{\I_{0}^{c}}}

\global\long\def\PF{\mathcal{\P}_{\bar{\Phi}}}
\global\long\def\PFc{\mathcal{\P}_{\bar{\Phi}^{c}}}

\global\long\def\RR{\mathbb{R}}

\global\long\def\R{\mathcal{R}}

\global\long\def\sgn{\textrm{sign}}

\global\long\def\mc#1{\mathcal{#1}}

\global\long\def\Omegat{\tilde{\Omega}}

\title{Matrix completion with column manipulation: Near-optimal sample-robustness-rank tradeoffs}

\author{Yudong Chen, Huan Xu, Constantine Caramanis, Sujay Sanghavi\thanks{Y. Chen is with the School of Operations Research and Informaiton Engineering, Cornell University  (yudong.chen@cornell.edu). H. Xu is with the Department of Mechanical Engineering, National University of Singapore (mpexuh@nus.edu.sg). C. Caramanis and S. Sanghavi are with the Department of Electrical and Computer Engineering, the University of Texas at Austin (constantine@utexas.edu, sanghavi@mail.utexas.edu). 
This paper was presented in part at the International Conference on Machine Learning, 2011.}}

\date{}

\maketitle

\begin{abstract}
This paper considers the problem of matrix completion when some number
of the columns are completely and arbitrarily corrupted, potentially
by a malicious adversary. It is well-known that standard algorithms
for matrix completion can return arbitrarily poor results, if even
a single column is corrupted. One direct application comes from robust
collaborative filtering. Here, some number of users are so-called
manipulators who try to skew the predictions of the algorithm by calibrating their inputs to the system. In this paper, we develop an efficient algorithm for this problem based on a combination of a trimming procedure and a convex program that minimizes the nuclear norm and the $\ell_{1,2}$ norm. Our theoretical results show that given a vanishing fraction of observed entries, it
is nevertheless possible to complete the underlying matrix
even when the number of corrupted columns grows. Significantly, our
results hold without any assumptions on the locations or values of the
observed entries of the manipulated columns. Moreover, we show by an information-theoretic argument that our guarantees are nearly optimal in terms of the fraction of sampled entries on the authentic columns, the fraction of corrupted columns, and the rank of the underlying matrix. Our results therefore
sharply characterize the tradeoffs between sample, robustness and rank in matrix
completion. 
\end{abstract}

\section{Introduction}

Previous work in low-rank matrix completion~\cite{candes2009exact,candes2010NearOptimalMC,gross2009anybasis,chen2013incoherence_arxiv}
has demonstrated the following remarkable fact: given a $m\times n$
matrix of rank $r$, if its entries are sampled uniformly at random,
then with high probability, the solution to a convex and in particular
tractable optimization problem yields exact reconstruction of the
matrix when only $O((m+n)r\log(m+n))$ entries are sampled. 

Yet as our simulations demonstrate, if even a few columns of this
matrix are corrupted, the output of these algorithms can be arbitrarily
skewed from the true matrix. This problem is particularly relevant
in so-called collaborative filtering, or recommender systems. Here,
based on only partial observation of users' preferences, one tries
to obtain accurate predictions for their unrevealed preferences. It
is well known and well documented~\cite{lam2004shilling,van2010manipulation}
that such recommender systems are susceptible to manipulation by malicious
users who can calibrate \emph{all} their inputs adversarially. It
is of great interest to develop efficiently scalable algorithms that
can successfully predict preferences of the honest users based on
the corrupted and partially observed data, while identifying the manipulators. 

The presence of partial observation and potentially adversarial input
makes \textit{a priori} identification of corrupted column versus good
column a challenging task. For example, a simple method that works fairly well
under full observation and purely random corruption, is to use
the correlation between the columns. Since the authentic columns of a low-rank
matrix are linearly correlated, under suitable conditions they can
be identified as those which have a high correlation with many other
columns. However, when partial observations are present, this method
fails since it is not immediately clear even how to compute the correlation---most
pairs of columns do not share an observed coordinate---let alone
finding the corrupted columns which can disguise themselves to look
like a partially observed authentic column. At first sight, it is unclear
how to accomplish the two tasks simultaneously: completing unobserved
entries, and identifying corrupted columns.\\

This paper studies this precise problem. We do so by exploiting the
algebraic structure of the problem: the non-corrupted columns form
a low-rank matrix, while the corrupted columns can be seen as a column-sparse
matrix. Thus, the mathematical problem we address is to decompose
a low-rank matrix from a column-sparse matrix, based on only partial
observation. Specifically, suppose we are given partial observation of a
matrix $M$, which can be written as
\begin{equation}
M=L_{0}+C_{0},\label{eq:setup}
\end{equation}
where $L_{0}$ is low-rank and $C_{0}$ has only a few non-zero columns.
Here the entries of $C_{0}$ may have arbitrary magnitude and can even be adversarially
built; the column/row space of $L_{0}$ as well as the positions of
non-zero columns of $C_{0}$ are unknown. With a subset of the entries
of $M$ observed, can we efficiently recover $L_{0}$ on the non-corrupted
columns, and also identify the non-zero columns of~$C_{0}$? And,
how do the rank and the number of corrupted columns impact the number
of observations needed?

We provide an affirmative answer to the first question, and a quantitative
solution to the second. In particular, we develop an efficient algorithm,
which is based on a trimming procedure followed by a convex
program that minimizes the nuclear norm and the $\ell_{1,2}$ norm.
We provide sufficient conditions under which this algorithm provably
recovers $L_{0}$ and identifies the corrupted columns. Our algorithm succeeds even when a vanishing fraction of randomly located
entries are observed and a significant fraction of the columns are
corrupted; moreover, the number of observations we need
depends near optimally, in an information-theoretic sense, on the rank of $L_{0}$ and the number of corrupted columns.  Significantly,
we do not assume anything about the values nor the locations of observations
on the corrupted columns. 

We note that our corruption model is very general. By making no assumption
on the corrupted columns, our results cover, but are not limited to, adversarial
manipulation. For example, the corrupted columns can also represent
persistent noise and abnormal sub-populations that are not well modeled
by a (known) probabilistic model. We discuss several such examples in Section~\ref{sec:app}.

Conceptually, our results establish the relation and tradeoffs between
three aspects of the problem: sample complexity (the number
of observed entries), model complexity (the rank of the matrix) and
adversary robustness (the number of arbitrarily corrupted columns). While
the interplay between sample and model complexities is a recurring
theme in modern work of statistics and machine learning, their relation
with robustness (particularly to arbitrary and adversarial corruption,
as opposed to neutral, stochastic noise) seems much less well understood.
Our results show that with more samples, one can not only estimate matrices of higher rank, but also be robust to more adversarial columns. Importantly,
we provide both (and nearly matching) upper and lower bounds, thus establishing a complete and sharp characterization of this phenomenon. To establish lower bounds under arbitrary corruption, we use techniques that are quite different from existing ones for stochastic corruption that largely rely on
Fano's inequality and the alike.

\paragraph{Paper Organization}

We postpone the discussion of related work to Section~\ref{sec:main}
after we state our main theorems. In Section~\ref{sec:app} we describe
several application motivating our study, followed by a summary of our
main technical contributions in Section~\ref{sec:contrib}. In Section~\ref{sec:setup}
we give the mathematical setup of the robust matrix completion problem
with corrupted columns. In Section~\ref{sec:main} we provide the
main results of the paper: a robust matrix completion algorithm, a
sufficient condition for the success of the algorithm, and a matching
inverse theorem showing the optimality of the algorithm. We also survey
relevant work in the literature and discuss their connection to our
results. In Section~\ref{sec:expt} we discuss implementation issues
and provide empirical results. We prove the two main theorems in Sections~\ref{sec:proof_main}
and \ref{sec:proof_inverse}, respectively, with some of the technical
details deferred to the appendix. The paper concludes with a discussion
in Section~\ref{sec:conclusion}.

\subsection{Motivating Applications\label{sec:app}}

Our investigation is motivated by several important problems in machine learning
and statistics, which we discuss below.

\paragraph{Manipulation-Robust Collaborative Filtering}

In online commerce and advertisement, companies collect user ratings
for products, and would like to predict user preferences based on these
incomplete ratings---a problem known as collaborative filtering~(CF). There is a large and growing literature on CF; most well-known
is the work on the Netflix prize~\cite{bennett2007netflix}, but
also see~\cite{adomavicius2005nextGenRecommender,schafer2001ecommerce}
and the references therein. Various CF algorithms have been developed~\cite{herlocker1999algorithmic,linden2003amazon,motwani2007tracingpath,moon2008autologistic,sandvig2007associationRule}.
A typical approach to cast it as a matrix completion problem:
the preferences across users are known to be correlated and thus modeled
as a low-rank matrix $L_{0}$, and the goal is to estimate $L_{0}$
from its partially observed entries. However, the quality of prediction
may be seriously hampered by (even a small number of) \emph{manipulators}---potentially
malicious users, who calibrate (possibly in a coordinated way) their
ratings \emph{and the entries they choose to rank} in an attempt
to skew predictions~\cite{van2010manipulation,lam2004shilling}.
In the matrix completion framework, this corresponds to the setting
where some of the columns of the observed matrix are provided by manipulative
users. As the ratings of the \emph{authentic} users correspond to
a low-rank matrix $L_{0}$, the corrupted ratings correspond to a
column-sparse matrix $C_{0}$. Therefore, in order to perform collaborative
filtering with robustness to manipulation, we need to identify the
non-zero columns of $C_{0}$ and at the same time recover $L_{0}$,
given only a set of incomplete entries. This falls precisely into
the scope of our problem.

\paragraph{Robust PCA}

In the robust Principal Component Analysis (PCA) problem~\cite{xu2013hrpca,Xu2010RobustPCA,lerman2012robustpcaneedle,mccoy2011two},
one is given a data matrix $ M $, of which most of the columns correspond
to authentic data points that lie in a low-dimensional space---the
space of principal components. The remaining columns are \emph{outliers},
which are not (known to be) captured by a low-dimensional linear model.
The goal is to negate the effect of outliers and recover the true
+principal components. In many situations such as problems in medical research
(see e.g.,~\cite{cesa2010partial}), there are unobserved variables/attributes
for each data point. The problem of robust PCA with partial observation---recovering
the principal components in the face of partially observed samples
and also corrupted points---falls directly into our framework.

\paragraph{Crowdsourcing}

Crowdsourcing has emerged as a popular approach for using human power
to solve learning problems. Here multiple-choice questions are distributed to several \emph{workers}, whose answers are then collected and aggregated in an attempt to obtain an accurate answer to each question. In a simplified setting called the \emph{Hammer-Spammer model}~\cite{Karger2011lowrank,Karger2011iterative},
a worker is either a \emph{hammer} who gives correct answers, or a
\emph{spammer} who answers completely randomly. A more general setting
is considered in~\cite{Karger2013multi}, where the spammers need
not follow a probabilistic model and may submit any answers they want,
for instance with an unknown bias, or even adversarially. This problem
can be mapped to our matrix framework, where rows correspond to questions and columns to workers, with $L_{0}$ representing the matrix of true answers from the hammers and $C_{0}$ the answers from the spammers. Each worker typically answers only a subset of the questions, leading to partial observation.

\paragraph{Model Mismatch}

More generally, the corrupted columns can encompass any observations
that are not captured by the assumed low-rank model. These observations
may be generated from an unknown population or affected by factors
beyond the knowledge of the modeler, but not necessarily adversarial.
Such mismatch between the models and data is ubiquitous. For instance,
in collaborative filtering there may exist a small set of atypical
users whose preferences are very weakly correlated with the majority
and thus difficult to infer using data from the majority. In PCA some
data points may simply not conform to the low-dimensional linear model.
The answers from some workers in crowdsourcing systems may be erroneous
as the data collecting process is non-ideal and not fully controllable.
It is difficult to accurately model or recover these columns, but our results guarantee that they do not hinder the recovery of the other columns.

\subsection{Main Contributions\label{sec:contrib}}

In this paper, we propose a new algorithm for matrix completion in
the presence of corrupted columns and provide performance guarantees.
Specifically, we have the following results:
\begin{enumerate}
\item We develop a two-step matrix completion algorithm, which first trims
the over-sampled columns of the matrix, and then solves a convex optimization
problem involving the nuclear norm and the $\ell_{1,2}$ norm. Our
algorithm extends the standard nuclear norm minimization approach
for matrix completion, and the use of trimming and the $\ell_{1,2}$
norm plays a crucial role in achieving robustness to arbitrary column-wise corruption.
\item For an $n\times n$ incoherent matrix with rank $r$ and a subset
of its columns arbitrarily corrupted, we show that if a fraction of
$p$ randomly located entries are observed in the uncorrupted columns,
then our algorithm provably identifies the corrupted columns and completes
the uncorrupted ones as long as $p$ obeys the usual condition $p\gtrsim\frac{r\log^{2}n}{n}$ for matrix
completion, and in addition the fraction
$\gamma$ of corrupted columns satisfies $\gamma\lesssim\frac{p}{r\sqrt{r}\log^{3}n}$. 
\item We further show that the two conditions are near-optimal, in the sense
that if $p\lesssim\frac{r\log n}{n}$ \emph{or} $\gamma\gtrsim\frac{p}{r}$,
then an adversary can corrupt the columns in such a way that \emph{all}
algorithms fail with probability bounded away from zero. Therefore,
our results establish tight bounds for the sample-robustness-rank
tradeoffs in matrix completion.
\item We develop a variant of the Augmented Lagrangian Multipliers (ALM)
method for solving the convex optimization problem in our algorithm. Empirical
results on synthetic data are provided, which corroborate with our
theoretical findings and show that our algorithm is more robust
than standard matrix completion algorithms.
\end{enumerate}

\section{Problem Setup\label{sec:setup}}

Suppose $M$ is a ground-truth matrix in $\RR^{m \times (n+n_c)} $. Among the
$n+n_{c}$ columns of $ M $,  $n$ of them (we will call them \emph{authentic} or \emph{non-corrupted}) span an $r$-dimensional subspace of $\RR^{m}$, and the remaining~$n_{c}$ columns are arbitrary (we will call them \emph{corrupted}). We only observe a subset of the entries of the matrix~$M$, and the goal is to infer the true subspace of the authentic columns and the identities of the corrupted ones.

Under the above setup, it is clear that the matrix $M$ can be
decomposed as $M=L_{0}+C_{0}.$ Here~$L_{0}  \in \RR^{m \times (n+n_c)}$ is the matrix containing the authentic columns, and therefore $\text{rank}(L_{0})=r$. The matrix~$C_{0} \in \RR^{m \times (n+n_c)}$ contains the corrupted columns, so at most $n_{c}$
of the columns of $C_{0}$ are non-zero. Let $I_{0} \subset [n+n_c]$
be the indices of the corrupted columns; that is, 
$ I_{0}:=\text{column-support}(C_{0}),  $
where $\left|I_{0}\right|=n_{c}$.
Let $\Omega\subseteq[m]\times[n+n_{c}]$
be the set of indices of the observed entries of $ M $, and $\P_{\Omega}$
the projection onto the matrices supported on $\Omega$, which is given by
\[
\left(\P_{\Omega}X\right)_{ij}=\begin{cases}
X_{ij}, & \quad(i,j)\in\Omega,\\
0, & \quad(i,j)\notin\Omega.
\end{cases}
\]
With this notation, our goal is to exactly recover from $\P_{\Omega}M$ the authentic columns 
in $L_{0}$ and the corresponding column space as well as the locations
$I_{0}$ of the non-zero columns of $C_{0}$.

\subsection{Assumptions}

In general, it is not always possible to complete a low-rank matrix
in the presence of corrupted columns. For example, if~$L_{0}$ has
only one non-zero column, it is impossible to distinguish $L_{0}$
from~$C_{0}$ even when $ M $ is fully observed. It is also well-known in the matrix completion literature~\cite{candes2009exact,gross2009anybasis,keshavan2009matrixafew}
that if~$L_{0}$ has only one non-zero row, or if one row or column of $L_{0}$ is completely unobserved, then asking to recover~$L_{0}$ from partial observations is problematic. To avoid these pathological situations, we will assume
that~$L_{0}$ satisfy the now standard incoherence condition~\cite{candes2009exact}
and the observed entries on the authentic columns of $L_{0}$ are
sampled at random. We note that we make no assumptions on the values
or locations of the observed entries of corrupted columns in $C_{0}$.

\subsubsection{Incoherence Condition}

Suppose $ L_0 $ has the Singular Value Decomposition (SVD) $L_{0}=U_{0}\Sigma_{0}V_{0}^{\top}$, where $ U_0 \in \RR^{m\times r} $, $ V_0 \in \RR^{(n+n_c)\times r} $ and $ \Sigma_0 \in \RR^{r\times r} $.
We use $\left\Vert \cdot\right\Vert _{2}$ to denote the vector~$\ell_{2}$
norm, and $e_{i}$ be the $i$-th standard basis vector whose dimension will
be clear in context.
\begin{assumption}
[Incoherence]\label{asm:incoherence}
The matrix $L_{0}$ is zero
on the columns in $\I_{0}$. Moreover, $L_{0}$ satisfies
the following two incoherence conditions with parameter $\mu$: 
\begin{align*}
\max_{1\le i\le m}\left\Vert U_{0}^{\top}e_{i}\right\Vert _{2}^{2} & \le\mu\frac{r}{m},\\
\max_{1\le j\le n+n_{c}}\left\Vert V_{0}^{\top}e_{j}\right\Vert _{2}^{2} & \le\mu\frac{r}{n}.
\end{align*}

\end{assumption}
Since the columns of $L_{0}$ in $\I_{0}$ are superposed with the
arbitrary $C_{0}$, there is no hope of recovering these columns.
Therefore, there is no loss of generality to assume $L_{0}$ is zero
on $\I_{0}$. Consequently, the matrix $V_{0}^{\top}$ has at most $n$ non-zero
columns (all in $I_{0}^{c}$), and accordingly the denominator on
the right hand side of the second inequality above is $n$ instead
of the full dimension $n+n_{c}$.

The two incoherence conditions are needed for completion of $ L_0 $ from partial observations {even with no corrupted columns}. 
The incoherence parameter $\mu$ is known to be small in various natural models and applications~\cite{candes2009exact,candes2010NearOptimalMC}.
The second inequality in Assumption~\ref{asm:incoherence}
is necessary in the presence of corrupted columns, \emph{even when
the matrix is fully observed}. This inequality essentially enforces
that the information about the column space of $L_{0}$ is spread
out among the columns. If, for instance, an authentic column of $L_{0}$
were not in the span of all the other columns, one could not hope
to distinguish it from a corrupted column (cf.~\cite{Xu2010RobustPCA}).

Finally, we note that previous work on matrix completion often imposes
a \emph{strong incoherence condition} $\max_{i,j}\left|\left(U_{0}V_{0}^{\top}\right)_{ij}\right|\le\sqrt{\frac{\mu r}{mn}}$~\cite{candes2009exact,candes2010NearOptimalMC,gross2009anybasis,Recht2009SimplerMC}.
We do not need this assumption, thus improving over these previous
results. Further discussion on this point is provided after our main theorems.

\subsubsection{Sampling Model}

Recall that $\I_{0}$ is the indices of
the corrupted columns. Let $\tilde{\Omega} := \Omega \cap \left( [m]\times\I_{0}^{c} \right)$
be the set of indices of observed entries on the non-corrupted columns.
We use the following definition.
\begin{defn}
[Bernoulli model]
\label{def:bernoulli}
Suppose $\Theta_{0} \subseteq [m]\times[n+n_{c}]$. A set $\Theta$ is said to be sampled
from the \emph{Bernoulli model with probabilities $\left\{ p_{j}\right\}_{j=1}^{n+n_c}$}
\emph{on $\Theta_{0}$} if each element $(i,j)$ of $\Theta_{0}$
is contained in $\Theta$ with probability $p_{j}$, independently
of all others. If $p_{j}=p$ for $j$, then $\Theta$ is said to be
sampled from the Bernoulli model on $ \Theta_0 $ with \emph{uniform} probability $p$.
\end{defn}
We can now specify our assumption on how the observed entries are
sampled.
\begin{assumption}
[Sampling]\label{asm:sampling}The set $\tilde{\Omega}$ is sampled
from the Bernoulli model with probabilities $\{p_{j}\}$ on $[m]\times I_{0}^{c}$,
where $p_{j}\ge p$ for all $j\in I_{0}^{c}$. Moreover, $\tilde{\Omega}$
is independent of $ \P_\Omega C_0 $, the observed entries on the corrupted columns. 
\end{assumption}
Note that our model is more general than the uniform sampling model
assumed in some previous work---we only require a \emph{lower bound
}on the observation probabilities of the non-corrupted columns, so some
columns may have an observation probability higher than
$p$. Importantly, the Bernoulli model is \emph{not} imposed on the corrupted columns. The adversary may choose
to reveal all entries on columns in $\I_{0}$ or just a fraction
of them, and the locations of these observed entries may be chosen
randomly or adversarially depending on $L_{0}$. The assumption of $\tilde{\Omega}$
being independent of the corrupted columns is needed for technical reasons.
We conjecture that it is only an artifact of our analysis and not
actually necessary, as indicated by our empirical results.

\subsubsection{Corrupted Columns}

Let $\gamma:=\frac{n_{c}}{n}$ be the ratio of the number of corrupted
columns to the number of authentic columns. Other than the independence
requirement above, we make no assumption whatsoever on the corrupted
columns in $ C_0 $. The incoherence assumption is imposed on the authentic $L_{0}$, not on~$M$ or~$C_{0}$, as is the sampling
assumption, and therefore the corrupted columns are not restricted in
any way by these. These columns need not follow any probabilistic
distributions, and they may be chosen by some adversary who aims to
skew one's inference of the non-corrupted columns. One consequence
of this is that we will not be able to recover the \emph{values} of the completely
corrupted columns of $C_{0}$, but we are able to reveal their \emph{identities}.

\section{Main Results: Algorithms, Guarantees and Limits\label{sec:main}}

The main result of this paper says that despite the corrupted columns and partial observation, we can simultaneously recover $L_{0}$, the non-corrupted columns, and identify $\I_{0}$, the position of the corrupted columns, as long as the number of corrupted columns and unobserved entries are controlled. Moreover, this can be achieved efficiently via a \emph{tractable} procedure,
given as Algorithm~\ref{alg:MP}.

\begin{algorithm}[b]
\protect\caption{Manipulator Pursuit}
\label{alg:MP}

\textbf{Input}: $\P_{\Omega}(M)$,$\Omega$,$\lambda$, $\rho$.

\textbf{Trimming}: For $j=1,\ldots,n+n_{c}$, if the number of observed
entries $h_{j}$ on the $j$-th column satisfies $h_{j}>\rho m$,
then randomly select $\rho m$ entries (by sampling without replacement)
from these $h_{j}$ entries and set the rest as unobserved. Let $\hat{\Omega}$
be the set of remaining observed indices. 

\textbf{Solve}~for~optimum~$(L^{\ast}$,~$C^{\ast}$): 
\begin{eqnarray}
\textrm{minimize}_{L,C} & \quad & \left\Vert L\right\Vert _{\ast}+\lambda\left\Vert C\right\Vert _{1,2}\label{eq:L12formulation}\\
\textrm{subject to} & \quad & \P_{\hat{\Omega}}(L+C)=\P_{\hat{\Omega}}(M)\nonumber 
\end{eqnarray}

\textbf{Set~}$\I^{*}=\text{column-support}\left(C^{*}\right):=\{j:\; C_{ij}^{\ast}\neq0\text{ for some }i\}$.

\textbf{Output}:~$L^{*}$, $ C^* $ and~$\I^{*}$. 
\end{algorithm}

The algorithm has two steps. In the first \emph{trimming} step, we find columns
with a large number of observed entries, and throw away some of these
entries randomly. This step is important, both in theory and empirically,
to achieve good performance: an adversary may choose
to reveal (and corrupt) a large number of entries on certain columns, which may
skew the next step of the algorithm; the trimming step protects against
this effect. Note that we cannot directly identify these over-sampled
corrupted columns by counting the number of observations---under the
(non-uniform) sampling model in Assumption~\ref{asm:sampling}, some
authentic columns are also allowed to have many observed entries. 

In the next step of the algorithm, we solve a convex program with
the trimmed observations as the input. The convex program, in fact
a Semidefinite Program (SDP), finds a pair $(L^*,C^*)$ that is consistent with
the observations and minimizes the weighted sum of the nuclear norm
$\left\Vert L\right\Vert_{*}$ and the matrix $\ell_{1,2}$
norm $\left\Vert C\right\Vert _{1,2}$, where $ \left\Vert L\right\Vert_{*} $ is the sum of singular values
of $L$ and a convex surrogate of its rank, and $\left\Vert C\right\Vert _{1,2}$ is the sum of the column $\ell_{2}$ norms of $C$ and a convex surrogate of its column sparsity.
The algorithm has two parameters: the threshold $0<\rho<1$ for trimming and
the coefficient $\lambda>0$ for the weighted sum in the convex program. Our theoretical results specify how to choose their values.

We say Algorithm~\ref{alg:MP} \emph{succeeds} if we have $\P_{U_{0}}(L^{*})=L^{*}$, $\P_{\I_{0}^{c}}(L^{*})=L_{0}$
 and $\I^{*}\subseteq\I_{0}$
for \emph{any} optimal solution $(L^{*},C^{*})$ of~\eqref{eq:L12formulation}, where $ \P_{U_{0}}(L^{*}) := U_0 U_0^\top L^* $ is the projection of the columns of $ L^* $ onto the column space of $ L_0 $, and $ \P_{\I_{0}^{c}}(L^{*}) $ is the projection of $ L^* $ onto the matrices supported on the column indices in $ I_0^c $, given by
\begin{align*}
\left[ \P_{\I_{0}^{c}}(L^{*}) \right]_{ij} = 
  \begin{cases}
  L^*_{ij}, & \text{if } j\not\in I_0,\\
  0, & \text{if } j\in I_0. 
  \end{cases}
\end{align*} 
That is, the algorithm succeeds we recover the true
column space of the original $L_{0}$ and complete its uncorrupted
columns, and at the same time identify the locations of the corrupted
columns. Note that the definition of success allows for $I^{*}\subsetneq I_0$.
In this case it may appear that some corrupted columns are unidentified
and included in $L^{*}$, but it is actually not a problem: the requirement
$\P_{U_{0}}(L^{*})=L^{*}$ means that these unidentified ``corrupted''
columns can be completed to lie in the true column space of $L_{0}$,
so they are essentially \emph{not corrupted}, as they are indistinguishable from a partially observed authentic column and do not affect the completion.

\subsection{Sufficient Conditions for Recovery\label{sec:Sufficient-Conditions-for}}

Our first main theorem guarantees that under some natural conditions,
our algorithm exactly recovers the non-corrupted columns and the identities
of the corrupted columns with high probability. Recall that $p$ is
a lower bound of the observation probability on the non-corrupted
columns, $\gamma:=\frac{n_{c}}{n}$ the ratio between the numbers
of corrupted and uncorrupted columns, and $ \rho $ the trimming threshold.
\begin{thm}
\label{thm:main}Let $\alpha:=\frac{\rho}{p}$. There exist universal
positive constant $c_{1}$ and $c_{2}$ for which the following holds. Suppose
the Assumptions~\ref{asm:incoherence} and~\ref{asm:sampling} hold.
If in Algorithm~\ref{alg:MP} we take 
\[
\lambda\in\left[\sqrt{\left(1+\frac{1}{\alpha}\right)\frac{\mu r\log(m+n)}{pn}},\frac{1}{48\sqrt{\sqrt{(1+\alpha)\mu r}\gamma n\log(m+n)}}\right],
\]
and $(p,\gamma)$ satisfies
\begin{align}
p & \ge c_{1}\left(1+\frac{1}{\alpha}\right)\frac{\mu r\log^{2}(m+n)}{\min(m,n)},\label{eq:p_cond}\\
\gamma & \le c_{2}\frac{\alpha}{1+\alpha\sqrt{\alpha}}\frac{p}{\mu r\sqrt{\mu r}\log^{3}(m+n)},\label{eq:gamma_cond}
\end{align}
then Algorithm~\ref{alg:MP} succeeds with probability at least $1-20(m+n)^{-5}$.
Note that the interval for $\lambda$ is non-empty under the condition~(\ref{eq:gamma_cond}).
\end{thm}
We prove this theorem in Section~\ref{sec:proof_main}.

The two conditions~(\ref{eq:p_cond}) and~(\ref{eq:gamma_cond}) have the natural interpretation that the
algorithm succeeds as long as there is sufficiently many observed
entries (in particular, more than the degrees of freedom of a rank-$r$
matrix), and the number of corrupted columns is not too large relative
to the number of observed entries. We discuss these two conditions
in more details in the next sub-section. The theorem also shows that the parameter $\lambda$ in the convex program~(\ref{eq:L12formulation}) can take any value in a certain range.

\subsubsection{Consequences}

We explore several consequences of Theorem~\ref{thm:main}. The conditions~(\ref{eq:p_cond}) and~(\ref{eq:gamma_cond}) above
involve the value of the parameter $\rho$ from trimming in Algorithm~\ref{alg:MP}.
The conditions become the least restrictive if $\alpha:=\frac{\rho}{p}=\Theta(1)$,
i.e., when $\rho$ is of the same order of $p$. Choosing $\rho$ optimally
in this way gives the following corollary.
\begin{cor}
[Optimal Bound]\label{cor:optimal_rho}There exist universal constant
$c_{1}$ and $c_{2}$ such that the following holds. Suppose the Assumptions~\ref{asm:incoherence} and~\ref{asm:sampling}
hold, and  we take $\rho=p$ and $\lambda=\sqrt{\frac{2\mu r\log(m+n)}{pn}}$
in Algorithm~\ref{alg:MP}.  Algorithm~\ref{alg:MP}
succeeds with probability at least $1-20(m+n)^{-5}$ as long as $(p,\gamma)$ satisfy~(\ref{eq:p_cond})
and~(\ref{eq:gamma_cond}) with $\alpha=1$. 
\end{cor}

For a more concrete example, suppose the observation probability satisfies $ p \gtrsim \frac{\sqrt{\mu^3 r^3} \log^3 n}{n^{1-\kappa}} $, then Corollary~\ref{cor:optimal_rho} guarantees success of our algorithms when the number of corrupted entries $ \gamma n $ is less than $n^\kappa $.

In a conference version~\cite{chen2011rmc_icml} of this paper, we
analyze the second step of Algorithm~\ref{alg:MP} (i.e., without
trimming, or equivalently $ \rho = 1 $) and show that it succeeds if $ (p,\gamma) $ satisfy (among other things) the condition 
\[
\gamma\lesssim\frac{p^{2}}{\left(1+\frac{\mu r}{p\sqrt{n}}\right)^{2}\mu^{3}r^{3}\log^{6}(m+n)}.
\]
This result is significantly improved by Corollary~\ref{cor:optimal_rho}
(in particular, compared to the condition~(\ref{eq:gamma_cond}) with $\alpha=1$),
which allows for an order-wise larger number of corrupted columns.
Our analysis reveals that the trimming step in Algorithm~\ref{alg:The-ALM-Algorithm}
is crucial to this improvement.
\begin{rem}
In practice, we may estimate the value of $p$ by using a robust mean
estimator (e.g., the median or trimmed mean) of the fraction of observed
entries over the columns. Given such an estimate $ \hat{p} $, we can set $\rho=\hat{p}$
and $\lambda=\sqrt{\frac{c\log(m+n)}{\hat{p}n}}$ for some constant $c$ (say $ 50 $),
and the algorithm's success is guaranteed by Theorem~\ref{thm:main}
and Corollary~\ref{cor:optimal_rho} for $\mu r=O(1)$. (Note
that while we may not know $n$, we do know the value of $n+n_{c}$,
which differs from $n$ by at most a factor of $2$ whenever $n_{c}\le n$.)
This approach is taken in our empirical studies in Section~\ref{sec:expt}.\\
\end{rem}

Setting $p=1$ in Corollary~\ref{cor:optimal_rho} immediately yields
a guarantee for the full observation setting.
\begin{cor}
[Full Observation]\label{cor:full_obs}Suppose the Assumptions~\ref{asm:incoherence} and~\ref{asm:sampling} hold with $ p=1 $. If we take $\rho=1$ and $\lambda=\sqrt{\frac{2\mu r\log(m+n)}{n}}$
in Algorithm~\ref{alg:MP}, and $\gamma:=\frac{n_{c}}{n}$ satisfies
\begin{align*}
\gamma & \le c_{1}'\frac{1}{\mu r\sqrt{\mu r}\log^{3}(m+n)}
\end{align*}
for some universal constant $c_{1}'$, then Algorithm~\ref{alg:MP}
succeeds with probability at least $1-20(m+n)^{-5}$.
\end{cor}
The full observation setting of our model corresponds to the Robust
PCA problem with \emph{sample-wise} corruption (cf. Section~\ref{sec:app}), which is previously
considered in~\cite{Xu2010NIPS_RobustPCA,Xu2010RobustPCA}. There
they propose an algorithm called \emph{Outlier Pursuit}, which is
similar to the second step of our Algorithm\ref{alg:MP} and shown to succeed in the full observation setting if $\gamma\lesssim\frac{1}{\mu r}$.
Our result in Corollary~\ref{cor:full_obs} is off by a small
factor of $\sqrt{\mu r}\log^{3}(m+n)$. This sub-optimality can be removed
by a more careful analysis in the setting with $ p $ close to $ 1 $, but we choose not to delve into it.\\

On the other hand, setting $\gamma=0$ gives a guarantee for the standard exact
matrix completion setting with clean observations. Our result is powerful
enough that it in fact improves upon some previous results in this setting.
\begin{cor}
[Matrix Completion]\label{cor:mc}Suppose $\gamma=0$ and the Assumption~\ref{asm:incoherence} and~\ref{asm:sampling}
hold. If we take $\rho=1$ and $\lambda\ge\sqrt{\frac{2\mu r\log(m+n)}{pn}}$
in Algorithm~\ref{alg:MP}, and $p$ satisfies 
\[
p\ge c_{1}''\frac{\mu r\log^{2}(m+n)}{\min(m,n)}
\]
for some universal constant $c_{1}''$, then Algorithm~\ref{alg:MP}
succeeds with probability at least $1-20(m+n)^{-5}$.
\end{cor}
Exact matrix completion is considered in the seminal
work~\cite{candes2009exact} and subsequently in~\cite{candes2010NearOptimalMC,gross2009anybasis,Recht2009SimplerMC},
in which the low-rank matrix $L_{0}$ is assumed to satisfied two incoherence
conditions: the standard incoherence condition with parameter $\mu$ as in Assumption~\ref{asm:incoherence}, and an additional \emph{strong incoherence condition} $\left\Vert UV\right\Vert _{\infty}\le\sqrt{\frac{\mu_{\text{str}}r}{mn}}$.
They show that $ L_0 $ can be exactly recovered via
nuclear norm minimization if $p\gtrsim\frac{\max\left\{ \mu,\mu_{\text{str}}\right\} r\log^{2}(m+n)}{\min\left\{ m,n\right\} }$.
Corollary~\ref{cor:mc} improves upon this result by removing the dependence
on the strong incoherence parameter $\mu_{\text{str}}$, which can
be as large as $\mu r$. This improvement was also observed in the recent work in~\cite{chen2013incoherence_arxiv,chen2014coherent_icml}. \\

We have seen that Theorem~\ref{thm:main} and Corollary~\ref{cor:optimal_rho}
give, as immediate corollaries, strong bounds for the special cases of full observation and standard matrix completion, which is a testament to the sharpness of our results. In fact, we show in the next sub-section that the conditions in Theorem~\ref{thm:main} are near-optimal.

\subsection{Information-Theoretic Limits for Recovery\label{sec:Information-Theoretic-Limits-for}}

Corollary~\ref{cor:optimal_rho} says that the conditions (\ref{eq:p_cond})
and (\ref{eq:gamma_cond}) with $\alpha=1$ are sufficient for our
algorithm to succeed. Theorem~\ref{thm:converse} below shows that these conditions
are in fact close to being information-theoretic (minimax) optimal. That is, they cannot
be significantly improved by any algorithm regardless of its computational
complexity. Note that the theorem tracks the values of $\mu$, $r$,
$p$ and $\gamma$, so all of them can scale in a non-trivial way
with respect to $n$.
\begin{thm}
\label{thm:converse}Suppose $m=n\ge4$, $\mu r\le\frac{n}{\log(2n)}$, and $ (p,\gamma) $ satisfy
\begin{align}
p & \le\frac{\mu r\log(2n)}{2n}\label{eq:p_necessary}\\
\textbf{ or}\quad\gamma & :=\frac{n_{c}}{n}\ge\frac{2p}{\mu r}.\label{eq:gamma_necessary}
\end{align}
Then any algorithm will fail to output the correct column space with probability
at least $\frac{1}{16}$; more precisely, for all measurable functions $\hat{L} $ of $ M $ and $ \Omega $,
\begin{align*}
\max_{L_0, C_0, \Omega \backslash \Omegat} \mathbb{P}\left[ \P_{U_0} (\hat{L}) \neq \hat{L} \right] 
\ge \frac{1}{16},
\end{align*}
where the maximization ranges over all matrix pairs $(L_0, C_0) $ and observed indices on the corrupted columns $ \Omega \backslash \Omegat $ that satisfy the Assumptions~\ref{asm:incoherence} and~\ref{asm:sampling}, and the probability is with respect to the distribution of the observed indices on the non-corrupted columns $ \Omegat $.
\end{thm}
We prove this theorem in Section~\ref{sec:proof_inverse}.

By comparing with Theorem~\ref{thm:converse}, we see that the conditions in Corollary~\ref{cor:optimal_rho} are close to the achievable limits. In particular, with $\alpha=1$, the condition~(\ref{eq:p_cond})
on $p$ matches~(\ref{eq:p_necessary}) up to one logarithmic factor,
and the condition~(\ref{eq:gamma_cond}) on $\gamma$ is worse than~(\ref{eq:gamma_necessary})
by a factor of $c\sqrt{\mu r}\log^{3}n$. In particular, both conditions
are optimal up to logarithmic factors in the case of constant rank
and incoherence $\mu r=O(1)$. It is of interest to study whether this small gap can be closed, potentially by tightening up the sufficient conditions in Theorem~\ref{thm:main} and Corollary~\ref{cor:optimal_rho}.

The failure condition~(\ref{eq:p_necessary}) is an extension
of a standard result for matrix completion in~\cite[Theorem 1.7]{candes2010NearOptimalMC}.
To gain some intuition on the second condition~(\ref{eq:gamma_necessary}),
we consider the case with $\mu r=1$, for which the condition becomes $n_{c}\gtrsim pn$.
This means that with probability bounded away from zero, the number of observed corrupted entries in the first row exceeds that of observed authentic entries in the same row. In this
case, if the corrupted entries in the other rows are chosen to be consistent with the true column space (on all but the first coordinates), then no algorithm can tell which of the two sets
of entries in the first row is actually authentic, and therefore recovery
of this row is impossible. Theorem~\ref{thm:converse} is proved
using an extension of the above argument---by demonstrating a particular
way of corrupting $n_{c}\gtrsim\frac{pn}{\mu r}$ columns that provably
confuses any algorithm.

\paragraph*{Implications for Robust PCA:}

Recall the Robust PCA setting with full observations ($p=1$) and the  {Outlier Pursuit} algorithm
discussed after Corollary~\ref{cor:full_obs} in Section~\ref{sec:Sufficient-Conditions-for}.  Theorem~\ref{thm:converse}
shows that $\gamma\gtrsim\frac{1}{\mu r}$ is necessary, so the guarantee for Outlier Pursuit given in~\cite{Xu2010RobustPCA} is order-wise optimal.

\subsection{Sample-Robustness-Rank Tradeoffs}

The results in the last two-subsections highlight the tradeoffs between sample complexity,
outlier robustness and model complexity~(matrix rank). In particular,
given a higher the observation probability $p$,  one can handle a
higher fraction $\gamma$ of corrupted columns and a higher rank $r$
of the underlying matrix. The other direction is also true:
with a smaller $p$, the fraction of allowable corrupted columns
and the allowable rank will necessarily become smaller, regardless
of the algorithm and the amount of computational. Theorems~\ref{thm:main}
and~\ref{thm:converse} provide the precise conditions that $p$, $\gamma$
and $r$ need to obey.

We emphasize that here we consider robustness to \emph{arbitrary and
possibly adversarial} corruption. Our results characterize, in terms of both upper and lower bounds, the tradeoffs between adversary robustness and
sample/model complexities. This can be put into the context of the study of modern \emph{high-dimensional statistics}~\cite{negahban2012unified_ss,buhlmann2011high},
where the relationship between sample and model complexities is a
central topic of interest. More recently, a line of work has focused
on the tradeoffs between computational complexity and various statistical
quantities~\cite{chandrasekaran2013tradeoff,berthet2013lowerSparsePCA,ma2013submatrix,zhang2014polynomial}.
Our results can be viewed as adding a new dimension to these recent lines of work: we
consider another axis of the problem---robustness (to adversarial
corruption)---and its relation to other statistical quantities.
Therefore, while we investigate a specific problem (matrix completion),
we expect  sample-robustness tradeoffs to be relevant in a broader
context.

Finally, we note that our empirical study in Section~\ref{sec:expt} demonstrate
the following phenomenon: If we further assume that the corrupted columns
are randomly generated and independent of each other, then our algorithm
can recover $L_{0}$ from a much higher number $n_{c}$ of corrupted
columns than is predicted by Theorems~\ref{thm:main} and~\ref{thm:converse}
(which require, among other things, $n_{c}=\gamma n\le1$). In particular,
the corrupted columns can significantly out-number the authentic columns.
This means our algorithm is useful well beyond the adversarial corruption
setting considered in the theorems above, and its actual performance
can become better if the corruption is more restricted and ``benign''.
A similar phenomenon is observed in~\cite{lerman2012robustpcaneedle,lin2014rpca} for the special case of \emph{full observation}.
Here we therefore see another level of sample-robustness-rank tradeoffs: If
we only ask for a weaker sense of robustness, namely, robustness against
randomly corrupted columns as opposed to arbitrary ones, then we have
more relaxed requirements on the observation probability, the
rank  and the number of corrupted columns. It is an interesting open problem to rigorously quantify the interplay between the nature of the corruption and the recovery performance.

\subsection{Connections to Prior Work and Innovation\label{sub:ConnectionsInnovation}}

Recent work in matrix completion shows that by using convex optimization~\cite{candes2009exact,candes2010NearOptimalMC,gross2009anybasis,Recht2009SimplerMC}
or other algorithms~\cite{keshavan2009matrixafew,jain2013altMin,cai2010singular},
one can exactly recover an $n\times n$ rank-$r$ matrix with high
probability from as few as $O(nr\mathrm{poly}\log n)$ (clean) entries.
Our paper extends this line of work and shows that even if all the
observed entries on some columns are completely corrupted (by possibly
adversarial noise), one can still recover the non-corrupted columns
as well as the identity of the corrupted ones. As discussed before,
our work also extends the work in~\cite{Xu2010NIPS_RobustPCA,Xu2010RobustPCA},
which only considers the full observation setting; see also~\cite{agarwal2012decomposition}
for results on the full observation setting with noise. The centerpiece
of our algorithm is a convex optimization problem that is a convex
proxy to a very natural but intractable algorithm for our task, namely,
finding a low-rank matrix $L$ and a column-sparse matrix $C$ consistent
with the observed data. Such convex surrogates for rank and support
functions have been used (often separately) in problems involving
low-rank matrices~\cite{recht2010guaranteed,candes2009exact}) and
in problems with group-sparsity~\cite{yuan2006group,Huang2010group}.
When this manuscript is under preparation, we learn about the very
recent work~\cite{klopp2014RobustMC}, which also studies robust
matrix completion under column-wise sparse corruption, albeit under a somewhat different setting. Their results are focused on the noisy setting with general sampling distributions, but do not guarantee exact recovery in the noiseless case. 

Our work is also related to the problem of separating a low-rank matrix
and an overall (element-wise) sparse matrix from their sum~\cite{candes2009robustPCA,chandrasekaran2011siam}
(this is sometimes called the low-rank-plus-sparse problem, or $L+S$
for short). This problem has also been studied under the partial observation
setting~\cite{candes2009robustPCA,chen2011LSarxiv,li2013constantCorruption}.
Compared to this line of work, our results indicate that separation
is possible even if the low-rank matrix is added with a \emph{column sparse
}matrix instead of an \emph{overall sparse} matrix. In particular,
we allow \emph{all} the observations from some columns to be completely corrupted. In contrast, existing guarantees for the $L+S$ problem require that from each row and column at least \emph{some} observations are clean, thus not suitable for our setting; this is also demonstrated
in our experiments. Moreover, although we do not pursue in this paper, our techniques allow us to establish results on separating three components---a low rank matrix, an element-wise sparse matrix, and a column-sparse matrix.

Besides the obvious difference in the problem setup, our paper also
departs from the previous work in terms of mathematical analysis.
In particular, in previous works in exact matrix completion and decomposition,
the intended outcome is known \textit{a priori}---their goal is to
output a matrix or a pair of matrices, exactly equal to the original
one(s). In our setting, however, the optimal solution of the convex
problem is in general neither the original low rank matrix $L_{0}$
nor the matrix $C_{0}$ which consists of only the corrupted columns.
This critical difference requires a novel analysis that builds on
a variant of the \emph{primal-dual witness} (or \emph{oracle problem})
method. This method has been applied to study support recovery in
problems involving sparsity~\cite{amini2009sparsePCA,lee201selection}.
Here we use the method for the recovery of the \emph{eigen space}
and \emph{column support.} A related problem is considered in~\cite{Xu2010NIPS_RobustPCA,Xu2010RobustPCA},
which, however, only studies with the full observation setting. The
presence of (many) missing entries makes the problem much more
complicated, as we need to deal with three matrix structures simultaneously,
i.e., low-rankness, column sparsity, and overall/element-wise sparsity.
This requires the introduction of new ingredients in the analysis;
in particular, one important technical innovation requires the development of new concentration results that involve these three structures, including bounds on the $\left\Vert \cdot\right\Vert _{\infty,2}$ norms of certain randomly sampled low-rank matrices (see Lemmas~\ref{lem:inf_2} and~\ref{lem:inf_2_order_1}).

\section{Implementation and Empirical Results\label{sec:expt}}

In this section, we discuss implementation issues of our algorithm
and provide empirical results.

\subsection{An ADMM Solver for the Convex Program}

The optimization problem~(\ref{eq:L12formulation}) is a semi-definite
program (SDP), and can be solved by off-the-shelf SDP solvers. However,
these general-purpose solvers can only handle small problems (e.g.,
400-by-400 matrices) and do not scale well to large datasets. Here we use
a family of first order algorithms called the Alternating Direction
Method of Multipliers (ADMM) methods~\cite{Boyd2011ADMM, Lin2009_ALM}, shown to
be effective on problems involving non-smooth objective functions.

We adapt this method to our partially observed, $\left\Vert \cdot\right\Vert _{\ast}+\lambda\left\Vert \cdot\right\Vert _{1,2}$-type
problem; see Algorithm~\ref{alg:The-ALM-Algorithm}. Here $\mathfrak{L}_{\epsilon}(S)$
is the entry-wise soft-thresholding operator: if $\left|S_{ij}\right|\le\epsilon$,
then set it to zero, and otherwise let $S_{ij}:=S_{ij}-\epsilon S_{ij}/\left|S_{ij}\right|$.
Similarly, $\mathfrak{C}_{\epsilon}(C)$ is the column-wise soft-thresholding
operator: if $\left\Vert C_{i}\right\Vert _{2}\le\epsilon$, then
set it to zero, and otherwise let $C_{i}:=C_{i}-\epsilon C_{i}/\left\Vert C_{i}\right\Vert _{2}$.
Note that the matrix $E^{(k)}$ accounts for the unobserved entries.
In our experiments, the parameters are set to $u_{0}=\left(\left\Vert M\right\Vert _{1,2}\right)^{-1}$
and $\alpha=1.1$, and the criterion for convergence is
$
\left\Vert M-E^{(k)}-L^{(k)}-C^{(k)}\right\Vert _{F}/\left\Vert M\right\Vert _{F}\le10^{-6}.
$

The main cost of Algorithm \ref{alg:The-ALM-Algorithm} is computing the SVD of the matrix $Z:=M-E^{(k)}-C^{(k)}+u_{k}^{-1}Y^{(k)}$
in each iteration. We can speed up the computation by taking advantage
of the specific structure of our problem, namely partial
observation and low-rankness. Observe that the iterate $Z$ can be written as the
sum of two matrices $Z=\left(M-E^{(k)}-L^{(k)}-C^{(k)}+u_{k}^{-1}Y^{(k)}\right)+L^{(k)}.$
A careful examination of Algorithm~\ref{alg:The-ALM-Algorithm} reveals
that the first matrix is non-zero only on the observed indices~$\Omega$,
while the second matrix has rank equal to the number of singular values
that remain non-zero after the soft-thresholding in the last iteration.
We can therefore employ a celebrated SVD routine called PROPACK~\cite{propack},
which can make use of such sparse and low-rank structures.
Using this strategy, we are able to apply the algorithm to moderately
large instances in our experiments, especially in the setting we care
most, i.e., when only a small number of entries are observed. 

\begin{algorithm}
\protect\caption{\label{alg:The-ALM-Algorithm} The ALM Algorithm for Robust Matrix
Completion}

\begin{algorithmic}

\STATE \textbf{input}: $\P_{\Omega}M\in\text{\ensuremath{\mathbb{R}}}^{m\times(n+n_{c})}$
(assuming~$\P_{\Omega^{c}}M=0$), $\Omega$, $\lambda$

\STATE \textbf{initialize}:~$Y^{(0)}=0$; $L^{(0)}=0$; $C^{(0)}=0$;
$E^{(0)}=0$; $u_{0}>0$; $\alpha>1$; $k=0$.

\WHILE{not~converged}

\STATE $(U,S,V)=\textrm{SVD}\left(M-E^{(k)}-C^{(k)}+u_{k}^{-1}Y^{(k)}\right)$;

\STATE $L^{(k+1)}=U\mathfrak{L}_{u_{k}^{-1}}\left(S\right)V^{\top}$;

\STATE $C^{(k+1)}=\mathfrak{C}_{\lambda u_{k}^{-1}}\left(M-E^{(k)}-L^{(k+1)}+u_{k}^{-1}Y^{(k)}\right)$;

\STATE $E^{(k+1)}=\P_{\Omega^{c}}\left(M-L^{(k+1)}-C^{(k+1)}+u_{k}^{-1}Y^{(k)}\right)$;

\STATE $Y^{(k+1)}=Y^{(k)}+u_{k}\left(M-E^{(k+1)}-L^{(k+1)}-C^{(k+1)}\right)$;

\STATE $u_{k+1}=\alpha u_{k}$;

\STATE $k\leftarrow k+1$;

\ENDWHILE

\RETURN $\left(L^{(k)},C^{(k)}\right)$

\end{algorithmic} 
\end{algorithm}

\subsection{Simulations}

We test the performance of our method on synthetic data. For a given
rank $r$, we generate two matrices $A\in\mathbb{R}^{m\times r}$
and $B\in\mathbb{R}^{n\times r}$ with i.i.d.\ standard Gaussian
entries, and then build the rank-$r$ matrix $L_{0}\in\mathbb{R}^{m\times(n+n_{c})}$
by $L_{0}=AB^{\top}$ padded with $n_{c}$ zero columns. The set of
observed entries on the authentic columns is generated according to
the Bernoulli model in Assumption~\ref{asm:sampling}. The observation
probabilities $\left\{ p_{j}\right\} $ on the authentic columns,
as well as the $n_{c}$ corrupted columns in $C_{0}$ and their observed
entries, are specified later. The observed matrix $\P_{\Omega}M=\P_{\Omega}\left(L_{0}+C_{0}\right)$
and the set of observed entries $\Omega$ are then given as input
to Algorithm~\ref{alg:MP}, with the convex program solved using
the ADMM solver described above. We set the parameters $\rho$ and
$\lambda$ in Algorithm~\ref{alg:MP} according to Corollary~\ref{cor:optimal_rho},
estimating $p_j$ by $1.1\times\text{median}\left(\left\{ \tilde{p}_{j}\right\} \right)$,
where $\tilde{p}_{j}$ is the empirical observation probability of
the $j$-th columns,

\subsubsection{Effect of Trimming}

\begin{figure}
\begin{centering}
\begin{tabular}{cc}
\includegraphics[trim=150 290 150 310, clip=1, width=0.4\columnwidth]{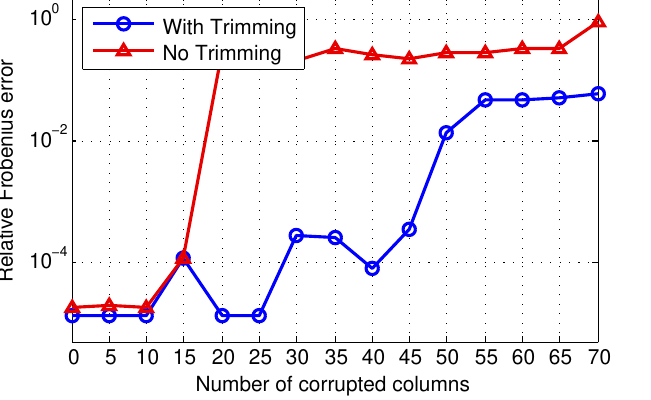} 
& 
\includegraphics[trim=150 290 150 310, clip=1,width=0.4\columnwidth]{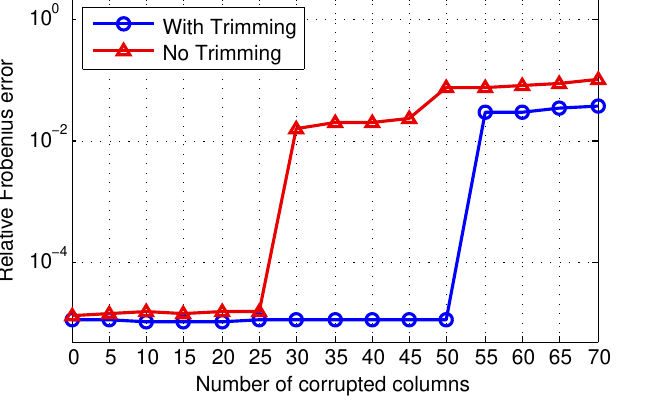}\tabularnewline
(a) & (b)\tabularnewline
\end{tabular}
\par\end{centering}

\protect\caption{\label{fig:trim_vs_no_trim}Comparison of the performance of Algorithm~\ref{alg:MP}
with and without trimming. The plots show the relative Frobenius norm
errors on recovering the uncorrupted columns of a $400\times400$ rank-$2$
matrix with observation probabilities (a) $p=0.2$, (b) $p=0.3$.
Each point in the plots is the average of $10$ trials.}
\end{figure}

In the first set of experiments, we study the performance of Algorithm~\ref{alg:MP}
with and without the trimming step. We consider recovering a matrix
$L_{0}$ with rank $r=2$ and dimensions $m\times(n+n_{c})=400\times400$.
The $n_{c}$ corrupted columns in $C_{0}$ are identical and equal
to a random column vector in $\RR^{m}$, with all of them
fully observed. The observation probability $p_{j}$ of the $j$-th
authentic equal to~$1$  if $j$ is a multiple of $3$, and equal
to $p$ otherwise, where we consider different values of~$ p $. Note that many authentic columns are fully observed,
so one cannot distinguish them from the corrupted columns based on only the number of observations. Figure~\ref{fig:trim_vs_no_trim} shows the relative errors of the output $L^{*}$ on the uncorrupted columns, i.e., $\left\Vert \P_{I_{0}^{c}}\left(L^{*}-L_{0}\right)\right\Vert _{F}/\left\Vert \P_{I_{0}^{c}}L_{0}\right\Vert _{F}$,
for different values of the observation probability $p$ and the number of corrupted columns $n_{c}$. Compared to no trimming, the trimming step often leads to much lower errors and allows for more corrupted columns. This agrees with our theoretical findings and shows that trimming is indeed crucial to good performance. \\

Having demonstrated the benefit of trimming when the $ p_j $'s are non-uniform, in the remaining experiments we set $p_{j}\equiv p$ for simplicity.

\subsubsection{Comparison with standard matrix completion and $L+S$}

While our theory and algorithm allow for the corrupted columns of
$C_{0}$ to have entries with arbitrarily large magnitude, we perform comparison in a more realistic setting with bounded corruption. In the second set of experiments, the $n_{c}$ non-zero columns of $C_{0}$ are identical, which equal the first column of $L_{0}$ on the locations of its observed entries, and are i.i.d.\ standard Gaussian on the other locations. These
columns are normalized to have the same norm as the first column of
$L_{0}$. The locations of the observed entries are also identical
across the columns of $C_{0}$, and are randomly selected according
to the Bernoulli model with probability $p$. Note that the columns
of $C_{0}$ have the same norm and observation probabilities as the
authentic columns. If we think of each column of $L_{0}$ as the ratings
of movies from an authentic user, then the above construction of $C_{0}$
mimics a rating manipulation scheme that is reported to be effective
in the literature~\cite{van2010manipulation}. In particular, the columns of $C_{0}$ are meant
to be similar to the ratings from an authentic user in $L_{0}$ on
the observed locations, while trying to skew the unobserved ones in
a coordinated fashion.

When only a small fraction of the entries are observed, the corrupted
columns $\P_{\Omega}(C_{0})$ can be viewed as a sparse matrix. Therefore,
to separate $L_{0}$ from $\P_{\Omega}(C_{0})$, one might think it
is possible to apply the techniques in \cite{candes2009robustPCA,chandrasekaran2011siam},
dubbed the $L+S$ approach, which decomposes a low-rank matrix and a sparse
matrix from their sum. In particular, 
one tries to decompose the input matrix $\P_{\Omega}(M)$ by solving the following convex program:
\begin{align}
\left(L^{*},S^{*}\right)=\arg & \min_{L,S} \; \left\Vert L\right\Vert _{*}+\lambda\left\Vert S\right\Vert _{1}\label{eq:L+S}\\
 & \text{s.t. }\P_{\Omega}(L+S) = \P_{\Omega}(M).\nonumber 
\end{align}
However, a central assumption of the $L+S$ approach, namely,
the support of the sparse matrix is spread out over the columns and
rows, is violated in the setup considered in this paper. Therefore,
it is no surprise that using the $L+S$ approach should not be successful.
This is indeed the case, as is illustrated numerically in our experiments. 

\begin{figure}
\begin{centering}
\begin{tabular}{cc}
\includegraphics[width=0.4\columnwidth]{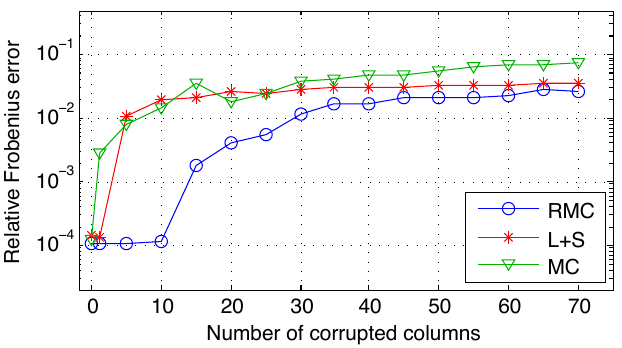} & \includegraphics[width=0.4\columnwidth]{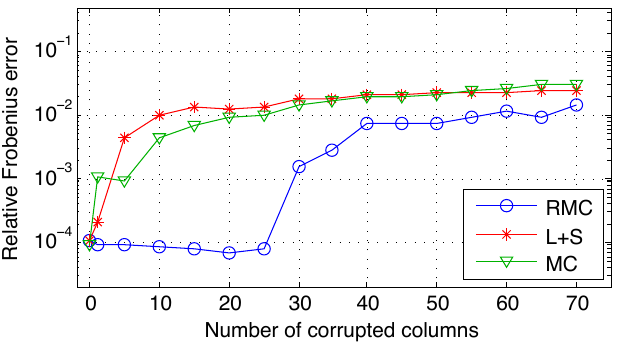}\tabularnewline
(a) & (b)\tabularnewline
\end{tabular}
\par\end{centering}

\protect\caption{\label{fig:compare3}Comparison of robust matrix completion in Algorithm~\ref{alg:MP}
(RMC), standard matrix completion (MC) and the $L+S$ approach.
The relative Frobenius norm errors are shown for recovering a $400\times400$ rank-$4$ matrix
with observation probabilities (a) $p=0.2$ and (b) $p=0.4$.  Each
point in the plots is the average of $10$ trials.}
\end{figure}

In particular, we compare our algorithm with the $L+S$ approach (with $\lambda$
set to $1/\sqrt{\max\left(m,n\right)}$ according to~\cite{candes2009robustPCA}),
as well as with standard matrix completion (which is equivalent to
solving~(\ref{eq:L+S}) with the additional constraint $S=0$). The
convex program~(\ref{eq:L+S}) is solved using the ADMM methods in~\cite{Lin2009_ALM}.
The results are shown in Figure~\ref{fig:compare3} for various
values of $p$ and $n_{c}$. We see that the $L+S$ and standard matrix
completion approaches are not robust under our setting, and our algorithm
has consistently better performance under both metrics considered. Moreover, with a higher observation probability~$p$, we can handle a larger number $n_{c}$ of corrupted columns and achieve essentially exact
recovery, which
is consistent with our theory.

\subsubsection{Random Corruption}

\begin{figure}
\begin{centering}
\begin{tabular}{cc}
\includegraphics[width=0.4\columnwidth]{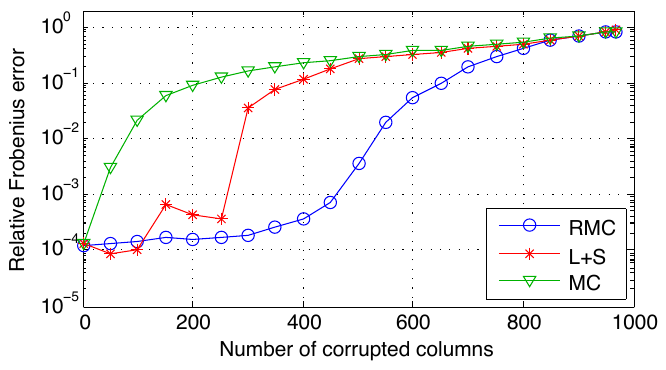} & \includegraphics[width=0.4\columnwidth]{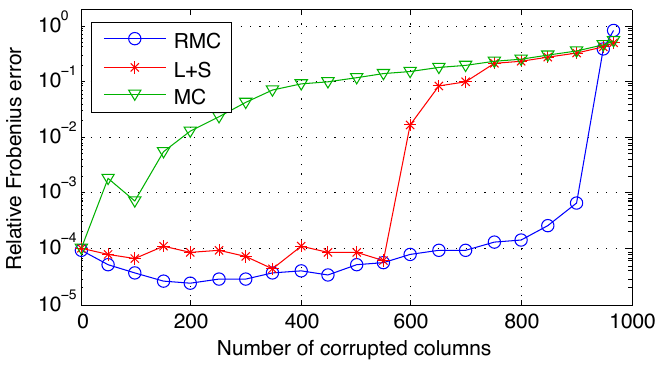}\tabularnewline
(a) & (b)\tabularnewline
\end{tabular}
\par\end{centering}

\protect\caption{\label{fig:dense} Comparison of robust matrix completion in Algorithm~\ref{alg:MP}
(RMC), standard matrix completion (MC) and the $L+S$ approach,
with randomly corrupted columns. The relative Frobenius norm
errors are shown for recovering a $200\times1000$ rank-$4$ matrix with observation probabilities (a) $p=0.3$ and (b) $p=0.6$. Each
point in the plots is the average of $10$ trials.}
\end{figure}

In this third set of experiments, we consider a more benign setting
of the corrupted columns, where these columns are generated randomly and independently
with i.i.d.\ Gaussian entries. The experiments are done under the setting
with rank $r=4$, $m=200$ rows and $n+n_{c}=1000$ columns. Figure~\ref{fig:dense}
shows the performance of the three algorithms for various $p$ and
$n_{c}$. Our algorithm again outperforms standard matrix completion
and the $L+S$ approaches. Perhaps more importantly, we see that our
algorithm succeeds under a much higher value of $n_{c}$ than in the
adversarial setting above. In particular, we recover the authentic
columns even when they are significantly out-numbered by the corrupted
columns, e.g., with $n=200$ and $n_{c}=800$. This result shows that with
such ``less adversarial'' corruption, the performance of our algorithm
is better than is guaranteed by our theory on worse case corruption. Rigorously
characterizing this phenomenon is an interesting future direction. 

\begin{figure}
\begin{centering}
\begin{tabular}{c}
\includegraphics[width=0.4\columnwidth]{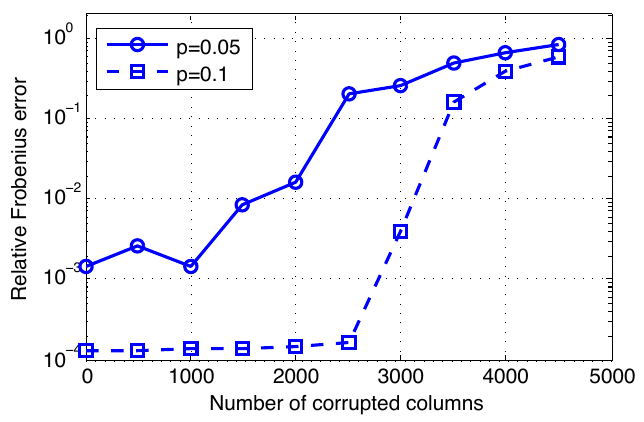}\tabularnewline
\end{tabular}
\par\end{centering}

\protect\caption{\label{fig:dense_large} Performance of our Algorithm~\ref{alg:MP} with random corruption. The relative Frobenius norm errors are shown for recovering a $1000\times5000$ rank-$8$ matrix with randomly corrupted columns and observation probabilities (a) $p=0.05$ and (b) $p=0.1$. Each 	point is the average of $10$ trials.}
\end{figure}

Finally, we demonstrate the applicability of our algorithms to larger
matrices with sparse observation. We consider a setting with rank
$r=8$, $m=1000$ rows and $n+n_{c}=5000$ columns, with observation
probability $p=0.05$ or $p=0.1$. The performance of our algorithm
is shown in Figure~\ref{fig:dense_large}. Again we see that our
algorithm is able to recover the true matrix even when there are many
corrupted columns. The average running time of each trial is less
than 2 minutes, indicating scalability to large problems.

\section{Proof of Theorem~\ref{thm:main}\label{sec:proof_main}}

In this section we prove the main Theorem~\ref{thm:main}. The proof
 requires a number of intermediate steps. Here
we provide a brief overview of the proof roadmap. By definition of
the success of Algorithm~\ref{alg:MP}, we need to show that
any optimal solution $\left(L^{*},C^{*}\right)$ of the program~(\ref{eq:L12formulation})
has the properties (i) $\P_{\I_{0}^{c}}L^{*}=L_{0}$, (ii) $\P_{U_{0}}L^{*}=L^{*}$
and (iii) $\I^{*}=\text{column-support}(C^{*})\subseteq\I_{0}$. A
central roadblock to this goal is that unless the adversary's corrupted
columns happen to be perfectly perpendicular to the column space of
the true low-rank matrix, \textit{$\left(L^{*},C^{*}\right)$ will
not be precisely equal to the ground truth} $\left(L_{0},C_{0}\right)$.
The reason is simple: if the corrupted columns have a non-perpendicular
component, then some part of that will be put into the $L^{*}$ matrix recovered by the optimization. Algorithmically, this matter is irrelevant:
as long as the corrupted columns are identified, and the recovered
$L^{*}$ matches the desired $L_{0}$ on the non-corrupted columns,
our objective is met, and the problem is solved. The analysis, however,
is significantly complicated: because $L^{*}\neq L_{0}$ in general
and we do not know what $L^{*}$ is exactly, we can no longer use
the standard approach in the matrix completion literature of proving
the ground-truth is the unique optimal solution of the convex program. \\

To prove the theorem, we use the idea of a \emph{primal-dual witness}:
we construct a primal solution $(\bar{L},\bar{C})$ and a dual certificate
$\bar{Q}$ such that:
\begin{itemize}
\item $(\bar{L},\bar{C})$ has the desired properties (i)--(iii);
\item $\bar{Q}$ certifies that any optimal solution to~(\ref{eq:L12formulation})
is either equal to $\left(\bar{L},\bar{C}\right)$, or is in a subspace defined by $\left(\bar{L},\bar{C}\right)$ and still has the properties (i)--(iii).
\end{itemize}
Beyond the above obstacle, challenges arise because of the simultaneous presence
of three matrix structures: low rank, entry-wise sparse, and column sparse.
This requires a number of additional innovations, including concentration
bounds involving these structures. 

In the rest of the section we present the details of the proof, which
is divided into several steps. In Section~\ref{sec:notation},
we provide the notation and preliminaries of the proof, and show
that it suffices to consider a simpler setting. In Section~\ref{sec:primal_contruct}
we construct the primal solution $(\bar{L},\bar{C})$ and study its
properties. In Section~\ref{sec:opt_cond}, we describe the conditions
that a dual certificate $\bar{Q}$ needs to satisfy.  We construct
the dual certificate $\bar{Q}$ in Section~\ref{sec:dual_construct},
and then prove that it indeed satisfies the desired conditions
with high probability in Section~\ref{sec:dual_validate}. The proofs of the technical lemmas are deferred to the appendix.

\subsection{Notation and Preliminaries\label{sec:notation}}

For a vector $x$, $x_{i}$ is its $i$-th entry. For a matrix $A$,
$A_{\cdot j}$ is its $j$-th column and $A_{ij}$ is its $(i,j)$-th
entry. Several standard matrix norms are used: $\left\Vert A\right\Vert _{\ast}$
is the nuclear norm (the sum of singular values), $\left\Vert A\right\Vert $
is the spectral/operator norm (the largest singular values), $\left\Vert A\right\Vert _{\infty}$
is the matrix infinity norm (the largest absolute value of the entries),
$\left\Vert A\right\Vert _{1,2}$ is the sum of $\ell_{2}$ norms
of the columns of $A$, $\left\Vert A\right\Vert _{\infty,2}$ is
the largest $\ell_{2}$ norm of the columns of $A$, and finally $\left\Vert A\right\Vert _{F}$
is the Frobenius norm. We also define $\ell_{(\infty,2)^{2}}$ norm
of a matrix by $\left\Vert A\right\Vert _{(\infty,2)^{2}}:=\max\left\{ \left\Vert A\right\Vert _{\infty,2},\left\Vert A^{\top}\right\Vert _{\infty,2}\right\} $,
which is the largest $\ell_{2}$ norm of the columns and rows of $A$.
For any positive integer $k$, $[k]:=\left\{ 1,2,\ldots,k\right\} $.
We also use the notation $a\wedge b:=\min\left\{ a,b\right\} $ and
$a\vee b:=\max\left\{ a,b\right\} .$ The letter $c$ and their derivatives
($c_{2}$ etc.) denote unspecified constants that are, however, universal
in that they are independent of $p$, $\gamma$, $\beta$, $\rho$,
$n$, $n_{c}$, $m_{}$ and~$r$. By \emph{with high probability}
(\emph{w.h.p.}), we mean with probability at least $1-c(m+n)^{-10}$
for some numerical constant $c>0$.

Recall that $\tilde{\Omega}:=\Omega\cap\left([m]\times\I_{0}^{c}\right)$
is the set of observed entries on the non-corrupted columns in $I_{0}^{c}$. We use $\Omega_{c}:=\Omega\cap\left([m]\times\I_{0}\right)$ to
denote the set of observed entries on the corrupted columns in $I_{0}$. 
We abuse notation by using $\Omega$ (and similarly $\Omega^{c}$,
$\tilde{\Omega}$, $\tilde{\Omega}^{c}$ etc) to denote both the set
of matrix entries and the linear subspace of matrices supported on
these entries. Similarly $\I_{0}$ and $\I_{0}^{c}$ denote both the
set of column indices and the linear subspace of matrices supported
on these columns. 
The operators $ \P_{\tilde{\Omega}}, \P_{\Omega_{c}}, \P_{I_{0}} $ and $ \P_{I_{0}^{c}}$ etc.\ are the corresponding projections onto the sets of matrices supported on $ \tilde{\Omega}, \Omega_{c}, I_{0}, I_{0}^{c} $ etc.

Denote the SVD of $L_{0}$ as $U_{0}\Sigma_{0}V_{0}^{\top}$, where
$U_{0}\in\mathbb{R}^{m\times r}$ and $V_{0}\in\mathbb{R}^{(n+n_{c})\times r}$.
Let $\P_{U_{0}}$ be the projection given by $\P_{U_{0}}A=U_{0}U_{0}^{\top}A$,
i.e., projecting each column of $A$ onto the column space of $L_{0}$,
where $A$ is any matrix with $m$ rows. The complimentary operation $\P_{U_{0}^{\bot}}A:=A-\P_{U_{0}}A$
projects the columns of $A$ onto the subspace orthogonal to the column
space of $L_{0}$. Similarly for the row space we define the projection
$\P_{V_{0}}A:=AV_{0}V_{0}^{\top}$. We define the subspace 
\[
\T_{0}:=\{U_{0}X^{\top}+YV_{0}^{\top}:X\in\RR^{(n+n_{c})\times r}\text{ with }\PIO X^{\top}=0,\; Y\in\RR^{m\times r}\}\subset\RR^{m\times(n+n_{c})},
\]
i.e., the set of matrices which has the same column or row space as
$L_{0}$ and is supported on the columns in $\I_{0}^{c}$; note that
$T_{0}\subset\I_{0}^{c}.$ The projection $\P_{T_{0}}$ is given by
\[
\P_{\T_{0}}A:=\P_{U_{0}}A+\P_{V_{0}}A-\P_{U_{0}}\P_{V_{0}}A=\P_{U_{0}}A+\P_{U_{0}^{\bot}}\P_{V_{0}}A
\]
for $A\in\RR^{m\times(n+n_{c})}$, and the complementary projection
is 
\[
\P_{\T_{0}^{\perp}}A:=A-\P_{T_{0}}A=(Id-U_{0}U_{0}^{\top})A(Id-V_{0}V_{0}^{\top}),
\]
where $Id$ is the identity matrix with appropriate dimension. We
note that the range of $\P_{T_{0}}$ is larger than $T_{0}$ since
the matrix $\P_{T_{0}}A$ may have non-zero columns in $\I_{0}$.
Nevertheless, when restricted to the subspace $I_{0}^{c}$, $\P_{T_{0}}$
is indeed the Euclidean projection onto $T_{0}$. Also note that the column-wise
projection $\P_{U_{0}}$ commutes with the row-wise projections $\P_{V_{0}}$,
$\PIO$ and $\PIOc$, since row-wise projections are given by right
multiplying a matrix, whereas column-wise projections are left multiplications.
We use $\mathcal{I}$ to denote the identity mapping on $\mathbb{R}^{m\times(n+n_{c})}$.

We provide a summary of the notation used in the proof in Table~\ref{tab:notation}.

\begin{table}

\protect\caption{\label{tab:notation}Summary of Notation}

\begin{centering}
\begin{tabular}{c|l}
\hline 
Notation & Meaning\tabularnewline
\hline 
$M$ & Input data matrix\tabularnewline
$\Omega$ & Set of observed indices\tabularnewline
$\tilde{\Omega}$ & Set of observed indices on the non-corrupted columns\tabularnewline
$\Omega_{c}$ & Set of observed indices on the corrupted columns\tabularnewline
$\hat{\Omega}$ & Trimmed set of observed indices\tabularnewline
$L^{*},C^{*}$ & An optimal solution to the program~(\ref{eq:L12formulation})\tabularnewline
$L_{0},C_{0}$ & True low-rank matrix and outlier matrix\tabularnewline
$U_{0},V_{0},T_{0}$ & The left and right singular vectors of $L_{0}$ and the corresponding
tangent space\tabularnewline
$I_{0}$ & Set of the indices of the corrupted columns (i.e., non-zero columns
of $C_{0}$)\tabularnewline
$\bar{L},\bar{C}$ & A solution to the oracle problem~(\ref{eq:oracle})\tabularnewline
$\bar{U},\bar{V},\bar{T}$ & The left and right singular vectors of $\bar{L}$ and the corresponding
tangent space\tabularnewline
$\bar{I}$ & The set of the indices of the non-zero columns of $\bar{C}$\tabularnewline
$\bar{H}$ & The column-wise normalized version of $\bar{C}$\tabularnewline
$\bar{Q}$ & The dual certificate corresponding to $\left(\bar{L},\bar{C}\right)$\tabularnewline
$\P_{T_{0}},\P_{\bar{U}},\P_{\bar{I}^{c}},\P_{\tilde{\Omega}}, $ etc. & Projection operators on $\mathbb{R}^{m\times(n+n_{c})}$\tabularnewline
$\mathcal{I}$ & The identity mapping on $\mathbb{R}^{m\times(n+n_{c})}$\tabularnewline
$Id$ & The identity matrix\tabularnewline
\hline 
\end{tabular}
\par\end{centering}

\end{table}

\subsubsection{Equivalent Models and Trimming\label{sec:equiv}}

It turns out that we may simplify the proof by transferring to an equivalent setting with a simpler observation model and no trimming. Let $\hat{p}:=\min\left\{ p,\rho\right\} $
and $\beta:=\frac{\rho}{\hat{p}}$. The conditions for $p,\gamma$
and $\lambda$ in Theorem~\ref{thm:main} can be written equivalently
as (with possibly different constants $c_{1}$ and $c_{2}$) 
\begin{align}
\hat{p} & \ge c_{1}\frac{\mu r\log^{2}(m+n)}{\min(m,n)},\label{eq:p_cond_eqv}\\
\gamma & \le c_{2}\frac{\hat{p}}{\mu r\sqrt{\beta\mu r}\log^{3}(m+n)},\label{eq:gamma_cond_eqv}\\
\lambda & \in\left[\sqrt{\frac{\mu r\log(m+n)}{\hat{p}n}},\frac{1}{48\sqrt{\sqrt{\beta\mu r}\gamma n\log(m+n)}}\right].\label{eq:lambda_cond_eqv}
\end{align}
We first note that the only randomness in the problem is the distribution
of $\tilde{\Omega},$ the set of observed indices on the non-corrupted
columns in $I_{0}^{c}$. We claim that it suffices to establish the theorem assuming uniform observation probability on $I_{0}^{c}$. 
To establish this claim, we need some notation. 
Without loss of generality we assume $I_{0}^{c}=[n]$. Let $\vec{p}$
be the vector in $\RR^{n}_+$ with elements $p_{1},\ldots,p_{n}$, where we
recall that $p_{j}\ge p\ge\hat{p}$ for all $j\in[n]$ by Assumption~\ref{asm:sampling}.
Denote by $\mathbb{P}_{\text{Ber}(\vec{p})}$ and $\mathbb{P}_{\text{UBer}(\hat{p}/4)}$
the probabilities calculated respectively when $\tilde{\Omega}$ follows the Bernoulli model with probabilities $\vec{p} = ( p_{j} )$, and when $\tilde{\Omega}$ follows the Bernoulli model with uniform probability
$\hat{p}/4$. The following lemma, proved in the appendix, connects the
success probabilities of Algorithm~\ref{alg:MP} under these two
models.
\begin{lem}
\label{lem:model}
Recall that $p_{j}\ge\hat{p}$ for all $j$,  and suppose that the
condition~(\ref{eq:p_cond_eqv}) holds with a sufficiently large
constant $c_{1}$. If $\mathbb{P}_{\textrm{UBer}(\hat{p}/4)}\left[success\right]\ge1-17(m+n)^{-5}$
, then $\mathbb{P}_{\text{Ber}(\vec{p})}\left[success\right]\ge1-20(m+n)^{-5}$. 
\end{lem}
The lemma implies that it suffices to prove Theorem~\ref{thm:main}
assuming $\tilde{\Omega}$ follows the Bernoulli model with uniform
probability $\hat{p}/4$.

Now define the set $\Omega':=\tilde{\Omega}\cup\left(\hat{\Omega}\cap([m]\times I_{0})\right)$,
which is the set of observed indices with only the columns in $I_{0}$
trimmed. If the condition~(\ref{eq:p_cond_eqv}) holds with a sufficiently large constant $c_{1}$, then w.h.p.\ with respect to $\mathbb{P}_{\textrm{UBer}(\hat{p}/4)}$,
$\Omega'$ is equal to $\hat{\Omega}$, the fully trimmed set. (This
is because by Bernstein's inequality, each uncorrupted column in $I_{0}^{c}$ has no more than $2\cdot\frac{\hat{p}}{4}m\le\rho m$ observed entries
w.h.p.\ and therefore is not changed by trimming.) In other words,
the convex program~(\ref{eq:L12formulation}) with $\Omega'$ as
the input is identical  to the one with input $\hat{\Omega}$ w.h.p.,
so it suffices to prove Algorithm~\ref{alg:MP} succeeds w.h.p.\ assuming
the columns in $I_{0}^{c}$ are not trimmed. Finally, note that after
trimming the number of remaining observations on each corrupted column
in $I_{0}$ is at most $\rho m$. Combining these observations, we
conclude that we may replace the sampling Assumption~\ref{asm:sampling}
with the following new Assumption~\ref{asm:sampling_2}, and study
Algorithm~\ref{alg:MP} without trimming (i.e., only the convex program).
Note that in Assumption~\ref{asm:sampling_2} we have changed the probability from $\hat{p}/4$ to $\hat{p},$
which only affects the constant $c_{1}$ in the condition~(\ref{eq:p_cond_eqv}).
\begin{assumption}
[Sampling 2]\label{asm:sampling_2}The set $\tilde{\Omega}$ is sampled
from the Bernoulli model with uniform probability $\hat{p}$ on $[m]\times I_{0}^{c}$,
and is independent of the locations of the observed entries on the
corrupted columns. For each $j\in\I_{0}$, we have $\left|\Omega\cap\left([m]\times\{j\}\right)\right|\le2\rho m$.
\end{assumption}
Summarizing the arguments above, we have established that in order
to prove Theorem~\ref{thm:main}, it suffices to prove the following: 
\begin{quotation}
Under Assumptions~\ref{asm:incoherence} and~\ref{asm:sampling_2},
if the conditions~(\ref{eq:p_cond_eqv})--(\ref{eq:lambda_cond_eqv})
hold, then with probability at least $1-16(m+n)^{-5}$, the program~(\ref{eq:L12formulation})
with $\Omega$ as the input succeeds, i.e., any optimal solution to
the program satisfies the properties (i)--(iii) stated at the beginning of this section.
\end{quotation}

\subsection{Primal Construction\label{sec:primal_contruct}}

We now construct the primal solution $\left(\bar{L},\bar{C}\right)$.
Recall that $\Omega_{c}$ is the observed indices on the corrupted
columns~$I_{0}$. Let $(\bar{L},\bar{C})$ be an optimal solution
to the following \emph{oracle problem}:
\begin{equation}
\begin{aligned}\min_{L,C}\quad & \left\Vert L\right\Vert _{*}+\lambda\left\Vert C\right\Vert _{1,2}\\
\text{s.t.}\quad & \P_{\Omega_{c}}(L+C)=\P_{\Omega_{c}}(M)\\
 & \PIOc(L)=L_{0}.\\
 & \P_{U_{0}}(L)=L\\
 & \P_{\I_{0}}(C)=C.
\end{aligned}
\label{eq:oracle}
\end{equation}
Note that we have imposed the desired properties of $\left(L^{*},C^{*}\right)$
as constraints in the oracle problem. Let $\bar{U}\bar{\Sigma}\bar{V}$
be the rank-$r$ SVD of $\bar{L}$ (the lemma below shows that $\bar{L}$
has rank $r$) and $\bar{I}:=\text{column-support}(\bar{C})$. We
define several subspaces and projections analogously to those for
$L_{0}$: $\P_{\bar{U}}A:=\bar{U}\bar{U}^{\top}A$, $\P_{\bar{U}^{\bot}}A=A-\P_{\bar{U}}A$,
$\P_{\bar{V}}:=A\bar{V}\bar{V}^{\top}$, $\bar{T}:=\left\{ \bar{U}X^{\top}+Y\bar{V}^{\top}:X\in\RR^{(n+n_{c})\times r},Y\in\RR^{m\times r}\right\} $,
$\P_{\bar{T}}A:=\P_{\bar{U}}A+\P_{\bar{V}}A-\P_{\bar{U}}\P_{\bar{V}}A$,
and $\P_{\bar{T}^{\bot}}A:=A-\P_{\bar{T}}A$. 

The following lemma, whose proof is given in the appendix, relates
some basic properties of the oracle solution $(\bar{L},\bar{C})$
to the ground truth~$(L_{0},C_{0})$. 
\begin{lem}
\label{lem:TandT0}We have the following: (a) $\P_{\bar{U}}=\P_{U_{0}}$
and $\bar{I}\subseteq I_{0}$; (b) $\max_{1\le j\le n+n_{c}}\left\Vert \left(\PIOc\bar{V}^{\top}\right)e_{j}\right\Vert _{2}\le\sqrt{\frac{\mu r}{n}}$;
(c) $\PIOc\PT=\PTO\PIOc\PT$; (d) $\PT\PIOc=\PT\PTO\PIOc$; (e) $\P_{\bar{\T}^{\bot}}\P_{\T_{0}^{\bot}}\PIOc=\P_{\T_{0}^{\bot}}\PIOc$. 
\end{lem}
Since all the constraint in~(\ref{eq:oracle}) are linear, by standard
convex analysis the optimal solution $(\bar{L},\bar{C})$ must satisfy
the KKT conditions. That is, there exist Lagrange multipliers $A_{1}$,
$A_{2}$, $A_{3}$ and $A_{4}$ (corresponding to the four constraints
in the oracle problem) and matrices $F$ and $G$ such that $\P_{\bar{T}}F=0$
, $\left\Vert F\right\Vert \le1$, $\P_{\bar{I}^{c}}\bar{H}=0$, $\bar{H}_{\cdot j}=\bar{C}_{\cdot j}/\left\Vert \bar{C}_{\cdot j}\right\Vert _{2}$
for all $j\in\bar{I}$, $G\in\bar{\I}^{c}$, $\left\Vert G\right\Vert _{\infty,2}\le1$,
and 
\begin{equation}
\bar{U}\bar{V}^{\top}+F+\PIOc A_{2}+\left(\mathcal{I}-\P_{U_{0}}\right)A_{3}=\lambda\left(\bar{H}+G\right)+\P_{\I_{0}^{c}}A_{4}=\P_{\Omega_{c}}A_{1};\label{eq:oracle_subgrad}
\end{equation}
here $\bar{U}\bar{V}^{\top}+F$ is a subgradient of $\left\Vert L\right\Vert _{*}$
at $\bar{L}$, and $\bar{H}+G$ is a subgradient of $\left\Vert C\right\Vert _{1,2}$
at $\bar{C}$. Also note that $\bar{H}$ is the column-wise normalized
version of $\bar{C}$ with unit-norm nonzero columns. Define the matrix
$\bar{H}':=\bar{H}+\PIO G$. The following lemma characterizes $\bar{H}'$
and is proved in the appendix
\begin{lem}
\label{lem:H}We have the following: (a) $\bar{H}'\in\Omega_{c}$;
(b) $\P_{\bar{I}}\bar{H}'=\bar{H}$; (c) $\left\Vert \PIc\bar{H}'\right\Vert _{\infty2}\le1$;
(d) \textup{$\bar{U}\P_{\I_{0}}\bar{V}^{\top}=\P_{U_{0}}(\lambda\bar{H}')$};
(e) $\bar{H}$ and $\bar{H}'$ are independent of $\tilde{\Omega}$\textup{.}
\end{lem}

\subsection{Success Condition\label{sec:opt_cond}}

Recall that in Section~\ref{sec:equiv} we show that it suffices
to prove the convex program~(\ref{eq:L12formulation}) succeeds without
trimming. The following proposition, proved in the appendix, provides
a deterministic sufficient condition for such success. The success
condition involves the quantities $\bar{T}$, $\bar{U}$, $\bar{V}$,
$\bar{I}$ and $\bar{H}$ of the oracle solution $(\bar{L},\bar{C})$
constructed in the last subsection.
\begin{prop}
\label{prop:opt_cond}If the following conditions hold: 
\begin{enumerate}
\item \textup{$\left\Vert \left(\hat{p}^{-1}\PTO\P_{\tilde{\Omega}}\PTO-\PTO\right)Z\right\Vert _{F}\le\frac{1}{2}\left\Vert Z\right\Vert _{F}$
for all $Z\in\I_{0}^{c}$.}
\item \textup{$\I_{0}\cap\textrm{range}\left(\P_{\bar{V}}\right)=\{0\}$}. 
\item There exists a matrix $\bar{Q}\in\RR^{m\times(n+n_{c})}$ (called
an approximate dual certificate) which satisfies

\begin{enumerate}
\item $\bar{Q}\in\Omega$;
\item $\bar{U}\bar{V}^{\top}-\P_{\bar{\T}}\bar{Q}=\P_{\bar{\T}}D$ for some
$D\in\mathbb{R}^{m\times(n+n_{c})}$ with $D\in\I_{0}^{c}$ and $\left\Vert D\right\Vert _{F}\le\sqrt{\frac{\hat{p}}{2}}\min\left\{ \frac{1}{4},\frac{\lambda}{4}\right\} $;
\item $\left\Vert \P_{\bar{\T}^{\bot}}\bar{Q}\right\Vert \le\frac{1}{2}$;
\item $\P_{\bar{\I}}\bar{Q}=\lambda\bar{H}$;
\item $\left\Vert \P_{\bar{\I}^{c}\cap\I_{0}}\bar{Q}\right\Vert _{\infty,2}\le\lambda$;
\item $\left\Vert \PIOc\bar{Q}\right\Vert _{\infty,2}\le\frac{\lambda}{2}$.
\end{enumerate}
\end{enumerate}
Then any optimal solution $(L^{*},C^{*})$ to the program~(\ref{eq:L12formulation})
must satisfy $\P_{\I_{0}^{c}}L^{*}=L_{0}$, $\P_{U_{0}}L^{*}=L^{*}$
and $\P_{\I_{0}}C^{*}=C^{*}$, which means Algorithm~\ref{alg:MP}
succeeds.
\end{prop}

\subsubsection{Approximate Isometry and Contraction}

We now show that the conditions 1 and 2 in Proposition~\ref{prop:opt_cond}
are satisfied w.h.p.\ under our model assumptions and the conditions~(\ref{eq:p_cond_eqv})--(\ref{eq:lambda_cond_eqv}).
Recall that by Assumption~\ref{asm:sampling_2} the set $\tilde{\Omega}$
follows the Bernoulli model with uniform probability $\hat{p}$.
The following lemma establishes the approximate isometry property
in the condition 1.
\begin{lem}
\label{lem:op}Suppose $\hat{p}\ge\frac{\mu r}{m\wedge n}\log(m+n)$,
then w.h.p.\ \textup{we have: for all $Z\in\I_{0}^{c}$,}
\begin{equation}
\left\Vert \left(\hat{p}^{-1}\PTO\P_{\tilde{\Omega}}\PTO-\PTO\right)Z\right\Vert _{F}\le\frac{1}{2}\left\Vert Z\right\Vert _{F}.\label{eq:invertibility}
\end{equation}

\end{lem}
The lemma is a variant of the standard approximate isometry inequality
in the literature of matrix completion/decomposition~\cite{candes2009robustPCA,chen2011LSarxiv,li2013constantCorruption}.
In particular, we note that the operator $\hat{p}^{-1}\PTO\P_{\tilde{\Omega}}\PTO-\PTO$
maps the subspace $T_{0}\subset\I_{0}^{c}$ to itself, so Lemma~\ref{lem:op} is an immediate consequence of Part 1) of Lemma 11 in~\cite{chen2011LSarxiv}. 

The next lemma, proved in the appendix, shows that the operator $\P_{\bar{V}}\PIO\P_{\bar{V}}$
is a contraction, which in particular implies the condition 2 in Proposition~\ref{prop:opt_cond}.
\begin{lem}
\label{lem:PVPI}If $\lambda^{2}\le\frac{1}{2\gamma n}$ , then $\left\Vert \P_{\bar{V}}\PIO\P_{\bar{V}}(Z)\right\Vert _{F}\le\frac{1}{2}\left\Vert Z\right\Vert _{F}$
and \textup{$\left\Vert \P_{\bar{V}}\PIO\P_{\bar{V}}(Z)\right\Vert \le\frac{1}{2}\left\Vert Z\right\Vert $
for any matrix $Z$.}
\end{lem}
Note that the requirements on $\hat{p}$ and $\lambda$ in the above
lemmas are satisfied under the conditions~(\ref{eq:p_cond_eqv})
and~(\ref{eq:lambda_cond_eqv}). We therefore have established the conditions 1 and 2 in Proposition~\ref{prop:opt_cond}.  To prove the theorem, it remains to construct a dual certificate $\bar{Q}$
obeying the conditions $3(a)$--$(f)$ in Proposition~\ref{prop:opt_cond} w.h.p., which is done in the next subsection.

\subsection{Dual Construction\label{sec:dual_construct}}

We build $\bar{Q}$ in two steps. In the first step we construct
a matrix $Q$ that satisfies all the requirements except $3(a)$. By
Lemma~\ref{lem:PVPI}, we know the operator $\P_{\bar{V}}\P_{\I_{0}^{c}}\P_{\bar{V}}=\P_{\bar{V}}-\P_{\bar{V}}\PIO\P_{\bar{V}}$
is invertible on $\textrm{range}\left(\P_{\bar{V}}\right)$ (as a
subspace of $\RR^{m\times(n+n_{c})}$), with its inverse given by
\begin{equation}
\mc B:=\left(\P_{\bar{V}}\P_{\I_{0}^{c}}\P_{\bar{V}}\right)^{-1}=\P_{\bar{V}}+\sum_{i=1}^{\infty}\left(\P_{\bar{V}}\PIO\P_{\bar{V}}\right)^{i}.\label{eq:inverse}
\end{equation}
We define a matrix $Q$ by
\[
Q:=\bar{U}\bar{V}^{\top}+\lambda\bar{H}'-\lambda\P_{U_{0}}\bar{H}'-\P_{\I_{0}^{c}}\P_{\bar{V}}\mc B\P_{\bar{V}}\P_{\bar{U}^{\bot}}\left(\lambda\bar{H}'\right).
\]
It is straightforward to check that $Q$ has the following properties
(proof in the appendix):
\begin{lem}
\label{lem:Q_properties}We have \textup{$\P_{\I_{0}}Q=\lambda\bar{H}'$,
$\P_{\bar{\T}}Q=\bar{U}\bar{V}^{\top}$, and} 
\begin{align*}
\left\Vert \P_{\bar{V}}\P_{\bar{U}^{\bot}}(\lambda\bar{H}')\right\Vert \le\left\Vert \lambda\bar{H}'\right\Vert \le\left\Vert \lambda\bar{H}'\right\Vert _{F} & \le\lambda\sqrt{\gamma n}.
\end{align*}

\end{lem}
While not needed in the sequel, it is a simple exercise to check that
Lemmas~\ref{lem:Q_properties} and~\ref{lem:PVPI} together imply
$\left\Vert \P_{\bar{\T}^{\bot}}Q\right\Vert \le3\lambda\sqrt{\gamma n}\le\frac{1}{2}$
and $\left\Vert \PIOc Q\right\Vert _{\infty,2}\le\left(1+2\lambda\sqrt{\gamma n}\right)\sqrt{\frac{\mu r}{n}}\le\frac{1}{2}\lambda$
under the condition~(\ref{eq:lambda_cond_eqv}). Therefore, $Q$
satisfies the condition 3 in Proposition~\ref{prop:opt_cond} except
for the requirement of being an element of~$\Omega$. Note that this
requirement can only potentially fail on the columns in $\I_{0}^{c}$
since $\P_{\I_{0}}Q=\lambda\bar{H}'\in\Omega_{c}$. As the second
step of building the dual certificate, we use the a variant of the
golfing scheme in~\cite{gross2009anybasis} to convert $Q$ to a
matrix $\bar{Q}$ that obeys this requirement. Set $k_{0}=20\log(m+n)$
and $p'=1-(1-\hat{p})^{1/k_{0}}$. Let $\tilde{\Omega}_{k}$, $k=1,\ldots,k_{0}$
be sets of entries sampled independently from the Bernoulli model
on $[m]\times I_{0}^{c}$ with uniform probability $p'$; that is,
$\mathbb{P}\left((i,j)\in\tilde{\Omega}_{k}\right)=p'$ independently
of all others for all $(i,j)\in[m]\times\I_{0}^{c}$ and $k\in[k_{0}]$.
We may assume $\tilde{\Omega}=\bigcup_{k=1}^{k_{0}}\tilde{\Omega}_{k}$,
which does not change the distribution of $\tilde{\Omega}$. Note
that $p'\ge\hat{p}/k_{0}\ge c_{1}\frac{\mu r\log(m+n)}{20(m\wedge n)}$
under the condition~(\ref{eq:p_cond_eqv}). We set $Y_{0}:=0$ and
define the matrices $\left\{ Y_{k}\right\} $ recursively by 
\[
Y_{k}:=Y_{k-1}+\frac{1}{p'}\P_{\tilde{\Omega}_{k}}\P_{\T_{0}}\left(\P_{\I_{0}^{c}}Q-Y_{k-1}\right),k=1,\ldots,k_{0}.
\]
The final dual certificate is given by $\bar{Q}=\P_{\I_{0}}Q+Y_{k_{0}}.$

\subsection{Verification of the Dual Certificate\label{sec:dual_validate}}

We now verify that the dual certificate $\bar{Q}$ constructed above
satisfies all the requirements $3(a)$--$3(f)$ in Proposition~\ref{prop:opt_cond}
under the conditions~(\ref{eq:p_cond_eqv})--(\ref{eq:lambda_cond_eqv}).
We have $\PIOc\bar{Q}=Y_{k_{0}}\in\tilde{\Omega}$ by construction
and $\P_{\I_{0}}\bar{Q}=\P_{\I_{0}}Q=\lambda\bar{H}'\in\Omega_{c}$
by part $(a)$ of Lemma~\ref{lem:H}, so the condition $3(a)$ holds.
Moreover, by part~$(b)$ and~$(c)$ of Lemma~\ref{lem:H} we have
$\PI\bar{Q}=\lambda\PI\bar{H}'=\bar{H}$ and $\left\Vert \P_{\bar{\I}^{c}\cap\I_{0}}\bar{Q}\right\Vert _{\infty.2}=\lambda\left\Vert \P_{\bar{\I}^{c}\cap\I_{0}}\bar{H}'\right\Vert _{\infty,2}\le\lambda$, so the conditions $3(d)$ and $3(e)$ are also satisfied. It remains
to verify $3(b)$, $3(c)$ and $3(f)$.

\subsubsection{Condition $3(b)$}

Define the linear operators $\mc A_{k}:=\PTO-\frac{1}{p'}\PTO\P_{\tilde{\Omega}_{k}}\PTO$
for $k=1,\ldots,k_{o}$ and the matrices $D_{k}=\PTO\left(\PIOc Q-Y_{k}\right)$
for $k=0,\ldots,k_{0}$. With this notation, we have $Y_{k}=Y_{k-1}+\frac{1}{p'}\P_{\tilde{\Omega}_{k}}D_{k-1}$
by definition, which implies
\begin{equation}
D_{k}=\left(\PTO-\frac{1}{p'}\PTO\P_{\tilde{\Omega}_{k}}\PTO\right)D_{k-1}=\mc A_{k}(D_{k-1}),\; k=1,\ldots,k_{0}.\label{eq:Yerror}
\end{equation}
It follows that with high probability,
\[
\left\Vert D_{k_{0}}\right\Vert _{F}=\left\Vert \mc A_{k_{0}}\mc A_{k_{0}-1}\cdots\mc A_{1}\left(D_{0}\right)\right\Vert _{F}\overset{(a)}{\le}\frac{1}{2^{k_{0}}}\left\Vert D_{0}\right\Vert _{F}\overset{(b)}{\le}\frac{1}{(m+n)^{10}}\left\Vert D_{0}\right\Vert _{F},
\]
where $(a)$ follows from Lemma \ref{lem:op} with $\tilde{\Omega}$
replaced by $\tilde{\Omega}_{k}$ and $(b)$ follows from our choice
of $k_{0}$. To bound $\left\Vert D_{0}\right\Vert _{F}$, we observe
that by definition of $Q$,
\begin{align}
D_{0}=\P_{\T_{0}}\P_{\I_{0}^{c}}Q & =\bar{U}\PIOc\bar{V}^{\top}+\P_{V_{0}}\P_{\I_{0}^{c}}\P_{\bar{V}}\mc B\P_{\bar{V}}\P_{\bar{U}^{\bot}}\left(\lambda\bar{H}'\right)\label{eq:PTPIcQ}
\end{align}
By (\ref{eq:inverse}) and Lemma \ref{lem:PVPI}, we know that for
any matrix $Z$,
\begin{equation}
\left\Vert \mc B(Z)\right\Vert _{F}\le\sum_{i=0}^{\infty}\left(\frac{1}{2}\right)^{i}\left\Vert Z\right\Vert _{F}\le2\left\Vert Z\right\Vert _{F}.\label{eq:B_norm}
\end{equation}
Combining the last two equations~(\ref{eq:PTPIcQ}) and~(\ref{eq:B_norm})
gives
\[
\left\Vert D_{0}\right\Vert _{F}\le\left\Vert \bar{U}\bar{V}^{\top}\right\Vert _{F}+2\left\Vert \lambda\bar{H}'\right\Vert _{F}\le\sqrt{r}+2\lambda\sqrt{\gamma n},
\]
where the last inequality follows from Lemma~\ref{lem:Q_properties}.
It follows that
\begin{equation}
\left\Vert D_{k_{0}}\right\Vert _{F}\le\frac{1}{(m+n)^{10}}\left(\sqrt{r}+2\lambda\sqrt{\gamma n}\right)\le\frac{\sqrt{\hat{p}}}{2}\min\left\{ \frac{1}{4},\frac{\lambda}{4}\right\} ,\label{eq:D_norm}
\end{equation}
where the last inequality follows from the conditions~(\ref{eq:p_cond_eqv})
and~(\ref{eq:lambda_cond_eqv}). On the other hand, since $\bar{Q}=\PIOc Y_{k_{0}}+\PIO Q$
and $\P_{\bar{\T}}Q=\bar{U}\bar{V}^{\top}$ by Lemma~, we have
\begin{align}
\bar{U}\bar{V}-\P_{\bar{\T}}\bar{Q} & =\P_{\bar{\T}}Q-\PT\left(\PIOc Y_{k_{0}}+\PIO Q\right)=\P_{\bar{T}}\PIOc\left(Q-Y_{k_{0}}\right)=\PT D_{k_{0}},\label{eq:equal_D}
\end{align}
where the last equality follows from Part $(d)$ of Lemma~\ref{lem:TandT0}.
We conclude that the condition 3(b) in Proposition~\ref{prop:opt_cond}
holds by combining~(\ref{eq:equal_D}), (\ref{eq:D_norm}) and the
fact that $D_{k_{0}}\in T_{0}\subseteq\I_{0}^{c}$.

\subsubsection{Condition $3(c)$}

We may write
\begin{align*}
\P_{\bar{\T}^{\bot}}\bar{Q} & =\P_{\bar{\T}^{\bot}}\left(\lambda\bar{H}'\right)+\P_{\bar{\T}^{\bot}}\PTO Y_{k_{0}}+\P_{\bar{\T}^{\bot}}\P_{\T_{0}^{\bot}}Y_{k_{0}}\\
 & =\P_{\bar{\T}^{\bot}}\left(\lambda\bar{H}'\right)+\left(\P_{\bar{\T}^{\bot}}\PTO\PIOc Q-\P_{\bar{\T}^{\bot}}D_{k_{0}}\right)+\P_{\T_{0}^{\bot}}Y_{k_{0}},
\end{align*}
where the first equality follows from definition of $\bar{Q}$, and
the second equality follows from $Y_{k_{0}}\in T_{0}\subseteq I_{0}^{c}$
and part $(e)$ of Lemma~\ref{lem:TandT0}. Hence we have 
\[
\left\Vert \P_{\bar{\T}^{\bot}}\bar{Q}\right\Vert \le\left\Vert \lambda\bar{H}'\right\Vert +\left\Vert D_{k_{0}}\right\Vert +\left\Vert \P_{\bar{\T}^{\bot}}\PTO\PIOc Q\right\Vert +\left\Vert \P_{\T_{0}^{\bot}}Y_{k_{0}}\right\Vert .
\]
The condition $3(c)$ holds if each of the terms above is upper bounded
by $\frac{1}{8}$. By Lemma~, we have $\left\Vert \lambda\bar{H}'\right\Vert \le\lambda\sqrt{\gamma n}\le\frac{1}{16}$,
where the last inequality holds under the condition~(\ref{eq:lambda_cond_eqv}).
In~(\ref{eq:D_norm}) we already showed that $\left\Vert D_{k_{0}}\right\Vert \le\left\Vert D_{k_{0}}\right\Vert _{F}\le\frac{\sqrt{\hat{p}}}{8}\le\frac{1}{8}$.
Moreover, using~(\ref{eq:PTPIcQ}), we have
\[
\left\Vert \P_{\bar{\T}^{\bot}}\PTO\PIOc Q\right\Vert \le\left\Vert \P_{\bar{V}^{\bot}}\P_{V_{0}}\P_{\I_{0}^{c}}\P_{\bar{V}}\mc B\P_{\bar{V}}\P_{\bar{U}^{\bot}}(\lambda\bar{H}')\right\Vert _{F}\overset{(a)}{\le}2\left\Vert \lambda\bar{H}'\right\Vert _{F}\le\frac{1}{8},
\]
where (a) follows~(\ref{eq:B_norm}) and the fact that projections
do not increase the Frobenius norm. It remains to bound $\left\Vert \P_{\T_{0}^{\bot}}Y_{k_{0}}\right\Vert $
by~$\frac{1}{8}$. 

For brevity we introduce some additional notation. Let $D_{0}^{U}:=\bar{U}\PIOc\bar{V}$,
and $D{}_{0}^{V}:=\P_{\bar{U}^{\bot}}\P_{V_{0}}\P_{\I_{0}^{c}}\mc B\P_{\bar{V}}\left(\lambda\bar{H}'\right)$,
$D_{k}^{U}:=\mc A_{k}\mc A_{k-1}\cdots\mc A_{1}(D_{0}^{U})$ and $D_{k}^{V}:=\mc A_{k}\mc A_{k-1}\cdots\mc A_{1}(D_{0}^{V})$
for $k=1,\ldots,k_{0}$. Note that for each $k\ge2$, $D_{k-1}^{U}$
and $D_{k-1}^{V}$ are independent of $\tilde{\Omega}_{k}$ by construction
the $\tilde{\Omega}_{k}$'s and part~$(e)$ of Lemma~\ref{lem:H}.
With these definitions, we have $D_{k}=D_{k}^{U}+D_{k}^{V}$ for $k=0,\ldots,k_{0}$
by~(\ref{eq:PTPIcQ}) and~(\ref{eq:Yerror}), and hence
\begin{equation}
Y_{k_{0}}=\sum_{k=1}^{k_{0}}\frac{1}{p'}\P_{\tilde{\Omega}_{k}}D_{k-1}=\sum_{i=1}^{k_{0}}\frac{1}{p'}\P_{\tilde{\Omega}_{k}}D_{k-1}^{U}+\sum_{k=1}^{k_{0}}\frac{1}{p'}\P_{\tilde{\Omega}_{k}}D_{k-1}^{V}.\label{eq:Ysum}
\end{equation}
Let $t$ be either $U$ or $V$. Since $D_{k-1}^{t}\in\T_{0}$ for
each $k$, we have
\begin{equation}
\sum_{k=1}^{k_{0}}\left\Vert \P_{\T_{0}^{\bot}}\frac{1}{p'}\P_{\tilde{\Omega}_{k}}D_{k-1}^{t}\right\Vert =\sum_{k=1}^{k_{0}}\left\Vert \P_{\T_{0}^{\bot}}\left(\frac{1}{p'}\P_{\tilde{\Omega}_{k}}D_{k-1}^{t}-D_{k-1}^{t}\right)\right\Vert \le\sum_{k=1}^{k_{0}}\left\Vert \left(\frac{1}{p'}\P_{\tilde{\Omega}_{k}}-\mc I\right)D_{k-1}^{t}\right\Vert .\label{eq:e1}
\end{equation}
To proceed, we need three lemmas involving the norms of a matrix after
certain random projections. Recall that $\tilde{\Omega}$ and $\tilde{\Omega}_{k}$
are sampled from the Bernoulli model with uniform probability $\hat{p}$
and $p'$, respectively. The first lemma bounds the spectral norm
using the $\ell_{\infty}$ and $\ell{}_{(\infty,2)^{2}}$ norm. This
lemma is proved in a recent report by the author~\cite{chen2013incoherence_arxiv},
but we provide a proof in the appendix for completeness.
\begin{lem}
\label{lem:op_inf} Let $Z$ be a fixed $m\times(n+n_{c})$ matrix
in $\I_{0}^{c}$. We have w.h.p.\ 
\[
\left\Vert \frac{1}{\hat{p}}\mathcal{P}_{\tilde{\Omega}}Z-Z\right\Vert \le\left(\frac{15\log(m+n)}{\hat{p}}\left\Vert Z\right\Vert _{\infty}+\sqrt{\frac{60\log(m+n)}{\hat{p}}}\left\Vert Z\right\Vert _{(\infty,2)^{2}}\right).
\]

\end{lem}
The next lemma, standard in matrix completion literature, further controls the $\ell_{\infty}$ norm.
\begin{lem}
\cite[Lemma 13, part 1]{chen2011LSarxiv}\label{lem:inf} Let $Z$
be a fixed $m\times(n+n_{c})$ matrix in $T_{0}$. If $\hat{p}>66\frac{\log(m+n)}{m\wedge n}$,
then w.h.p.\ we have 
\[
\left\Vert \frac{1}{\hat{p}}\PTO\mathcal{P}_{\tilde{\Omega}}\PTO Z-\PTO Z\right\Vert _{\infty}\le\frac{1}{2}\left\Vert Z\right\Vert _{\infty}.
\]

\end{lem}
The third lemma is new, which controls the $\ell_{(\infty,2)^{2}}$
norm. See the appendix for a proof.
\begin{lem}
\label{lem:inf_2}
The following holds for some constant $c_{0}>0$
and any fixed matrix $Z\in T_{0}$. If $\hat{p}\ge c_{0}\frac{\mu r\log(m+n)}{m\wedge n}$,
then we have w.h.p.\ 
\begin{align*}
\left\Vert \frac{1}{\hat{p}}\mathcal{P}_{T_{0}}\mathcal{P}_{\tilde{\Omega}}\mathcal{P}_{T_{0}}Z-\mathcal{P}_{T_{0}}Z\right\Vert _{\infty,2} & \le\frac{40\log(m+n)}{\hat{p}}\sqrt{\frac{\mu r}{n\wedge m}}\left\Vert Z\right\Vert _{\infty}+\sqrt{\frac{250\mu r\log(m+n)}{\hat{p}(n\wedge m)}}\left\Vert Z\right\Vert _{\infty,2}\\
 & \le\frac{1}{2}\sqrt{\frac{\log(m+n)}{\hat{p}}}\left\Vert Z\right\Vert _{\infty}+\frac{1}{2}\left\Vert Z\right\Vert _{\infty,2}.
\end{align*}
The same bound holds with the $\left\Vert \cdot\right\Vert _{\infty,2}$
norm replaced by the $\left\Vert \cdot\right\Vert _{(\infty,2)^{2}}$
norm.
\end{lem}
Applying Lemma~\ref{lem:op_inf} with $\tilde{\Omega}$ replaced
by $\tilde{\Omega}_{k}$ to the R.H.S. of~(\ref{eq:e1}) and using~$p'\ge\frac{\hat{p}}{20\log(m+n)}\gtrsim\frac{\mu r\log(m+n)}{m\wedge n}$
under the condition~(\ref{eq:p_cond_eqv}), we have for $t=U$ or~$V$,
\begin{equation}
\sum_{k=1}^{k_{0}}\left\Vert \left(\frac{1}{p'}\P_{\tilde{\Omega}_{k}}-\mc I\right)D_{k-1}^{t}\right\Vert \le\sum_{k=1}^{k_{0}}\frac{15\log^{2}(m+n)}{\hat{p}}\left\Vert D_{k-1}^{t}\right\Vert _{\infty}+\sum_{k=1}^{k_{0}}\sqrt{\frac{60\log^{2}(m+n)}{\hat{p}}}\left\Vert D_{k-1}^{t}\right\Vert _{(\infty,2)^{2}}.\label{eq:e2}
\end{equation}
We then apply Lemmas~\ref{lem:inf} and~\ref{lem:inf_2} with $\tilde{\Omega}$
replaced by $\tilde{\Omega}_{k}$ to the two norms in the last R.H.S.,
which gives
\begin{align*}
\left\Vert D_{k-1}^{t}\right\Vert _{\infty} & =\left\Vert \mc A_{k-1}\mc A_{i-2}\cdots\mc A_{1}\left(D_{0}^{t}\right)\right\Vert _{\infty}\le\frac{1}{2^{k-1}}\left\Vert D_{0}^{t}\right\Vert _{\infty},\\
\left\Vert D_{k-1}^{t}\right\Vert _{(\infty,2)^{2}} & =\left\Vert \mc A_{k-1}\mc A_{i-2}\cdots\mc A_{1}\left(D_{0}^{t}\right)\right\Vert _{(\infty,2)^{2}}\le\frac{1}{2^{k-1}}\left\Vert D_{0}^{t}\right\Vert _{(\infty,2)^{2}}+\frac{k-1}{2^{k-1}}\sqrt{\frac{\log^{2}(m+n)}{\hat{p}}}\left\Vert D_{0}^{t}\right\Vert _{\infty}.
\end{align*}
It follows that 
\begin{align}
\sum_{k=1}^{k_{0}}\frac{\log(m+n)}{p'}\left\Vert D_{k-1}^{t}\right\Vert _{\infty}+\sum_{k=1}^{k_{0}}\sqrt{\frac{\log(m+n)}{p'}}\left\Vert D_{k-1}^{t}\right\Vert _{(\infty,2)^{2}}\qquad\qquad\qquad\nonumber \\
\le\frac{6\log^{2}(m+n)}{\hat{p}}\left\Vert D_{0}^{t}\right\Vert _{\infty}+2\sqrt{\frac{\log^{2}(m+n)}{\hat{p}}}\left\Vert D_{0}^{t}\right\Vert _{(\infty,2)^{2}}.\label{eq:e3}
\end{align}
Combining~(\ref{eq:Ysum})--(\ref{eq:e3}), we obtain
\begin{align*}
\left\Vert \P_{\T_{0}^{\bot}}Y_{k_{0}}\right\Vert  & \le\frac{90\log^{2}(m+n)}{\hat{p}}\left(\left\Vert D_{0}^{U}\right\Vert _{\infty}+\left\Vert D_{0}^{V}\right\Vert _{\infty}\right)+16\sqrt{\frac{\log^{2}(m+n)}{\hat{p}}}\left(\left\Vert D_{0}^{U}\right\Vert _{(\infty,2)^{2}}+\left\Vert D_{0}^{V}\right\Vert _{(\infty,2)^{2}}\right).
\end{align*}
The following lemma, proved in the appendix, bounds the norms of $D_{0}^{U}$
and $D_{0}^{V}$ above. The lemma relies on the second part of Assumption~\ref{asm:sampling_2},
which is a consequence of the trimming procedure in Algorithm~\ref{alg:MP}.
\begin{lem}
\label{lem:PV_H}Recall that $\beta:=\frac{\rho}{\hat{p}}$. Under
Assumptions~\ref{asm:incoherence} and~\ref{asm:sampling_2}, we
have
\begin{align*}
\left\Vert D_{0}^{U}\right\Vert _{\infty} & \le\sqrt{\frac{\mu^{2}r^{2}}{mn}},\\
\left\Vert D_{0}^{U}\right\Vert _{\infty,2}\le\sqrt{\frac{\mu r}{n}}, & \quad\left\Vert D_{0}^{U}\right\Vert _{(\infty,2)^{2}}\le\sqrt{\frac{\mu r}{m\wedge n}},\\
\left\Vert D_{0}^{V}\right\Vert _{\infty}\le\left\Vert D_{0}^{V}\right\Vert _{\infty,2} & \le4\lambda^{2}\gamma\mu r\sqrt{\beta\hat{p}n},\\
\left\Vert D_{0}^{V}\right\Vert _{(\infty,2)^{2}}\le\left\Vert D_{0}^{V}\right\Vert _{F} & \le4\lambda^{2}\gamma n\sqrt{\mu r\beta\hat{p}}.
\end{align*}

\end{lem}
\noindent Using this lemma, we conclude that
\begin{align*}
\left\Vert \P_{\T_{0}^{\bot}}Y_{k_{0}}\right\Vert \le & \frac{90\mu r\log^{2}(m+n)}{\hat{p}\sqrt{mn}}+90\cdot4\lambda^{2}\gamma\mu r\sqrt{\frac{\beta n}{\hat{p}}}\log^{2}(m+n)+16\sqrt{\frac{\mu r\log^{2}(m+n)}{\hat{p}(m\wedge n)}}\\
 & \quad+48\lambda^{2}\gamma n\sqrt{\mu r\beta}\log(m+n).
\end{align*}
One checks that each term above is bounded by $\frac{1}{32}$ under
the conditions~(\ref{eq:p_cond_eqv}) and~(\ref{eq:lambda_cond_eqv}).
 This means that $\left\Vert \P_{\T_{0}^{\bot}}Y_{k_{0}}\right\Vert \le\frac{1}{8}$,
proving the condition~$3(c)$ in Proposition~\ref{prop:opt_cond}.

\subsubsection{Condition $3(f)$}

We need to show $\left\Vert \PIOc\bar{Q}\right\Vert _{\infty,2}=\left\Vert Y_{k_{0}}\right\Vert _{\infty,2}\le\frac{\lambda}{2}$.
By~(\ref{eq:Ysum}), we have
\begin{align}
 & \left\Vert Y_{k_{0}}\right\Vert _{\infty,2}\nonumber \\
\le & \underbrace{\sum_{k=1}^{k_{0}}\left\Vert \left(\frac{1}{p'}\P_{\tilde{\Omega}_{k}}-\mc I\right)D_{k-1}^{U}\right\Vert _{\infty,2}+\sum_{k=1}^{k_{0}}\left\Vert D_{k-1}^{U}\right\Vert _{\infty,2}}_{S_{1}}+\underbrace{\sum_{k=1}^{k_{0}}\left\Vert \left(\frac{1}{p'}\P_{\tilde{\Omega}_{k}}-\mc I\right)D_{k-1}^{V}\right\Vert _{\infty,2}+\sum_{k=1}^{k_{0}}\left\Vert D_{k-1}^{V}\right\Vert _{\infty,2}}_{S_{2}}.\label{eq:f1}
\end{align}
It suffices to bound each of $S_{1}$ and $S_{2}$ by $\frac{\lambda}{4}$.
We need the following lemma, which is proved in the appendix.
\begin{lem}
\label{lem:inf_2_order_1}
For any fixed matrix $Z\in\T_{0}$, we have w.h.p., 
\begin{align*}
\left\Vert \frac{1}{\hat{p}}\mathcal{P}_{\tilde{\Omega}}Z-Z\right\Vert _{\infty,2} & \le\frac{20\log(m+n)}{\hat{p}}\left\Vert Z\right\Vert _{\infty}+\sqrt{\frac{50\log(m+n)}{\hat{p}}}\left\Vert Z\right\Vert _{\infty,2}.
\end{align*}

\end{lem}
Using the lemma with $\tilde{\Omega}$ replaced by $\tilde{\Omega}_{k}$,
we have w.h.p.\ 
\begin{align*}
S_{1} & \le\sum_{k=1}^{k_{0}}\frac{20\log(m+n)}{p'}\left\Vert D_{k-1}^{U}\right\Vert _{\infty}+2\sum_{k=1}^{k_{0}}\sqrt{\frac{50\log(m+n)}{p'}}\left\Vert D_{k-1}^{U}\right\Vert _{\infty,2}.
\end{align*}
Thanks to the second part of Lemma~\ref{lem:inf_2}, we know that~(\ref{eq:e3})
holds with $\left\Vert \cdot\right\Vert _{(\infty,2)^{2}}$ replaced
by $\left\Vert \cdot\right\Vert _{\infty,2}$. Using this, we obtain
that w.h.p.
\[
S_{1}\le\frac{120\log^{2}(m+n)}{\hat{p}}\left\Vert D_{0}^{U}\right\Vert _{\infty}+4\sqrt{\frac{50\log^{2}(m+n)}{\hat{p}}}\left\Vert D_{0}^{U}\right\Vert _{\infty,2}\le\frac{120\mu r\log^{2}(m+n)}{\hat{p}\sqrt{mn}}+4\sqrt{\frac{50\mu r\log^{2}(m+n)}{\hat{p}n}},
\]
where the last inequality follows from Lemma~\ref{lem:PV_H}. The
last R.H.S. is no more than $\frac{\lambda}{4}$ under the conditions~(\ref{eq:p_cond_eqv})
and~(\ref{eq:lambda_cond_eqv}). 

Turning to the term $S_{2}$ in~(\ref{eq:f1}), we apply Lemma~\ref{lem:inf_2_order_1}
with $\tilde{\Omega}$ replaced by $\tilde{\Omega}_{k}$ to obtain
that w.h.p., 
\begin{align*}
S_{2} & \le\sum_{k=1}^{k_{0}}\frac{20\log(m+n)}{p'}\left\Vert D_{k-1}^{V}\right\Vert _{\infty}+2\sum_{k=1}^{k_{0}}\sqrt{\frac{50\log(m+n)}{p'}}\left\Vert D_{k-1}^{V}\right\Vert _{\infty,2}.
\end{align*}
Since~(\ref{eq:e3}) holds with $\left\Vert \cdot\right\Vert _{(\infty,2)^{2}}$
replaced by $\left\Vert \cdot\right\Vert _{\infty,2}$, we obtain
that w.h.p.,
\[
S_{2}\le\frac{120\log^{2}(m+n)}{\hat{p}}\left\Vert D_{0}^{V}\right\Vert _{\infty}+4\sqrt{\frac{50\log^{2}(m+n)}{\hat{p}}}\left\Vert D_{0}^{V}\right\Vert _{\infty,2}.
\]
It then follows from Lemma~\ref{lem:PV_H} that w.h.p.,
\[
S_{2}\le480\lambda^{2}\gamma\mu r\sqrt{\frac{\beta n}{\hat{p}}}\log^{2}(m+n)+16\lambda^{2}\gamma\mu r\sqrt{50\beta n\log^{2}(m+n)}\le600\lambda^{2}\gamma\mu r\sqrt{\frac{\beta n}{\hat{p}}}\log^{2}(m+n).
\]
The last R.H.S. is bounded by $\frac{\lambda}{4}$ w.h.p.\ under the
conditions~(\ref{eq:p_cond_eqv}) and~(\ref{eq:lambda_cond_eqv}).
 This establishes the condition 3(f) in Proposition~\ref{prop:opt_cond}.
Finally, note that each random event above holds w.h.p., so by the
union bound they hold simultaneously with probability at least $1-20(m+n)^{-5}$.
This completes the proof of Theorem~\ref{thm:main}.

\section{Proof of Theorem~\ref{thm:converse}\label{sec:proof_inverse}}

We consider the two conditions~(\ref{eq:p_necessary}) and~(\ref{eq:gamma_necessary})
separately.

\subsection{Condition~(\ref{eq:p_necessary}): $p\le\frac{\mu r\log(2n)}{2n}$\label{sec:case1}}

In this case we use a modified argument from~\cite[Theorem 1.7]{candes2010NearOptimalMC}
to establish the impossibility of determining the \emph{column space}
(i.e., the left singular vectors). We may assume $n_{c}=0$.

\begin{figure}

\begin{centering}
\includegraphics[scale=0.35]{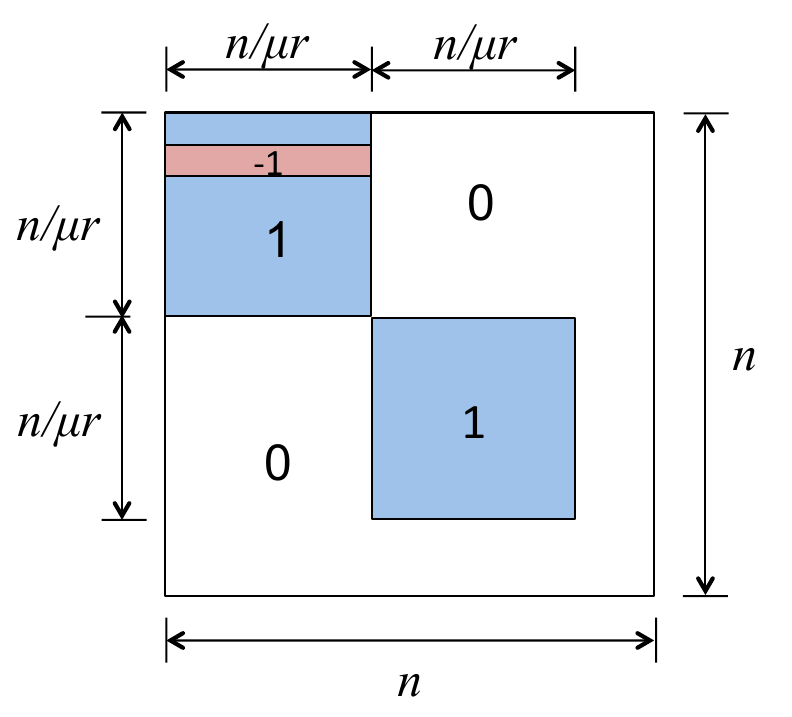}
\par\end{centering}

\protect\caption{\label{fig:converse_p}An illustration of $L$ constructed in Section~\ref{sec:case1}
with rank $r=2$.}
\end{figure}

Without loss of generality, assume that $s:=\frac{n}{\mu r}$ is an
integer. We use $e_{i}$ to denote the $i$-th standard basis whose
dimension will become clear in context. For $k\in[r]$, define the
set 
\begin{equation}
B_{k}=\left\{ (k-1)s+1,(k-1)s+2,\ldots,ks\right\} .\label{eq:B}
\end{equation}
Consider the matrix $L=\sum_{k=1}^{r}u_{k}v_{k}^{\top}\in\RR^{n},$
where the (unnormalized) singular vectors $u_{k}\in\mathbb{R}^{n}$
and $v_{k}\in\mathbb{R}^{n}$ are given by
\[
u_{k}=\sum_{i\in B_{k}}\omega_{i}e_{i},\quad v_{k}=\sum_{k\in B_{k}}e_{i},
\]
where the $\omega_{i}$'s take values in $\left\{ -1,1\right\} $.
Clearly, $L$ has rank-$r$ and incoherence parameter $\mu$, and
is a block diagonal matrix with $r$ blocks of size $s\times s$.
In particular, each row of a block is either all $1$ or $-1$ with
its sign determined by $\omega_{i}$. An illustration of $L$ is given
in Figure~\ref{fig:converse_p}. Therefore, in order to uniquely
determine the left singular vectors $u_{k}$ from the observed entries
of $L$, we must be on event that there is at least one observed entry
on every row $i$ of each diagonal block, since otherwise there would
be no information on $w_{i}$. Under the Bernoulli sampling model
in Assumption \ref{asm:sampling}, the probability of this event is
$\pi=\left[1-\left(1-p\right)^{s}\right]^{n}.$ Using the premise
$2p\le\frac{\log(2n)}{s}\le1$ of the theorem and the inequality $1-x+x^{2}/2>e^{-x},\forall x\ge0$,
we have 
\[
1-p\ge1-\frac{\log(2n)}{2s}\ge1-\frac{\log(2n)}{s}+\frac{\log^{2}(2n)}{2s^{2}}>e^{-\log(2n)/s}.
\]
It follows that 
\[
\pi\le\exp\left[-n(1-p)^{s}\right]\le\exp\left[-ne^{-\log2n}\right]=\exp(-1/2)\le\frac{3}{4},
\]
where the first inequality follows from $1-x\le e^{-x}$. Therefore,
with probability $1-\pi\ge\frac{1}{4}$, there exists one row of a
diagonal block that is unobserved, in which case the $u_{k}$'s cannot
be determined. It is easy to see that this implies the conclusion of the theorem.

\subsection{Condition~(\ref{eq:gamma_necessary}): $\gamma\ge\frac{2p}{\mu r}$\label{sec:case2}}

W.L.O.G. we assume $ps$ is a positive integer. Under the above condition,
we have $n_{c}=\gamma n\ge2ps>ps$, where $s:=\frac{n}{\mu r}$ as
before. We prove the theorem by constructing a family of candidate
solutions $\left(L_{i},C_{i}\right)$, $i=1,2,\ldots,2M$ and showing
it is difficult to accurately distinguish them based on the observed
data $\Omega$ and $\P_{\Omega}\left(L_{i}+C_{i}\right)$. In this
subsection, we use capital letters ($B_{1}$, $I_{2}$, $J$, etc.)
to denote sets of \emph{column} indices (i.e., subsets of $[n+n_{c}]$),
and Greek letters ($\Omega,\Theta$, $\xi$ etc.) to denote sets of
\emph{entry} indices (i.e., subsets of $[n]\times[n+n_{c}]$).

\begin{figure}
\begin{centering}
\includegraphics[scale=0.35]{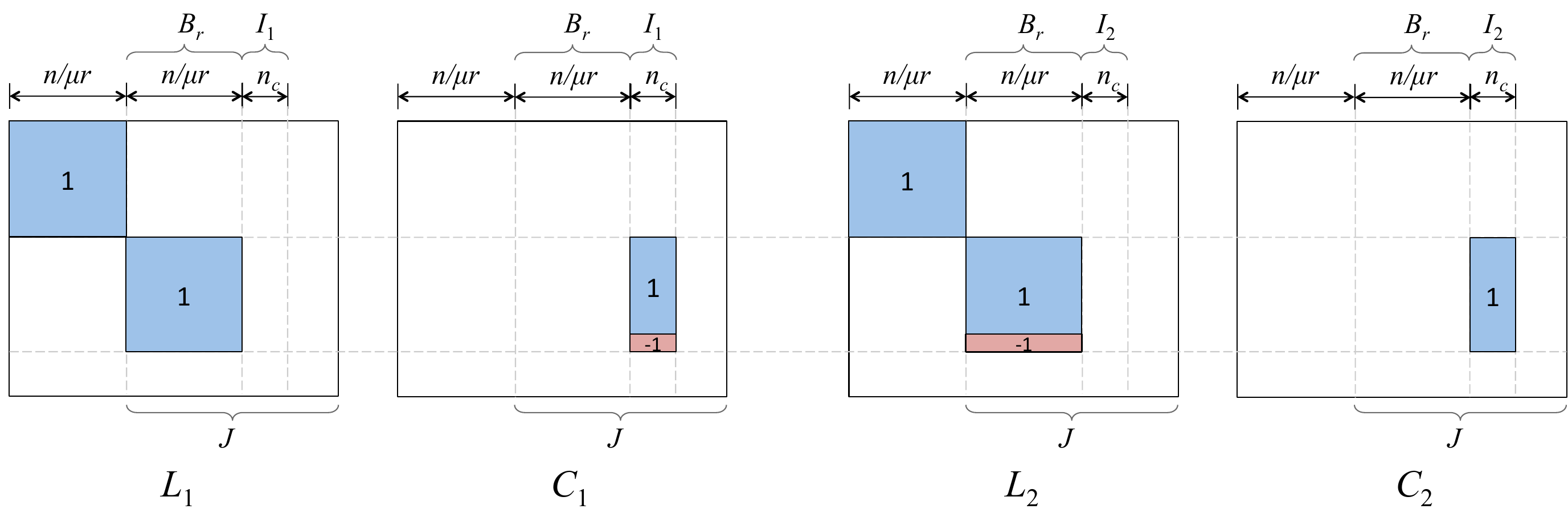}
\par\end{centering}

\protect\caption{\label{fig:converse_gamma}An illustration of $\left(L_{1},C_{1}\right)$
and $\left(L_{2},C_{2}\right)$ constructed in Section~\ref{sec:case2}
with rank $r=2$.}
\end{figure}

Let $J:=\{(r-1)s+1,\ldots,rs+n_{c}\}$. Recall the definition of the
$B_{k}$'s in~(\ref{eq:B}), which satisfies $B_{r}\subseteq J$.
We further let $I:=J\backslash B_{r}$ and 
\begin{align*}
u_{k} & =\sum_{i\in B_{k}}e_{i},k\in[r];\quad\bar{u}_{r}=-e_{rs}+\sum_{i\in B_{r},i\ne rs}e_{i};\\
v_{k} & =\sum_{i\in B_{k}}e_{i},k\in[r];\quad w=\sum_{i\in I}e_{i}.
\end{align*}
We build two candidate solutions $\left(L_{1},C_{1}\right)$ and $\left(L_{2},C_{2}\right)$
as follows:
\[
L_{1}=\sum_{k=1}^{r}u_{k}v_{k}^{\top},\quad C_{1}=\bar{u}_{r}w^\top
\qquad L_{2}=\sum_{k=1}^{r-1}u_{k}v_{k}^{\top}+\bar{u}_{r}v_{r}^{\top},\quad C_{2}=u_{r}w^\top.
\]
We illustrate them in Figure~\ref{fig:converse_gamma}. 
 Let $M:={s+n_{c} \choose s}$. In the definition of the $\left(L_{1},C_{1}\right)$,
if we let the set $B_{k}$ vary in all $M$ possible subsets of $J$
with size $s$ (i.e., we permute the columns in $J$), then we get
$M$ different candidates $\left(L_{i},C_{i}\right),i=1,3,\ldots,2M-1$.
Similarly, by varying $B_{k}$ in$\left(L_{2},C_{2}\right)$ we can
get another $M$ candidates $\left(L_{i},C_{i}\right),i=2,4,\ldots,2M$.
We thus have defined a family of $2M$ pairs. Let $I_{i}:=\text{column-support}(C_{i}).$
Note that for the $L_{i}$'s, only the locations of the last $s$
authentic columns vary in $J$, and the sign of these columns' $rs$-th
row changes. The corrupted columns in $C_{i}$ are identical to the
last $s$ authentic columns of $L_{i}$ except with the sign of the
$rs$-th row flipped. Therefore, to recover the column space of $L_{i}$,
one needs to determine the sign of the $rs$-th row. The idea of the
proof is simple: under the Bernoulli model and with $n_{c}>2ps$ columns
in $I_{i}$, with positive probability the $rs$-row has roughly as
many observed $1$'s as $-1$'s, so there is no way to determine which
sign is authentic. 

We make this precise by specifying the set of observed entries $\Omega=\tilde{\Omega}_{i}\cup\Omega_{c,i}$,
for each candidate $i\in[2M]$. According to our assumption, the observations
$\tilde{\Omega}_{i}$ on the authentic columns follow the Bernoulli
model with uniform probability $p$. It remains to specify the observations
$\Omega_{c,i}$ on the corrupted columns. Recall Definition~\ref{def:bernoulli} of the Bernoulli model, and let $\Omega_{c,i}^{+}$
be drawn from the Bernoulli model on $[rs-1]\times I_{i}$ with uniform
probability $p$; this will be the observed entries on the first $rs-1$
rows of the corrupted columns. Let $\Gamma_{i}$ be independent from
$\tilde{\Omega}_{i}$ and drawn according to the Bernoulli model on
$[s]$ with uniform probability $p$ . If $\left|\Gamma_{i}\right|\ge n_{c}$,
then $\Omega_{c,i}^{-}$, the set of observed entries on the $rs$-th
row of the corrupted columns, is set as $\Omega_{c,i}^{-}=\left\{ rs\right\} \times I_{i}$.
If $\left|\Gamma_{i}\right|=t<n_{c}$, then we set $\Omega_{c_{i}}^{-}=\left\{ rs\right\} \times I_{i}\left(t\right)$,
where $I_{i}(t)$ denotes the $t$ smallest indices in $I_{i}$. The
set of observed entries on the corrupted columns $I_{i}$ is then
given by $\Omega_{c,i}=\Omega_{c,i}^{+}\cup\Omega_{c,i}^{-}$. We
see that the authentic observations $\tilde{\Omega}_{i}$ are independent
of $C_{i}$ and $\Omega_{c,i}$, so Assumption~\ref{asm:sampling}
is satisfied. In the sequel, we use $\mathbb{P}_{L_{i},C_{i}}$ to
denote the probability computed under the $i$-th candidate solution
$\left(L_{i},C_{i}\right)$.

\begin{figure}
\begin{centering}
\includegraphics[scale=0.35]{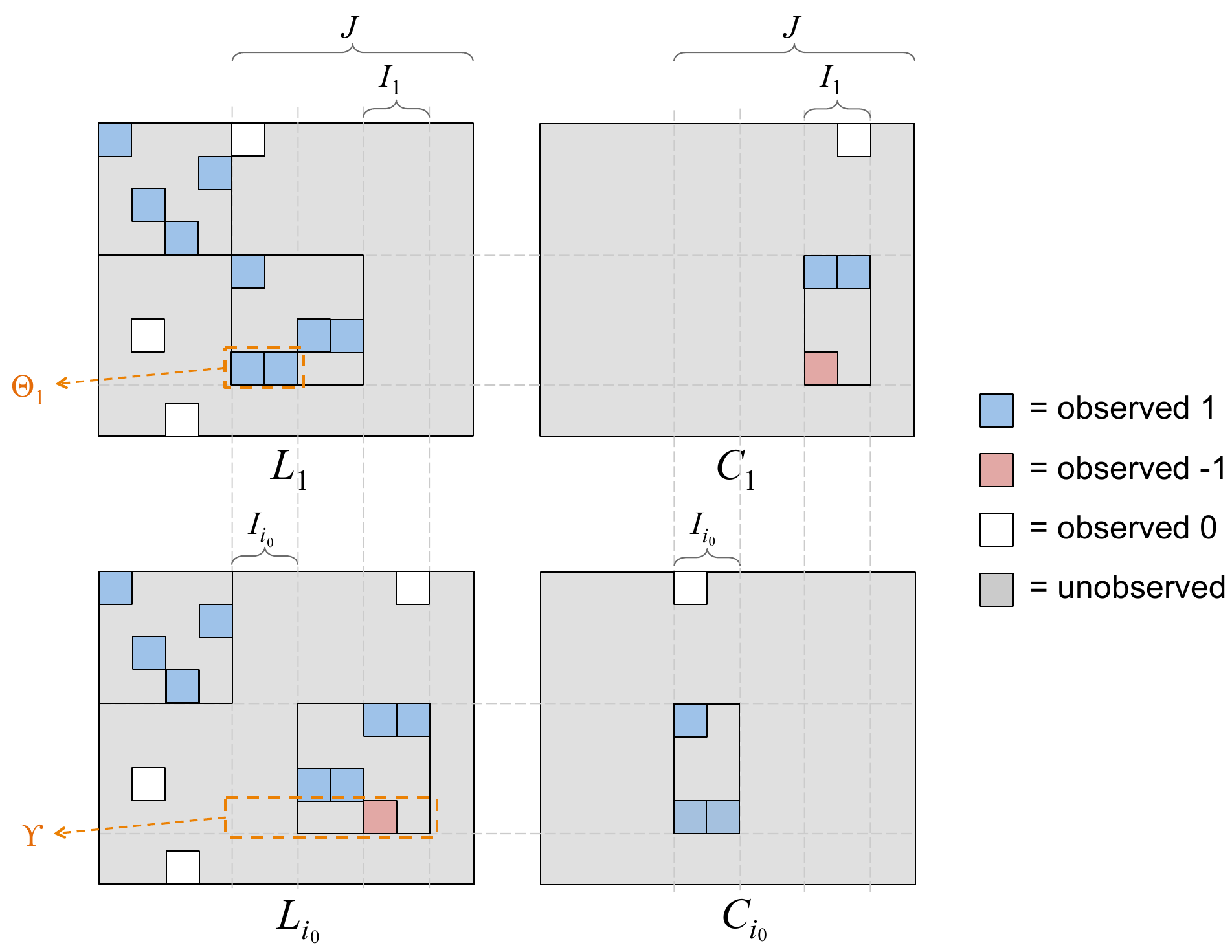}
\par\end{centering}

\protect\caption{\label{fig:converse_gamma_omega}An illustration of the two solutions
$\left(L_{1},C_{1}\right)$, $\left(L_{i_{0}},C_{i_{0}}\right)$ and
the locations of the observed entries $\Omega$ in Section~\ref{sec:case2},
where $\left|\Theta_{1}\right|\le n_{c}$. In this case the two solutions
generate the same observed data $\protect\P_{\Omega}\left(L_{1}+C_{1}\right)=\protect\P_{\Omega}\left(L_{i_{0}}+L_{i_{0}}\right)$
and it is impossible to distinguish between them.}
\end{figure}

Now suppose the true solution is the first candidate $\left(L_{1},C_{1}\right)$.
Let $\Theta_{1}:=\tilde{\Omega}_{1}\cap(\left\{ rs\right\} \times J)$
be the set of observations on the $rs$-th row of the authentic columns
in $J$. If we define the event 
\[
\mathcal{E}:=\left\{ \left|\Gamma_{1}\right|\le\left|\Theta_{1}\right|\le ps\right\} ,
\]
then we have 
\begin{align*}
\mathbb{P}_{L_{1},C_{1}}\left[\mathcal{E}\right] & \overset{(i)}{\ge}\frac{1}{2}\mathbb{P}_{L_{1},C_{1}}\left[\left|\Gamma_{1}\right|<ps\text{ and }\left|\Theta_{1}\right|<ps\right]\\
 & \overset{(ii)}{\ge}\frac{1}{2}\cdot\mathbb{P}_{L_{1},C_{1}}\left[\left|\Gamma_{1}\right|<ps\right]\cdot\mathbb{P}_{L_{1},C_{1}}\left[\left|\Theta_{1}\right|<ps\right]\overset{(iii)}{\ge}\frac{1}{8}
\end{align*}
here $(i)$ follows from symmetry, and $(ii)$--$(iii)$ hold because
$\left|\Theta_{1}\right|$ and $\left|\Gamma_{1}\right|$ are independent
and both follow the Binomial distribution with $s$ trials and probability
$p$, whose median is $ps$. On this event $\mathcal{E}$, we can
always find another candidate solution $i_{0}\in\left\{ 2,4,\ldots,2M\right\} $
(which means the last row of $L_{i}$ has a negative sign so the column
space is different) such that $\Theta_{1}=\left\{ rs\right\} \times I_{i_{0}}\left(\left|\Theta_{1}\right|\right)$;
this is because $\Theta_{1}\subseteq\left\{ rs\right\} \times J$
and the $I_{i}$'s enumerates the subsets of $J$ with size $n_{c}>ps\ge\left|\Theta_{1}\right|$.
See Figure~\ref{fig:converse_gamma_omega} for an illustration. Let
$\omega\subseteq[n]\times[n+n_{c}]$ be a realization of $\Omega$
that is consistent with $\mathcal{E}$, i.e., it satisfies $\mathbb{P}_{L_{1},C_{1}}\left[\Omega=\omega\text{ and }\mathcal{E}\right]>0$.
We claim that (proved below) for any such $\omega$, we have 
\begin{align*}
\mathbb{P}_{L_{1},C_{1}}\left[\Omega=\omega\right] & \le\mathbb{P}_{L_{i_{0}},C_{i_{0}}}\left[\Omega=\omega\right],
\end{align*}
and
\begin{align*}
\P_{\Omega}\left(L_{1}+C_{1}\right) & =Z:=\mathcal{P}_{\Omega}\left(L_{i_{0}}+C_{i_{0}}\right)\quad\quad\text{for } \Omega = \omega.
\end{align*}
This means the observed data is identical under both candidate solutions,
but the $i_{0}$-th candidate has a higher likelihood. In this case,
the maximum likelihood estimator (MLE), which is given by 
\[
f\left(\omega,Z\right):=\arg\max_{(L_{i},C_{i})}\mathbb{P}_{L_{i},C_{i}}\left[\Omega=\omega,\mathcal{P}_{\Omega}\left(L_{i_{0}}+C_{i_{0}}\right)=Z\right]
\]
will incorrectly output a solution other than $(L_{1},C_{1})$ with probability at least $\frac{1}{2}$.
The above argument in fact holds if any one of the $\left(L_{i},C_{i}\right)$'s
is the true solution. Therefore, the average probability of error
for the MLE is at least $\frac{1}{2}\cdot\mathbb{P}_{L_{1},C_{1}}\left[\mathcal{E}\right]\ge\frac{1}{16}$.
Since the MLE minimizes the average probability of error, which in turn lower bounds the worst case error probability, we conclude
that any estimator makes an error with worst case probability at least $\frac{1}{16}$.
This proves the theorem.

\textbf{Proof of the claim: }
When $ \Omega = \omega $, the equality $\P_{\Omega}\left(L_{1}+C_{1}\right)=\mathcal{P}_{\Omega}\left(L_{i_{0}}+C_{i_{0}}\right)$
holds by construction of the $\left(L_{i},C_{i}\right)$'s and the
assumption on $\omega$ (cf.\ Figure~\ref{fig:converse_gamma_omega}).
To prove the inequality, we note the distribution of $\Omega$ under
$\left(L_{1},C_{1}\right)$ and $\left(L_{i_{0}},C_{i_{0}}\right)$
only differs on the entries in $\Upsilon:=\left\{ rs\right\} \times J$.
Let $\xi:=\omega\cap\left(\left\{ rs\right\} \times I_{i_{0}}\right)$
and $\zeta:=\omega\cap\left(\left\{ rs\right\} \times I_{1}\right).$
Note that because $\omega$ is consistent with $\mathcal{E}$, we
have $\left|\zeta\right|\le\left|\xi\right|\le ps<n_{c}$; moreover,
the observed entries in $\Upsilon$ are either on the columns $I_{i}$
or $I_{i_{0}}$, so $\omega\cap\Upsilon=\xi\cup\zeta$. Let $g(\cdot)$
denote the probability mass function of the Binomial distribution
with $s$ trials and probability $p$. Then, according to our specification
of $\Omega$ under each candidate solution, we have
\begin{align*}
\frac{\mathbb{P}_{L_{1},C_{1}}\left[\Omega=\omega\right]}{\mathbb{P}_{L_{i_{0}},C_{i_{0}}}\left[\Omega=\omega\right]} & =\frac{\mathbb{P}_{L_{1},C_{1}}\left[\Omega\cap\Upsilon=\omega\cap\Upsilon\right]}{\mathbb{P}_{L_{i_{0}},C_{i_{0}}}\left[\Omega\cap\Upsilon=\omega\cap\Upsilon\right]}\\
 & =\frac{p^{\left|\xi\right|}(1-p)^{s-\left|\xi\right|}g(\left|\zeta\right|)}{g(\left|\xi\right|)p^{\left|\zeta\right|}(1-p)^{s-\left|\zeta\right|}}=\frac{g(\left|\zeta\right|)}{g(\left|\xi\right|)}\cdot\left(\frac{p}{1-p}\right)^{\left|\xi\right|-\left|\zeta\right|}.
\end{align*}
Observe that $g(\cdot)$ is unimodal with mode $ps$, and $\left|\zeta\right|\le\left|\xi\right|\le ps$,
so $g(\left|\zeta\right|)\le g(\left|\xi\right|)$. Moreover, we have
$\left(\frac{p}{1-p}\right)^{\left|\xi\right|-\left|\zeta\right|}\le1$
by the assumption $p\le\frac{1}{2}$. This means 
\[
\frac{\mathbb{P}_{L_{1},C_{1}}\left[\Omega=\omega\right]}{\mathbb{P}_{L_{i_{0}},C_{i_{0}}}\left[\Omega=\omega\right]}\le1,
\]
proving the claim.

\section{Conclusion\label{sec:conclusion}}

In this paper, we study the problem of completing a low-rank matrix
from sparsely observed entries when observations from some columns
are completely and arbitrarily corrupted. We propose a new algorithm
based on trimming and convex optimization, and provide performance
guarantees showing its robustness to column-wise corruption. We further
show that the performance of our algorithm is close to the information-theoretic
limit under adversarial corruption, thus achieving near-optimal tradeoffs
between sample complexity, robustness and rank. 

Immediate future directions include removing the sub-optimality in
bounds and allowing for noise and sparse corruption. It may be possible
to further improve the robustness of matrix completion by combining
our approach with other outlier detection techniques~\cite{Huber1981Robstat,maronna2006robuststat}.
As our work is motivated by the practical applications in collaborative
filtering and crowdsourcing, it is important to study in more depth
the computational aspects and develop fast online/parallel algorithms.
A more systematic exploration of the relation between sample complexity,
model complexity, computational complexity and robustness, will also
be of much theoretical and practical interest.

\section*{Acknowledgment}

We are grateful to the anonymous reviewers for their helpful suggestions on improving the quality of the manuscript. Y. Chen  was supported by NSF grant CIF-31712-23800, ONR MURI grant N00014-11-1-0688, and a start-up fund from the School of Operations Research and Information Engineering at Cornell University. The work of H. Xu was partially supported by the Ministry of Education of Singapore through AcRF Tier Two grants R-265-000-443-112 and R265-000-519-112, and A*STAR SERC PSF grant R-265-000-540-305. C. Caramanis acknowledges NSF grants 1056028, 1302435 and 1116955. His research was also partially supported by the U.S. DoT through the Data-Supported Transportation Operations and Planning (D-STOP) Tier 1 University Transportation Center. S. Sanghavi would like to acknowledge NSF grants 0954059 and 1302435. 

\appendixpage

\appendix

\section{Proof of Lemma~\ref{lem:model}\label{sec:proof_lem_model}}

We need a simple observation first: the convex program~(\ref{eq:L12formulation})
has a monotonicity property, that is, having more observed entries
on the uncorrupted columns only makes the program more likely to succeed.
\begin{lem}
[Monotonicity]\label{lem:monotone}Suppose the indices set $\Omega_{1}$
and $\Omega_{2}$ are such that $\Omega_{1}\cap\left([m]\times I_{0}^{c}\right)\subseteq\Omega_{2}\cap\left([m]\times I_{0}^{c}\right)$
and $\Omega_{1}\cap\left([m]\times I_{0}\right)=\Omega_{2}\cap\left([m]\times I_{0}\right)$.
If the program~(\ref{eq:L12formulation}) with $\hat{\Omega}=\Omega_{1}$
as the input succeeds, then using $\hat{\Omega}=\Omega_{2}$ as the
input also succeeds. \end{lem}
\begin{proof}
Define the set 
\[
\mathfrak{X}:=\left\{ (L,C):\P_{I_{0}^{c}}(L)=L_{0},\P_{U_{0}}(L)=L,\P_{I_{0}}(C)=C,\P_{\Omega_{1}\cap\left([m]\times I_{0}\right)}\left(L+C\right)=\P_{\Omega_{1}\cap\left([m]\times I_{0}\right)}\left(M\right)\right\} ,
\]
which are the solutions that correspond to the success of the algorithm and are consistent
on the entries in $\Omega_{1}\cap\left([m]\times I_{0}\right)=\Omega_{2}\cap\left([m]\times I_{0}\right)$.
Observe that any solution in $\mathfrak{X}$ is feasible to the program
with $\hat{\Omega}$ equal to $\Omega_{1}$ or $\Omega_{2}$. Suppose
$(L^{*'},C^{*'})$ is any optimal solution to the program~(\ref{eq:L12formulation})
with $\Omega_{2}$. By optimality we must have $\left\Vert L^{*'}\right\Vert _{*}+\lambda\left\Vert C^{*'}\right\Vert _{1,2}\le\left\Vert L\right\Vert _{*}+\lambda\left\Vert C\right\Vert _{1,2}$,
$\forall\left(L,C\right)\in\mathfrak{X}$. On the other hand, the
program with $ $$\Omega_{1}$ succeeds by assumption, meaning that
any optimal solution $\left(L^{*},C^{*}\right)$ of it must be in
the set $\mathfrak{X}$. It follows that $(L^{*'},C^{*'})$ has an
objective value lower or equal to $(L^{*},C^{*})$. But $(L^{*'},C^{*'})$
is also feasible to the program with $\Omega_{1}$ since $\Omega_{1}\subseteq\Omega_{2}$,
so $(L^{*'},C^{*'})$ is optimal to the program with $\Omega_{1}$
and hence in the set $\mathfrak{X}$. This means the program with
$\Omega_{2}$ succeeds.
\end{proof}
We turn to the proof of Lemma~\ref{lem:model}. Given a vector $\vec{k}\in\RR^{n}$ with elements $k_{j}$, let $\mathbb{P}_{\text{Unif}(\vec{k})}$
denote the probability when $\tilde{\Omega}$ follows the uniform
model with parameter $\vec{k}$, meaning that the observed entries
on the $j$-th column is sampled uniformly at random without replacement
from all size-$k_{j}$ subsets of the entries in this column. Recall
that $h_{j}$ is the number of observed entries on the $j$-th column
before trimming. We use $\left\lfloor x\right\rfloor $ to denote
the largest integer no more than $x$. We have the following chain
of inequalities:
\begin{align*}
\mathbb{P}_{\text{Ber}(\vec{p})}\left[success\right] & =\sum_{k_{1}=1}^{m}\cdots\sum_{k_{n}=1}^{m}\mathbb{P}_{\text{Ber}(\vec{p})}\left[success\vert h_{j}=k_{j},j\in[n]\right]\mathbb{P}_{\text{Ber}(\vec{p})}\left[h_{j}=k_{j},j\in[n]\right]\\
 & \ge\sum_{k_{1}=\left\lfloor \hat{p}m/2\right\rfloor }^{m}\cdots\sum_{k_{n}=\left\lfloor \hat{p}m/2\right\rfloor }^{m}\mathbb{P}_{\text{Ber}(\vec{p})}\left[success\vert h_{j}=k_{j},j\in[n]\right]\mathbb{P}_{\text{Ber}(\vec{p})}\left[h_{j}=k_{j},j\in[n]\right]\\
 & \overset{(a)}{=}\sum_{k_{1}=\left\lfloor \hat{p}m/2\right\rfloor }^{m}\cdots\sum_{k_{n}=\left\lfloor \hat{p}m/2\right\rfloor }^{m}\mathbb{P}_{\text{Unif}(\vec{k})}\left[success\right]\mathbb{P}_{\text{Ber}(\vec{p})}\left[h_{j}=k_{j},j\in[n]\right]\\
 & \overset{(b)}{=}\sum_{k_{1}=\left\lfloor \hat{p}m/2\right\rfloor }^{m}\cdots\sum_{k_{n}=\left\lfloor \hat{p}m/2\right\rfloor }^{m}\mathbb{P}_{\text{Unif}\left(\vec{k}\wedge\left\lfloor \rho m\right\rfloor \right)}\left[success\right]\mathbb{P}_{\text{Ber}(\vec{p})}\left[h_{j}=k_{j},j\in[n]\right]\\
 & \overset{(c)}{\ge}\sum_{k_{1}=\left\lfloor \hat{p}m/2\right\rfloor }^{m}\cdots\sum_{k_{n}=\left\lfloor \hat{p}m/2\right\rfloor }^{m}\mathbb{P}_{\text{Unif}\left(\left\lfloor \hat{p}m/2\right\rfloor \right)}\left[success\right]\mathbb{P}_{\text{Ber}(\vec{p})}\left[h_{j}=k_{j},j\in[n]\right]\\
 & =\mathbb{P}_{\text{Unif}\left(\left\lfloor \hat{p}m/2\right\rfloor \right)}\left[success\right]\mathbb{P}_{\text{Ber}(\vec{p})}\left[h_{j}\ge\left\lfloor \hat{p}m/2\right\rfloor ,j\in[n]\right]\\
 & \overset{(d)}{\ge}\mathbb{P}_{\text{Unif}\left(\left\lfloor \hat{p}m/2\right\rfloor \right)}\left[success\right]\left(1-(m+n)^{-10}\right),
\end{align*}
where $(a)$ follows from the fact that the conditional distribution
of a set following the Bernoulli model given its cardinality is the
same as sampling uniformly without replacement, $(b)$ is a consequence
of the trimming step in Algorithm~\ref{alg:MP}, as a uniform subset
of a uniformly sampled set is still uniform, $(c)$ follows from $\rho m\ge\hat{p}m/2$
and the monotonicity in Lemma~\ref{lem:monotone}, and finally $(d)$
follows from the Bernstein inequality under the condition~(\ref{eq:p_cond_eqv})
with $c_{1}$ large enough. The probability in $(d)$ can be bounded
by similar reasoning as follows: 
\begin{align*}
 & \mathbb{P}_{\text{Unif}\left(\left\lfloor \hat{p}m/2\right\rfloor \right)}\left[success\right]\\
\ge & \mathbb{P}_{\text{Unif}\left(\left\lfloor \hat{p}m/2\right\rfloor \right)}\left[success\right]\sum_{k_{1}=1}^{\left\lfloor \hat{p}m/2\right\rfloor }\cdots\sum_{k_{n}=1}^{\left\lfloor \hat{p}m/2\right\rfloor }\mathbb{P}_{\text{UBer}(\hat{p}/4)}\left[h_{j}=k_{j},j\in[n]\right]\\
\overset{(a)}{\ge} & \sum_{k_{1}=1}^{\left\lfloor \hat{p}m/2\right\rfloor }\cdots\sum_{k_{n}=1}^{\left\lfloor \hat{p}m/2\right\rfloor }\mathbb{P}_{\text{Unif}(\vec{k})}\left[success\right]\mathbb{P}_{\text{UBer}(\hat{p}/4)}\left[h_{j}=k_{j},j\in[n]\right]\\
\overset{(b)}{=} & \sum_{k_{1}=1}^{\left\lfloor \hat{p}m/2\right\rfloor }\cdots\sum_{k_{n}=1}^{\left\lfloor \hat{p}m/2\right\rfloor }\mathbb{P}_{\text{UBer}(\hat{p}/4)}\left[success\vert h_{j}=k_{j},j\in[n]\right]\mathbb{P}_{\text{UBer}(\hat{p}/4)}\left[h_{j}=k_{j},j\in[n]\right]\\
= & \sum_{k_{1}=1}^{\left\lfloor \hat{p}m/2\right\rfloor }\cdots\sum_{k_{n}=1}^{\left\lfloor \hat{p}m/2\right\rfloor }\mathbb{P}_{\text{UBer}(\hat{p}/4)}\left[success,h_{j}=k_{j},j\in[n]\right]\\
= & \mathbb{P}_{\text{UBer}(\hat{p}/4)}\left[success\right]-\sum_{\vec{k}:\exists j\in[n],k_{j}>\left\lfloor \hat{p}m/2\right\rfloor }\mathbb{P}_{\text{UBer}(\hat{p}/4)}\left[success,h_{j}=k_{j},j\in[n]\right]\\
\ge & \mathbb{P}_{\text{UBer}(\hat{p}/4)}\left[success\right]-\mathbb{P}_{\text{UBer}(\hat{p}/4)}\left[\exists j\in[n],h_{j}>\left\lfloor p'm/2\right\rfloor \right]\\
\overset{(c)}{\ge} & \mathbb{P}_{\text{UBer}(\hat{p}/4)}\left[success\right]-(m+n)^{-10},
\end{align*}
where $(a)$ follows from the monotonicity Lemma~\ref{lem:monotone},
$(b)$ follows from the fact that conditional Bernoulli distribution
is uniform, and $(c)$ follows from the Bernstein inequality under
the condition~(\ref{eq:p_cond_eqv}). Combining pieces, we obtain
\[
\mathbb{P}_{\text{Ber}(\vec{p})}\left[success\right]
\ge\left(1-(m+n)^{-10}\right)\left(\mathbb{P}_{\text{UBer}(\hat{p}/4)}\left[success\right]-(m+n)^{-10}\right).
\]
The lemma follows.

\section{Proof of Lemmas in Section~\ref{sec:primal_contruct}}

In this section, we prove the lemmas used in Section~\ref{sec:primal_contruct}.

\subsection{Proof of Lemma~\ref{lem:TandT0}}

Let $\textrm{col}(Z)$ denote the column space of a matrix $Z$. Observe
that $\P_{U_{0}}\bar{L}=\bar{L}$ implies $\textrm{col}(\bar{L})\subseteq\textrm{col}(U_{0})$,
and $\PIOc(\bar{L})=L_{0}$ implies $\textrm{col}(\bar{L})\supseteq\textrm{col}(U_{0})$.
It follows that $\textrm{col}(\bar{U})=\textrm{col}(\bar{L})=\textrm{col}(U_{0})$.
Because $\bar{C}$ satisfies the last constraint in the oracle problem~(\ref{eq:oracle}),
we have $\bar{I}\in I_{0}$. This proves part (a) of the lemma. A
consequence is that  $\textrm{rank}(\bar{L})=\textrm{rank}(L_{0})=r$. 

Since $\PIOc\bar{L}=L_{0}$, we conclude that the matrix $\bar{V}_{c}^{\top}:=\PIOc\bar{V}^{\top}$
has the same rank-$r$ row space as $V_{0}^{\top}$ . Therefore, $\bar{V}_{c}^{\top}\bar{V}_{c}\in\RR^{r\times r}$
is positive definite and there exists a symmetric and invertible matrix
$K_{1}\in\RR^{r\times r}$ with $K_{1}^{2}=\bar{V}_{c}^{\top}\bar{V}_{c}$
and $\left\Vert K_{1}\right\Vert \le\left\Vert \bar{V}_{c}\right\Vert \le\left\Vert \bar{V}\right\Vert \le1$.
This implies that $K_{1}^{-1}\bar{V}_{c}^{\top}$ has orthonormal
rows spanning the same row space as $V_{0}^{\top}$. Because $V_{0}^{\top}$
also has orthonormal rows, there must exist an orthonormal matrix
$K_{2}\in\RR^{r\times r}$ such that $K_{2}K_{1}^{-1}\bar{V}_{c}^{\top}=V_{0}^{\top}$.
Hence we have $\bar{V}_{c}^{\top}=NV_{0}^{\top},$ where the matrix
$N:=K_{2}^{-1}K_{1}\in\mathbb{R}^{r\times r}$ is invertible. It follows
that
\[
\max_{1\le j\le n+n_{c}}\left\Vert \left(\PIOc\bar{V}^{\top}\right)e_{j}\right\Vert _{2}^{2}=\max_{j}\left\Vert K_{2}^{-1}K_{1}V_{0}^{\top}e_{j}\right\Vert _{2}^{2}\le\left\Vert K_{2}^{-1}\right\Vert ^{2}\left\Vert K_{1}\right\Vert ^{2}\max_{j}\left\Vert V_{0}^{\top}e_{j}\right\Vert _{2}^{2}\le\frac{\mu r}{n},
\]
where in the last inequality we use the incoherence of $L_{0}$ in
Assumption~\ref{asm:incoherence}. This proves part $(b)$.

Now consider part $(c)$. Let $Z$ be an arbitrary matrix in $\RR^{m\times(n+n_{c})}$.
By part (a) of the lemma, we have $\P_{U_{0}}\P_{\bar{U}}\left(\PIOc Z\right)=\P_{\bar{U}}\left(\PIOc Z\right)$.
We also have 
\[
\PIOc\P_{\bar{U}^{\bot}}\P_{\bar{V}}Z=\left(\P_{\bar{U}^{\bot}}Z\right)\bar{V}\PIOc\left(\bar{V}^{\top}\right)=\left(\P_{\bar{U}^{\bot}}Z\right)\bar{V}\bar{V}_{c}^{\top},
\]
where the  R.H.S. spans the same row space as $V_{0}^{\top}$ by the
discussion in the last paragraph. It follows that
\begin{align*}
\PTO\PIOc\PT Z & =\PTO\P_{\bar{U}}\PIOc Z+\PTO\PIOc\P_{\bar{U}^{\bot}}\P_{\bar{V}}Z=\P_{\bar{U}}\PIOc Z+\PIOc\P_{\bar{U}^{\bot}}\P_{\bar{V}}Z=\PIOc\PT Z.
\end{align*}

For part $(d)$, the previous discussion shows that $\bar{V}_{c}=V_{0}N^{\top}$.
Therefore, for any $Y\in\RR^{m\times(n+n_{c})}$, we have 
\[
\left(\PIOc Y\right)V_{0}V_{0}^{\top}\bar{V}\bar{V}^{\top}=\left(\PIOc Y\right)V_{0}V_{0}^{\top}\bar{V}_{c}\bar{V}^{\top}=\left(\PIOc Y\right)V_{0}V_{0}^{\top}V_{0}N^{\top}\bar{V}^{\top}=\left(\PIOc Y\right)\bar{V}_{c}\bar{V}^{\top}=\left(\PIOc Y\right)\bar{V}\bar{V}^{\top}.
\]
Applying this equality with $Y=\P_{\bar{U}^{\bot}}Z$, we obtain
\begin{align*}
\PT\PTO\PIOc Z & =\P_{\bar{U}}\left(\PIOc Z\right)+\left(\PIOc\left(\mc I-\P_{U_{0}}\right)Z\right)V_{0}V_{0}^{\top}\bar{V}\bar{V}^{\top}=\P_{\bar{U}}\PIOc Z+\left(\PIOc\P_{\bar{U}^{\bot}}Z\right)\bar{V}\bar{V}^{\top}=\PT\PIOc Z.
\end{align*}

Finally, to prove part $(e)$, we note that
\begin{align*}
\P_{\bar{\T}^{\bot}}\P_{\T_{0}^{\bot}}\PIOc Z & =\left(\mc I-\PT\right)\left(\mc I-\PTO\right)\PIOc Z.
\end{align*}
Expanding the last R.H.S and applying part $(d)$ of the lemma gives
the desired result.

\subsection{Proof of Lemma~\ref{lem:H}}

Applying $\P_{I_{0}}$ to both sides of the last equality in~(\ref{eq:oracle_subgrad})
proves part~$(a)$ of the lemma. Part~$(b)$ follows from $G\in\bar{I}^{c}$,
and part~$(c)$ follows from $\P_{\bar{I}^{c}}\bar{H}'=\P_{\bar{I}^{c}}\P_{I_{0}}G$.
Applying the projection $\P_{\bar{U}}\P_{\I_{0}}=\P_{U_{0}}\P_{\I_{0}}$
to both sides of the first equality in~(\ref{eq:oracle_subgrad}),
we obtain part~$(d)$. Finally, note that $\bar{H}$ and $\bar{H}'$
are determined by the oracle program~(\ref{eq:oracle}), which only
depends on $\P_{\Omega_{c}}M=\P_{\Omega_{c}}C_{0}$ and dose not involve
$\tilde{\Omega}$. Therefore, independence between $\tilde{\Omega}$
and $\P_{\Omega_{c}}C_{0}$ imposed in Assumption~\ref{asm:sampling_2}
implies part~$(e)$.

\section{Proof of Proposition~\ref{prop:opt_cond}}

To prove the proposition, we need a technical lemma.
\begin{lem}
\label{lem:Tperp>T-1NonOr}Suppose (\ref{eq:invertibility}) holds,
then for any $\Delta_{l},\Delta_{c}\in\RR^{m\times(n+n_{c})}$ with
$\P_{\Omega}\Delta_{l}+\P_{\Omega}\Delta_{c}=0$, we have
\[
\left\Vert \PIOc\P_{\bar{\T}}\Delta_{l}\right\Vert _{F}\le\sqrt{\frac{2}{\hat{p}}}\left(\left\Vert \P_{\bar{\T}^{\bot}}\Delta_{l}\right\Vert _{*}+\left\Vert \PIOc\Delta_{c}\right\Vert _{1,2}\right).
\]
\end{lem}
\begin{proof}
Since $\P_{\Omega}\Delta_{c}=-\P_{\Omega}\Delta_{l}$, we have 
\begin{align*}
\left\Vert \PIOc\Delta_{c}\right\Vert _{1,2}\ge\left\Vert \P_{\tilde{\Omega}}\Delta_{c}\right\Vert _{F} & =\left\Vert \P_{\tilde{\Omega}}\Delta_{l}\right\Vert _{F}.
\end{align*}
By triangle inequality, we get 
\[
\left\Vert \P_{\tilde{\Omega}}\Delta_{l}\right\Vert _{F}\ge\left\Vert \P_{\tilde{\Omega}}\P_{\bar{\T}}\Delta_{l}\right\Vert _{F}-\left\Vert \P_{\tilde{\Omega}}\P_{\bar{\T}^{\bot}}\Delta_{l}\right\Vert _{F}\ge\left\Vert \P_{\tilde{\Omega}}\P_{\I_{0}^{c}}\P_{\bar{\T}}\Delta_{l}\right\Vert _{F}-\left\Vert \P_{\bar{\T}^{\bot}}\Delta_{l}\right\Vert _{\ast}.
\]
We bound the first term in the last R.H.S.: 
\begin{align*}
\left\Vert \P_{\tilde{\Omega}}\P_{\I_{0}^{c}}\P_{\bar{\T}}\Delta_{l}\right\Vert _{F}^{2} & =\left\langle \P_{\tilde{\Omega}}\P_{\mathcal{I}_{0}^{c}}\P_{\bar{\T}}\Delta_{l},\P_{\tilde{\Omega}}\P_{\mathcal{I}_{0}^{c}}\P_{\bar{\T}}\Delta_{l}\right\rangle \\
 & \overset{(a)}{=}\left\langle \P_{\mathcal{I}_{0}^{c}}\P_{\bar{\T}}\Delta_{l},\PTO\P_{\tilde{\Omega}}\PTO\P_{\mathcal{I}_{0}^{c}}\P_{\bar{\T}}\Delta_{l}\right\rangle \\
 & \overset{(b)}{=}\left\langle \P_{\mathcal{I}_{0}^{c}}\P_{\bar{\T}}\Delta_{l},\left(\PTO\P_{\tilde{\Omega}}\PTO\right)\P_{\mathcal{I}_{0}^{c}}\P_{\bar{\T}}\Delta_{l}-\hat{p}\P_{T_{0}}\P_{\mathcal{I}_{0}^{c}}\P_{\bar{\T}}\Delta_{l}+\hat{p}\P_{\mathcal{I}_{0}^{c}}\P_{\bar{\T}}\Delta_{l}\right\rangle \\
 & \overset{(c)}{\ge}\frac{\hat{p}}{2}\left\Vert \PIOc\P_{\bar{\T}}\Delta_{l}\right\Vert _{F}^{2}.
\end{align*}
where $(a)$ follows from Part (c) of Lemma~\ref{lem:TandT0} and
the fact that $\P_{T_{0}}$ is a projection when restricted to $I_{0}^{c}$,
$(b)$ uses Part (c) of Lemma~\ref{lem:TandT0} again, and $(c)$
uses (\ref{eq:invertibility}). Combining the last three equations
proves the lemma.
\end{proof}
Back to the proof of Proposition~\ref{prop:opt_cond}. Suppose $\left(L^{*},C^{*}\right)=(\bar{L}+\Delta_{l},\;\bar{C}+\Delta_{c})$
is an optimal solution to (\ref{eq:L12formulation}), with $\P_{\Omega}\Delta_{l}+\P_{\Omega}\Delta_{c}=0$.
Take any matrix $F\in\bar{\T}^{\perp}$ such that $\left\Vert F\right\Vert =1$,
$\left\langle F,\; P_{\bar{T}^{\perp}}\Delta_{l}\right\rangle =\left\Vert P_{\bar{T}^{\perp}}\Delta_{l}\right\Vert _{\ast}$
and another matrix $G\in\bar{\I}^{c}$ such that $\left\Vert G\right\Vert _{\infty,2}=1$,
$\left\langle G,\; P_{\bar{\I}^{c}}\Delta_{c}\right\rangle =\left\Vert P_{\bar{\I}^{c}}\Delta_{c}\right\Vert _{1,2}=\left\Vert P_{\bar{\I}^{c}\cap\I_{0}}\Delta_{c}\right\Vert _{1,2}+\left\Vert \PIOc\Delta_{c}\right\Vert _{1,2}$.
Then $\bar{U}\bar{V}^{\top}+F$ is a subgradient of $\left\Vert \bar{L}\right\Vert _{\ast}$
and $P_{\bar{\I}}\bar{Q}+\lambda G$ is a subgradient of $\lambda\left\Vert \bar{C}\right\Vert _{1,2}$.
By optimality of~$\left(L^{*},C^{*}\right)$, we have
\begin{align*}
0 & \ge\left\Vert \bar{L}+\Delta_{l}\right\Vert _{\ast}+\lambda\left\Vert \bar{C}+\Delta_{c}\right\Vert _{1,2}-\left\Vert \bar{L}\right\Vert _{\ast}-\lambda\left\Vert \bar{C}\right\Vert _{1,2}\\
 & \overset{(i)}{\ge}\left\langle \bar{U}\bar{V}^{\top}+F,\Delta_{l}\right\rangle +\left\langle \PIO\bar{Q}+\lambda G,\Delta_{c}\right\rangle \\
 & \overset{(ii)}{=}\left\Vert \P_{\bar{\T}^{\bot}}\Delta_{l}\right\Vert _{\ast}+\lambda\left(\left\Vert \P_{\bar{\I}^{c}\cap\I_{0}}\Delta_{c}\right\Vert _{1,2}+\left\Vert \PIOc\Delta_{c}\right\Vert _{1,2}\right)+\left\langle \bar{U}\bar{V}^{\top}-\bar{Q},\Delta_{l}\right\rangle +\left\langle \PI\bar{Q}-\bar{Q},\Delta_{c}\right\rangle 
\end{align*}
where~$(i)$ follows from the definition of a subgradient, and~$(ii)$
is due to Condition $(a)$ and $\P_{\Omega}\Delta_{l}+\P_{\Omega}\Delta_{c}=0$.
Now observe that Conditions~$3(b)$ and~$3(c)$ imply
\[
\left\langle \bar{U}\bar{V}^{\top}-\bar{Q},\;\Delta_{l}\right\rangle =\left\langle \PT D-\P_{\bar{\T}^{\bot}}\bar{Q},\;\Delta_{l}\right\rangle \ge-\sqrt{\frac{\hat{p}}{2}}\min\left\{ \frac{1}{4},\frac{\lambda}{4}\right\} \left\Vert \PIOc\PT\Delta_{l}\right\Vert _{F}-\frac{1}{2}\left\Vert \P_{\bar{\T}^{\bot}}\Delta_{l}\right\Vert _{\ast},
\]
and Conditions~$3(e)$ and~$3(f)$ imply
\[
\left\langle \PI\bar{Q}-\bar{Q},\Delta_{c}\right\rangle \ge-\lambda\left\Vert \P_{\bar{\I}^{c}\cap\I_{0}}\Delta_{c}\right\Vert _{1,2}-\frac{\lambda}{2}\left\Vert \PIOc\Delta_{c}\right\Vert _{1,2}
\]
Putting together, we obtain
\begin{align*}
0 & \ge\frac{1}{2}\left\Vert \P_{\bar{\T}^{\bot}}\Delta_{l}\right\Vert _{\ast}+\frac{1}{2}\lambda\left\Vert \PIOc\Delta_{c}\right\Vert _{1,2}-\sqrt{\frac{\hat{p}}{2}}\left\Vert \PIOc\PT\Delta_{l}\right\Vert _{F}\\
 & \overset{(iii)}{\ge}\frac{1}{2}\left\Vert \P_{\bar{\T}^{\bot}}\Delta_{l}\right\Vert _{\ast}+\frac{1}{2}\lambda\left\Vert \PIOc\Delta_{c}\right\Vert _{1,2}-\min\left\{ \frac{1}{4},\frac{\lambda}{4}\right\} \left(\left\Vert \P_{\bar{\T}^{\bot}}\Delta_{l}\right\Vert _{*}+\left\Vert \PIOc\Delta_{c}\right\Vert _{1,2}\right)\\
 & \ge\frac{1}{4}\left\Vert \P_{\bar{\T}^{\bot}}\Delta_{l}\right\Vert _{F}+\frac{1}{4}\lambda\left\Vert \PIOc\Delta_{c}\right\Vert _{1,2}\ge0,
\end{align*}
where $(iii)$ follows from Lemma~\ref{lem:Tperp>T-1NonOr}. Therefore,
we must have
\[
\left\Vert \P_{\bar{\T}^{\bot}}\Delta_{l}\right\Vert _{F}=\left\Vert \PIOc\Delta_{c}\right\Vert _{1,2}=0,
\]
which means $\Delta_{l}\in\bar{T}$, $\P_{\I_{0}^{c}}\Delta_{c}=0$
and $\P_{I_{0}}C^{*}=C^{*}$. It follows that $\PTO\PIOc\PT\Delta_{l}=\PIOc\PT\Delta_{l}=\PIOc\Delta_{l}$
by Part~(c) of Lemma~\ref{lem:TandT0}, and $\P_{\tilde{\Omega}}\Delta_{l}=-\P_{\tilde{\Omega}}\Delta_{c}=0$,
so $\PIOc\Delta_{l}\in\T_{0}\cap\tilde{\Omega}$. But this intersection
is trivial by Condition~1 in the proposition, so $\PIOc\Delta_{l}=0$
and thus $\PIOc L^{*}=L_{0}$. Furthermore, we have
\begin{align*}
\P_{\bar{U}^{\perp}}\Delta_{l} & =\P_{\bar{U}^{\perp}}\PT\Delta_{l}=\P_{\bar{V}}\P_{\bar{U}^{\perp}}\Delta_{l}
\end{align*}
and thus $\P_{\bar{U}^{\perp}}\Delta_{l}\in\textrm{range}\left(\P_{\bar{V}}\right)$.
But we also have $\P_{\bar{U}^{\perp}}\Delta_{l}=\P_{\bar{U}^{\perp}}\left(\P_{I_{0}^{c}}+\P_{I_{0}}\right)\Delta_{l}=\P_{\bar{U}^{\perp}}\P_{I_{0}}\Delta_{l}\in\mathcal{I}_{0}$.
This implies $\P_{\bar{U}^{\perp}}\Delta_{l}=0$ by Condition~2 in
the proposition. This shows that $\P_{U_{0}}\Delta_{l}=\P_{\bar{U}}\Delta_{l}=\Delta_{l}$,
where the first equality follows from part~(a) of Lemma~\ref{lem:TandT0}.
This completes the proof of the proposition.

\section{Proof of Lemma~\ref{lem:PVPI}}

 For any matrices $A$ and $B$, we have 
\begin{equation}
\left\Vert AB\right\Vert _{F}\le\left\Vert A\right\Vert \left\Vert B\right\Vert _{F},\label{eq:opF_submultipy}
\end{equation}
which follows from $\left\Vert AB\right\Vert _{F}^{2}=\sum_{j}\left\Vert ABe_{j}\right\Vert _{2}^{2}\le\sum_{j}\left\Vert A\right\Vert ^{2}\left\Vert Be_{j}\right\Vert _{2}^{2}=\left\Vert A\right\Vert ^{2}\left\Vert B\right\Vert _{F}^{2}.$
Using part~$(d)$ of Lemma~\ref{lem:H}, we know $\P_{I_{0}}\bar{V}^{\top}=\lambda\bar{U}^{\top}\bar{H}'.$
It follows that for any matrix $Z$, 
\[
\P_{\bar{V}}\PIO\P_{\bar{V}}(Z)=\PIO\left(Z\bar{V}\bar{V}^{\top}\right)\bar{V}\bar{V}^{\top}=Z\bar{V}\left(\P_{I_{0}}\bar{V}^{\top}\right)\left(\P_{I_{0}}\bar{V}^{\top}\right)^{\top}\bar{V}^{\top}=\lambda^{2}Z\bar{V}\left(\bar{U}^{\top}\bar{H}'\right)\left(\bar{H}'^{\top}\bar{U}\right)\bar{V}^{\top}.
\]
Using (\ref{eq:opF_submultipy}), we obtain 
\[
\left\Vert \P_{\bar{V}}\PIO\P_{\bar{V}}(Z)\right\Vert _{F}\le\lambda^{2}\left\Vert Z\right\Vert _{F}\left\Vert \bar{V}\right\Vert ^{2}\left\Vert \bar{U}\right\Vert ^{2}\left\Vert \bar{H}'\right\Vert ^{2}\overset{(i)}{\le}\lambda^{2}\gamma n\left\Vert Z\right\Vert _{F}\le\frac{1}{2}\left\Vert Z\right\Vert _{F},
\]
where the inequality $(i)$ follows from $\left\Vert \bar{H}'\right\Vert _{\infty,2}\le1$
and $\bar{H}'\in I_{0}$ has at most $\gamma n$ non-zero columns.
The second part of the lemma is a proved in similar manner using the
sub-multiplicity of the matrix spectral norm.

\section{Proof of Lemma~\ref{lem:Q_properties}}

By part~$(d)$ of Lemma~\ref{lem:H}, we have 
\[
\P_{\I_{0}}Q=\bar{U}\P_{I_{0}}\bar{V}^{\top}+\lambda\bar{H}'-\lambda\P_{U_{0}}\bar{H}'=\lambda\bar{H}'.
\]
Using~(\ref{eq:inverse}) and $\P_{U_{0}}=\P_{\bar{U}}$, we have
\[
\P_{\bar{\T}}Q=\bar{U}\bar{V}^{\top}+\left(\lambda\P_{\bar{U}}\bar{H}'+\lambda\P_{\bar{U}^{\bot}}\P_{\bar{V}}\bar{H}'\right)-\lambda\P_{U_{0}}\bar{H}'-\lambda\left(\P_{\bar{V}}\P_{\I_{0}^{c}}\P_{\bar{V}}\right)\mc B\P_{\bar{V}}\P_{\bar{U}^{\bot}}\bar{H}'=\bar{U}\bar{V}^{\top}.
\]
This proves the two equalities in the lemma. Observe that $\bar{H}'\in I_{0}$
has at most $n_{c}=\gamma n$ non-zero columns, each of which has
norm at most one by part~$(b)$ and~$(c)$ of Lemma~\ref{lem:H}.
It follows that $\left\Vert \bar{H}'\right\Vert \le\left\Vert \bar{H}'\right\Vert _{F}\le\sqrt{\gamma n}$.
We also have $\left\Vert \P_{\bar{V}}\P_{\bar{U}^{\bot}}\bar{H}'\right\Vert \le\left\Vert Id-\bar{U}\bar{U}^{\top}\right\Vert \left\Vert \bar{H}'\right\Vert \left\Vert \bar{V}\bar{V}^{\top}\right\Vert \le\left\Vert \bar{H}'\right\Vert $
by sub-multiplicity of the spectral norm. This proves the first set
of inequalities in the lemma.

\section{Proof of Lemmas in Section~\ref{sec:dual_validate}}

In this section, we prove the lemmas used in Section~\ref{sec:dual_validate}.

\subsection{Proof of Lemma \ref{lem:PV_H}}

Recall that $D_{0}^{U}:=\bar{U}\PIc(\bar{V})$, and $D{}_{0}^{V}:=\P_{\bar{U}^{\bot}}\P_{V_{0}}\P_{\I_{0}^{c}}\mc B\P_{\bar{V}}\left(\lambda\bar{H}'\right)$.
The first three inequalities follow directly from the incoherence
Assumption~\ref{asm:incoherence} and part (b) of Lemma~\ref{lem:TandT0}.
Now, by Assumption~\ref{asm:sampling_2} and part~$(a)$ of Lemma~\ref{lem:H},
we know each column of $\bar{H}'$ has at most $2\rho m$ non-zeros.
Because $\bar{U}$ has the same column space as $U_{0}$ by Lemma~\ref{lem:TandT0},
$\bar{U}$ satisfies the same incoherence property as $U_{0}$ given
in Assumption~\ref{asm:incoherence}. Therefore, we have 
\[
\left\Vert e_{a}^{\top}\bar{H}^{\top}\bar{U}\right\Vert _{2}\le\left\Vert \bar{H}e_{a}\right\Vert _{1}\left\Vert \bar{U}^{\top}\right\Vert _{\infty,2}\le\sqrt{2\rho m}\left\Vert \bar{H}e_{a}\right\Vert _{2}\cdot\sqrt{\frac{\mu r}{m}}=\sqrt{2\rho\mu r}.
\]
It follows that 
\[
\left\Vert \bar{H}^{\top}\bar{U}\right\Vert \le\left\Vert \bar{H}^{\top}\bar{U}\right\Vert _{F}\le\sqrt{\gamma n}\sqrt{2\rho\mu r}=\sqrt{\gamma n}\sqrt{2\beta\hat{p}\mu r},
\]
where we use the definition $\beta:=\frac{\rho}{\hat{p}}$. Using
Lemma~\ref{lem:PVPI} and the fact that $\left\Vert \bar{H}\right\Vert \le\sqrt{\gamma n}$,
we get 
\begin{align}
\left\Vert \mc B\left(\bar{H}\bar{H}^{\top}\bar{U}\bar{V}^{\top}\right)\right\Vert  & =\left\Vert \P_{\I_{0}^{c}}\P_{\bar{V}}\sum_{i=0}^{\infty}\left(\P_{\bar{V}}\PIO\P_{\bar{V}}\right)^{i}\left(\bar{H}\bar{H}^{\top}\bar{U}\bar{V}^{\top}\right)\right\Vert \nonumber \\
 & \le\left(\sum_{i=0}^{\infty}\frac{1}{2}\right)\left\Vert \bar{H}\right\Vert \left\Vert \bar{H}^{\top}\bar{U}\right\Vert \left\Vert \bar{V}^{\top}\right\Vert \le4\gamma n\sqrt{\beta\hat{p}\mu r}.\label{eq:pvh1}
\end{align}
On the other hand, note that by part~$(d)$ of Lemma~\ref{lem:H}
we have $\P_{I_{0}}\bar{V}^{\top}=\lambda\bar{U}^{\top}\bar{H}'.$
Since $\bar{H}'\in I_{0}$, we have 
\begin{align}
\left\Vert \lambda\left(\P_{\bar{U}^{\bot}}\P_{V_{0}}\P_{\I_{0}^{c}}\mc B\P_{\bar{V}}\bar{H}'\right)e_{j}\right\Vert _{2} & =\lambda\left\Vert \left(Id-\bar{U}\bar{U}^{\top}\right)\mc B\left(\bar{H}'\bar{V}\bar{V}^{\top}\right)V_{0}V_{0}^{\top}e_{j}\right\Vert _{2}\nonumber \\
 & =\lambda\left\Vert \left(Id-\bar{U}\bar{U}^{\top}\right)\mc B\left(\bar{H}'(\P_{I_{0}}\bar{V}^{\top})^{\top}\bar{V}^{\top}\right)V_{0}V_{0}^{\top}e_{j}\right\Vert _{2}\nonumber \\
 & =\lambda^{2}\left\Vert \left(Id-\bar{U}\bar{U}^{\top}\right)\mc B\left(\bar{H}'\bar{H}^{'\top}\bar{U}\bar{V}^{\top}\right)V_{0}V_{0}^{\top}e_{j}\right\Vert _{2}.\label{eq:pvh2}
\end{align}
Combining~(\ref{eq:pvh1}) and~(\ref{eq:pvh2}), we obtain 
\begin{align*}
\left\Vert D_{0}^{V}\right\Vert _{\infty,2}=\max_{j}\left\Vert \lambda\left(\P_{\bar{U}^{\bot}}\P_{V_{0}}\P_{\I_{0}^{c}}\mc B\P_{\bar{V}}\bar{H}'\right)e_{j}\right\Vert _{2} & \le\lambda^{2}\left\Vert Id-\bar{U}\bar{U}^{\top}\right\Vert \left\Vert \mc B\left(\bar{H}\bar{H}^{\top}\bar{U}\bar{V}^{\top}\right)\right\Vert \max_{j}\left\Vert V_{0}V_{0}^{\top}e_{j}\right\Vert _{2}\\
 & \le\lambda^{2}\cdot1\cdot4\gamma n\sqrt{\beta\hat{p}\mu r}\cdot\sqrt{\frac{\mu r}{n}}=4\lambda^{2}\gamma\mu r\sqrt{\beta\hat{p}n},
\end{align*}
which proves the forth equation in the lemma. The last equation in
the lemma can be established in a similar manner using Lemma~\ref{lem:PVPI}:
\begin{align*}
\left\Vert D_{0}^{V}\right\Vert _{F}=\left\Vert \P_{\bar{U}^{\bot}}\P_{V_{0}}\mc B\P_{\bar{V}}\left(\lambda\bar{H}'\right)\right\Vert _{F} & \le\lambda^{2}\left\Vert Id-\bar{U}\bar{U}^{\top}\right\Vert \left\Vert \mc B\left(\bar{H}\bar{H}^{\top}\bar{U}\bar{V}^{\top}\right)\right\Vert _{F}\left\Vert V_{0}V_{0}^{\top}\right\Vert \\
 & \le2\lambda^{2}\left\Vert \bar{H}\bar{H}^{\top}\bar{U}\bar{V}^{\top}\right\Vert _{F}\\
 & \le2\lambda^{2}\left\Vert \bar{H}\right\Vert \left\Vert \bar{H}^{\top}\bar{U}\right\Vert _{F}\left\Vert \bar{V}^{\top}\right\Vert \\
 & \le2\lambda^{2}\cdot\sqrt{\gamma n}\cdot\sqrt{\gamma n}\sqrt{2\beta\hat{p}\mu r}\cdot1.
\end{align*}

\subsection{Proof of Lemma~\ref{lem:inf_2}}

Let $e_{i}$ be the $i$-th standard basis whose dimension will become
clear in the context. The following inequality is used repeatedly:
from the incoherence Assumption~\ref{asm:incoherence}, we have 
\begin{equation}
\left\Vert \P_{T_{0}}\left(e_{i}e_{j}^{\top}\right)\right\Vert _{F}^{2}=\left\Vert \P_{U_{0}}e_{i}\right\Vert _{2}^{2}+\left\Vert \P_{V_{0}}e_{j}\right\Vert _{2}^{2}-\left\Vert \P_{U_{0}}e_{i}\right\Vert _{2}^{2}\left\Vert \P_{V_{0}}e_{j}\right\Vert _{2}^{2}\le\frac{2\mu r}{n\wedge m},\forall i\in[m],j\in[n+n_{c}].\label{eq:basic_inequality}
\end{equation}
We also need the matrix Bernstein inequality, restated below.
\begin{thm}
[Matrix Bernstein \cite{tropp2010matrixmtg}]\label{lem:matrix_bernstein}Let $X_{1},\ldots,X_{N}\in\mathbb{R}^{m\times n}$
be independent zero mean random matrices. Suppose there exist two
numbers $B$ and $\sigma^{2}$ such that 
\[
\max\left\{ \left\Vert \mathbb{E}\sum_{k=1}^{N}X_{k}X_{k}^{\top}\right\Vert ,\left\Vert \mathbb{E}\sum_{k=1}^{N}X_{k}^{\top}X_{k}\right\Vert \right\} \le\sigma^{2}
\]
and $\left\Vert X_{k}\right\Vert \le B$ almost surely for all $k$.
Then with probability at least $1-2(m+n)^{-12}$, we have 
\[
\left\Vert \sum_{k=1}^{N}X_{k}\right\Vert \le20B\log(m+n)+\sqrt{50\sigma^{2}\log(m+n)}.
\]

\end{thm}
We now turn to the proof of the lemma.
\begin{proof}
(of Lemma~\ref{lem:inf_2}) 
Observe that $\frac{1}{\hat{p}}\mathcal{P}_{T_{0}}\mathcal{P}_{\tilde{\Omega}}\mathcal{P}_{T_{0}}Z-\mathcal{P}_{T_{0}}Z\in I_{0}^{c}$
for any matrix $Z\in T_{0}\subseteq I_{0}^{c}$. Fix an index $b\in\I_{0}^{c}$.
For each $(i,j)\in[m]\times\I_{0}^{c}$, let $\delta_{(ij)}$ be the
indicator variable which equals one if and only if $(i,j)\in\tilde{\Omega}$.
We have $\mathbb{P}\left[\delta_{(ij)}=1\right]=\hat{p}$ by assumption~\ref{asm:sampling_2}.
Define 
\[
S_{(ij)}:=\left(\frac{1}{\hat{p}}\delta_{(ij)}-1\right)Z_{ij}\P_{T_{0}}(e_{i}e_{j}^{\top})e_{b},
\]
which is a column vector in $\mathbb{R}^{m}$. Since $\P_{T_{0}}Z=Z$
for $Z\in T_{0}$, the $b$-th column of the matrix $\left(\frac{1}{p}\P_{T_{0}}\P_{\tilde{\Omega}}-\P_{T_{0}}\right)Z$
can be written as
\[
\left(\left(\frac{1}{\hat{p}}\P_{T_{0}}\P_{\Omegat}-\mathcal{I}\right)Z\right)e_{b}=\sum_{(i,j)\in[m]\times\I_{0}^{c}}S_{(ij)},
\]
which is the sum of independent vectors in $\mathbb{R}^{m}$. Note
that $\mathbb{E}\left[S_{(ij)}\right]=0$ and 

\[
\left\Vert S_{(ij)}\right\Vert _{2}\le\left|\frac{1}{\hat{p}}\delta_{(ij)}-1\right|\left|Z_{ij}\right|\left\Vert \P_{T_{0}}(e_{i}e_{j}^{\top})\right\Vert _{F}\le\frac{1}{\hat{p}}\sqrt{\frac{2\mu r}{n\wedge m}}\left\Vert Z\right\Vert _{\infty},\;\text{a.s.},
\]
where the second inequality follows from~(\ref{eq:basic_inequality}).
We also have 
\[
\left|\mathbb{E}\left[\sum_{(i,j)\in[m]\times\I_{0}^{c}}S_{(ij)}^{\top}S_{(ij)}\right]\right|=\left|\sum_{i,j}\mathbb{E}\left[\left(\frac{1}{p}\delta_{(ij)}-1\right)^{2}\right]Z_{ij}^{2}\left\Vert \P_{T_{0}}\left(e_{i}e_{j}^{\top}\right)e_{b}\right\Vert _{2}^{2}\right|=\frac{1-\hat{p}}{\hat{p}}\sum_{i,j}Z_{ij}^{2}\left\Vert \P_{T_{0}}\left(e_{i}e_{j}^{\top}\right)e_{b}\right\Vert _{2}^{2}.
\]
We bound the term in the summand in the last R.H.S. Recall that $Id$
denotes the identity matrix. For each $(i,j)\in[m]\times\I_{0}^{c}$,
we have
\begin{align*}
\left\Vert \P_{T_{0}}\left(e_{i}e_{j}^{\top}\right)e_{b}\right\Vert _{2} & =\left\Vert U_{0}U_{0}^{\top}e_{i}e_{j}^{\top}e_{b}+\left(Id-U_{0}U_{0}^{\top}\right)e_{i}e_{j}^{\top}V_{0}V_{0}^{\top}e_{b}\right\Vert _{2}\\
 & =\begin{cases}
\left\Vert U_{0}U_{0}^{\top}e_{i}+\left(Id-U_{0}U_{0}^{\top}\right)e_{i}\left\Vert V_{0}^{\top}e_{b}\right\Vert _{2}^{2}\right\Vert _{2}, & \text{ if }j=b,\\
\left\Vert \left(Id-U_{0}U_{0}^{\top}\right)e_{i}e_{j}^{\top}V_{0}V_{0}^{\top}e_{b}\right\Vert _{2}, & \text{ if }j\neq b,
\end{cases}\\
 & \le\begin{cases}
\left\Vert U_{0}^{\top}e_{i}\right\Vert _{2}+\left\Vert V_{0}^{\top}e_{b}\right\Vert _{2}^{2}, & \text{ if }j=b,\\
\left|e_{j}^{\top}V_{0}V_{0}^{\top}e_{b}\right|, & \text{ if }j\neq b,
\end{cases}\\
 & \le\begin{cases}
2\sqrt{\frac{\mu r}{m\wedge n}}, & \text{ if }j=b,\\
\left|e_{j}^{\top}V_{0}V_{0}^{\top}e_{b}\right|, & \text{ if }j\neq b,
\end{cases}
\end{align*}
where in the last inequality we use $\left\Vert V_{0}^{\top}e_{b}\right\Vert _{2}\le1$
and the incoherence Assumption~\ref{asm:incoherence}. It follows
that 
\begin{align*}
\left\Vert \mathbb{E}\left[\sum_{i,j}S_{(ij)}S_{(ij)}^{\top}\right]\right\Vert  & =\left|\mathbb{E}\left[\sum_{i,j}S_{(ij)}^{\top}S_{(ij)}\right]\right|\\
 & \le\frac{1}{\hat{p}}\sum_{i\in[m],j=b}Z_{ij}^{2}\frac{4\mu r}{n\wedge m}+\frac{1}{\hat{p}}\sum_{i\in[m],j\neq b}Z_{ij}^{2}\left|e_{j}^{\top}VV^{\top}e_{b}\right|^{2}\\
 & =\frac{4\mu r}{\hat{p}(n\wedge m)}\sum_{i}Z_{ib}^{2}+\frac{1}{\hat{p}}\sum_{j\neq b}\left|e_{j}^{\top}VV^{\top}e_{b}\right|^{2}\sum_{i}Z_{ij}^{2}\\
 & \le\frac{4}{\hat{p}}\frac{\mu r}{n\wedge m}\left\Vert Z\right\Vert _{\infty,2}^{2}+\frac{1}{\hat{p}}\left\Vert VV^{\top}e_{b}\right\Vert _{2}^{2}\left\Vert Z\right\Vert _{\infty,2}^{2}\\
 & \le\frac{4}{\hat{p}}\frac{\mu r}{n\wedge m}\left\Vert Z\right\Vert _{\infty,2}^{2}+\frac{1}{\hat{p}}\frac{\mu r}{n}\cdot\left\Vert Z\right\Vert _{\infty,2}^{2}\le\frac{5\mu r}{\hat{p}(n\wedge m)}\left\Vert Z\right\Vert _{\infty,2}^{2}.
\end{align*}
Treating $\{S_{(ij)}\}$ as zero-padded $m\times n$ matrices and
applying the Matrix Bernstein inequality in Theorem~\ref{lem:matrix_bernstein},
we obtain that with probability at least $1-2(m+n)^{-12}$,
\begin{align*}
\left\Vert \left(\left(\frac{1}{\hat{p}}\P_{T_{0}}\P_{\Omegat}-\mathcal{I}\right)Z\right)e_{b}\right\Vert _{2} & \le20\frac{1}{\hat{p}}\sqrt{\frac{2\mu r}{n\wedge m}}\left\Vert Z\right\Vert _{\infty}\log(m+n)+\sqrt{50\cdot\frac{5\mu r}{\hat{p}(n\wedge m)}\left\Vert Z\right\Vert _{\infty,2}^{2}\log(m+n)}\\
 & \le\frac{1}{2}\sqrt{\frac{\log(m+n)}{\hat{p}}}\left\Vert Z\right\Vert _{\infty}+\frac{1}{2}\left\Vert Z\right\Vert _{\infty,2},
\end{align*}
where the second inequality holds provided $c_{0}$ in the condition
of the lemma is sufficiently large. In a similar fashion we can prove
that for each $a\in[m]$ and with probability at least $1-2(m+n)^{-12}$,
\begin{align*}
\left\Vert e_{a}^{\top}\left(\left(\frac{1}{\hat{p}}\P_{T_{0}}\P_{\Omegat}-\mathcal{I}\right)Z\right)\right\Vert  & \le\frac{40}{\hat{p}}\sqrt{\frac{\mu r}{n\wedge m}}\left\Vert Z\right\Vert _{\infty}\log(m+n)+\sqrt{\frac{250\mu r}{\hat{p}(n\wedge m)}\log(m+n)}\left\Vert Z\right\Vert _{\infty,2}\\
 & \le\frac{1}{2}\sqrt{\frac{\log(m+n)}{\hat{p}}}\left\Vert Z\right\Vert _{\infty}+\frac{1}{2}\left\Vert Z^{\top}\right\Vert _{\infty,2}.
\end{align*}
 The lemma follows from a union bound over all indices $a\in[m]$
and $b\in\I_{0}^{c}$. 
\end{proof}

\subsection{Proof of Lemma~\ref{lem:inf_2_order_1}}

Observe that $\frac{1}{\hat{p}}\mathcal{P}_{\tilde{\Omega}}Z-Z\in I_{0}^{c}$
for any matrix $Z\in T_{0}\subseteq I_{0}^{c}$. Fix an index $b\in\I_{0}^{c}$.
For each $i\in[m]$, we recall that the indicator variable $\delta_{(ib)}$
defined in the last section, and define the vector 
\[
\xi_{(i)}:=Z_{ib}\left(\frac{1}{\hat{p}}\delta_{(ib)}-1\right)e_{i}\in\mathbb{R}^{m}.
\]
Then the $b$-th column of the matrix $\frac{1}{\hat{p}}\mathcal{P}_{\tilde{\Omega}}Z-Z$
can be written as 
\[
\left(\frac{1}{\hat{p}}\mathcal{P}_{\tilde{\Omega}}Z-Z\right)e_{b}=\sum_{i\in[m]}\xi_{(i)}.
\]
which is the sum of independent vectors. Note that each $\xi_{(i)}$
has mean zero and satisfies $\left\Vert \xi_{(i)}\right\Vert _{2}\le\left(\frac{1}{\hat{p}}-1\right)Z_{ib}\le\frac{1}{\hat{p}}\left\Vert Z\right\Vert _{\infty}$
a.s. Moreover, we have 
\begin{align*}
\max\left\{ \left\Vert \mathbb{E}\sum_{i\in[m]}\xi_{(i)}^{\top}\xi_{(i)}\right\Vert ,\left\Vert \mathbb{E}\sum_{i\in[m]}\xi_{(i)}\xi_{(i)}^{\top}\right\Vert \right\}  & =\max\left\{ \left|\frac{1-\hat{p}}{\hat{p}}\sum_{i}Z_{ib}^{2}\right|,\left\Vert \frac{1-\hat{p}}{\hat{p}}\sum_{i}Z_{ib}^{2}e_{i}e_{i}^{\top}\right\Vert \right\} \le\frac{1}{\hat{p}}\left\Vert Z\right\Vert _{\infty,2}^{2}.
\end{align*}
Treating $\{\xi_{(i)}\}$ as zero-padded $m\times n$ matrices and
applying the matrix Bernstein inequality in Theorem~\ref{lem:matrix_bernstein},
we obtain that with probability at least $1-2(m+n)^{-12}$,
\[
\left\Vert \left(\frac{1}{\hat{p}}\mathcal{P}_{\tilde{\Omega}}Z-Z\right)e_{b}\right\Vert _{2}\le\frac{20\log(m+n)}{\hat{p}}\left\Vert Z\right\Vert _{\infty}+\sqrt{\frac{50\log(m+n)}{\hat{p}}}\left\Vert Z\right\Vert _{\infty,2}.
\]
The lemma follows from a union bound over all indices $b\in\I_{0}^{c}$.

\subsection{Proof of Lemma~\ref{lem:op_inf}}

Recall the indicator variables $\left\{ \delta_{(ij)}\right\} $
defined in the last section. Since $Z\in\I_{0}^{c}$, we may write
\[
\frac{1}{\hat{p}}\P_{\hat{\Omega}}Z-Z=\sum_{(i,j)\in[m]\times\I_{0}^{c}}S_{(ij)}:=\sum_{(i,j)\in[m]\times\I_{0}^{c}}\left(\frac{1}{\hat{p}}\delta_{(ij)}-1\right)Z_{ij}e_{i}e_{j}^{\top},
\]
where $\left\{ S_{(ij)}\right\} $ are independent matrices satisfying
$\mathbb{E}[S_{(ij)}]=0$ and $\left\Vert S_{(ij)}\right\Vert \le\frac{1}{\hat{p}}\left\Vert Z\right\Vert _{\infty}.$
Moreover, we have 
\[
\mathbb{E}\sum_{(i,j)\in[m]\times\I_{0}^{c}}S_{(ij)}^{\top}S_{(ij)}=\sum_{(i,j)\in[m]\times\I_{0}^{c}}Z_{ij}^{2}e_{i}e_{j}^{\top}e_{j}e_{i}^{\top}\mathbb{E}\left(\frac{1}{\hat{p}}\delta_{ij}-1\right)^{2}=\sum_{(i,j)\in[m]\times\I_{0}^{c}}\frac{1-\hat{p}}{\hat{p}}Z_{ij}^{2}e_{i}e_{i}^{\top}
\]
and thus 
\[
\left\Vert \mathbb{E}\sum_{(i,j)\in[m]\times\I_{0}^{c}}S_{(ij)}^{\top}S_{(ij)}\right\Vert \le\frac{1}{\hat{p}}\max_{i\in[m]}\left|\sum_{j\in I_{0}^{c}}Z_{ij}^{2}\right|\le\frac{1}{\hat{p}}\left\Vert Z\right\Vert _{(\infty,2)^{2}}^{2}.
\]
We can bound $\left\Vert \mathbb{E}\sum_{(i,j)\in[m]\times\I_{0}^{c}}S_{(ij)}S_{(ij)}^{\top}\right\Vert $
in a similar way. Applying the matrix Bernstein inequality in Theorem
\ref{lem:matrix_bernstein} proves the lemma.

\bibliographystyle{plain}
\bibliography{rmc2015}

\end{document}